\documentclass[10pt,twocolumn,letterpaper]{article}

\usepackage{iccv}
\usepackage{times}
\usepackage{epsfig}
\usepackage{graphicx}
\usepackage{amsmath}
\usepackage{amsthm}
\usepackage{amssymb}
\usepackage{amsfonts}
\usepackage{booktabs}
\usepackage{mathrsfs}
\usepackage[utf8]{inputenc}
\usepackage[T1]{fontenc}
\usepackage{url}
\usepackage{nicefrac}
\usepackage{microtype}
\usepackage{xcolor}
\usepackage{multirow}
\usepackage{placeins}
\usepackage{bbm}
\usepackage{svg}
\usepackage{thm-restate}
\usepackage{enumerate}
\usepackage[caption=false]{subfig}
\usepackage{pgfplots}
\usepackage{cite}
\pgfplotsset{width=10cm,compat=1.9}
\usepgfplotslibrary{groupplots}
\usepackage{hyperref}
\hypersetup{colorlinks,allcolors=black}

\usepackage[title]{appendix}

\iccvfinalcopy % *** Uncomment this line for the final submission
\def\iccvPaperID{11116} % *** Enter the ICCV Paper ID here

\title{Noisy Image Segmentation With Soft-Dice}

\declaretheoremstyle[
    headfont=\bfseries, 
    bodyfont=\normalfont\itshape,
    headpunct={},
    spacebelow=\parsep,
    spaceabove=\parsep,
    mdframed={
        innertopmargin=6pt,
        innerbottommargin=6pt, 
        skipabove=12pt, 
        skipbelow=12pt} 
]{framedstyle}

\declaretheorem[name=Theorem]{theorem}

\declaretheorem[name=Definition]{definition}

\author{
  Marcus Nordstr\"om \thanks{Author is also affiliated with RaySearch Laboratories.}\\
  Department of Mathematics\\
  KTH Royal Institute of Technology\\
  Stockholm, Sweden \\
  {\tt\small marcno@kth.se} \\
  \and
  Henrik Hult \\
  Department of Mathematics\\
  KTH Royal Institute of Technology\\
  Stockholm, Sweden \\
  {\tt\small{hult@kth.se}} \\
  \and
  Atsuto Maki \\
  Department of Computer Science\\
  KTH Royal Institute of Technology\\
  Stockholm, Sweden \\
  {\tt\small{atsuto@kth.se}} \\
  \and
  Fredrik L\"ofman \\
  Department of Machine Learning \\
  RaySearch Laboratories \\
  Stockholm, Sweden \\
  {\tt\small fredrik.lofman@raysearchlabs.com} \\
}

\begin{document}
\maketitle

\begin{abstract}
This paper presents a study on the soft-Dice loss, one of the most popular loss functions in medical image segmentation, 
for situations where noise is present in target labels.
In particular, the set of optimal solutions are characterized and sharp bounds on the volume bias of these solutions are provided.
It is further shown that a sequence of soft segmentations converging to optimal soft-Dice also converges to optimal Dice when converted to hard segmentations using thresholding.
This is an important result because soft-Dice is often used as a proxy for
maximizing the Dice metric.
Finally, experiments confirming the theoretical results are provided.
\end{abstract}

\section{Introduction}
Many state-of-the-art methods for tasks in image analysis today are based on supervised learning methods.
For the problem of medical image segmentation, 
a family of the most commonly used models are deep neural networks with U-net like architecture~\cite{ronneberger2015u}.
They  
% have an enormous amount of 
are with numerous
supporting literature and frequently placed amongst the top submissions in medical image segmentation competitions.
For reviews, see \cite{siddique2021u} and \cite{du2020medical}.

Common to these machine learning methods, are that they are trained using some data set and some loss function.
The loss function quantifies the distance between the model's prediction associated with an image and the true labels associated with the same image.
Furthermore, this quantification is done in such a way that it can be targeted by continuous optimization methods, like the stochastic gradient descent schemes which are 
% commonly
widely used in the field. 
For segmentation problems only consisting of one target structure, that is, two class problems, binary cross-entropy has long been considered the standard choice.
This loss function has several convenient theoretical properties, e.g. being convex, and is known to be reliable in practice.
\begin{figure}[htb!]
  \centering

 % j\begin{groupplot}[group style={group size=7 by 9, horizontal sep=0.05cm, vertical sep=0.05cm},height=3.70cm,width=3.70cm,xmin=0,xmax=1,ymin=0,ymax=1, xticklabels=\empty, yticklabels=\empty]
\begin{tikzpicture}
  
  \begin{groupplot}[group style={group size=3 by 3, horizontal sep=0.05cm, vertical sep=0.05cm},height=4.2cm,width=4.2cm,xmajorgrids,ymajorgrids,xtick={0.4,0.7,1.0,1.3,1.6},xmin=0.4,xmax=1.6,ytick={0.4,0.7,1.0,1.3,1.6},ymin=0.4,ymax=1.6]
  
    \nextgroupplot[title=marginals,xtick=\empty,ytick=\empty,yticklabel pos = right, xmin=0, xmax=1,ymin=0,ymax=1]
    \addplot graphics[xmin=0,xmax=1,ymin=0,ymax=1] {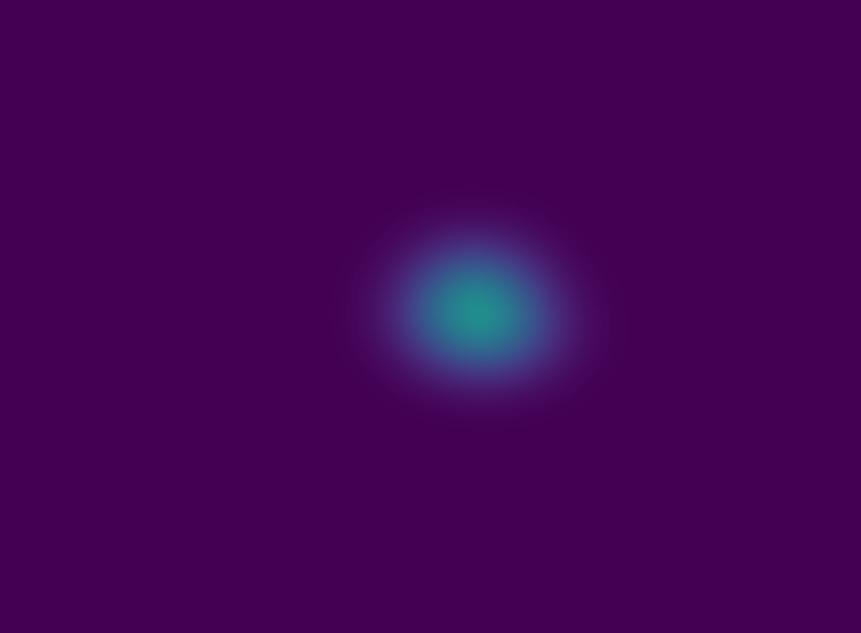};
    
    \nextgroupplot[title=cross-entropy,xtick=\empty,ytick=\empty,yticklabel pos = right, xmin=0, xmax=1,ymin=0,ymax=1]
    \addplot graphics[xmin=0,xmax=1,ymin=0,ymax=1] {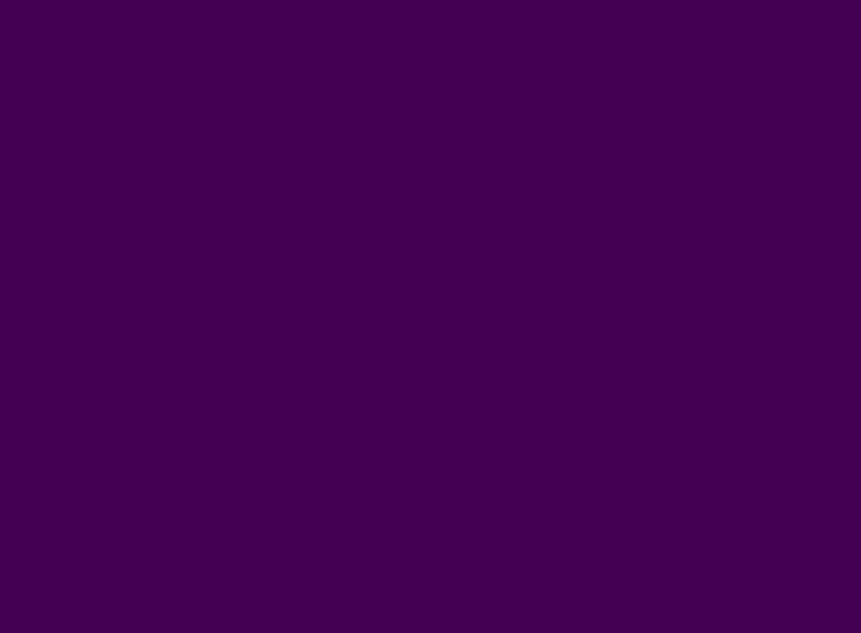};
    
    \nextgroupplot[title=soft-Dice,xtick=\empty,ytick=\empty,yticklabel pos = right, xmin=0, xmax=1,ymin=0,ymax=1]
    \addplot graphics[xmin=0,xmax=1,ymin=0,ymax=1] {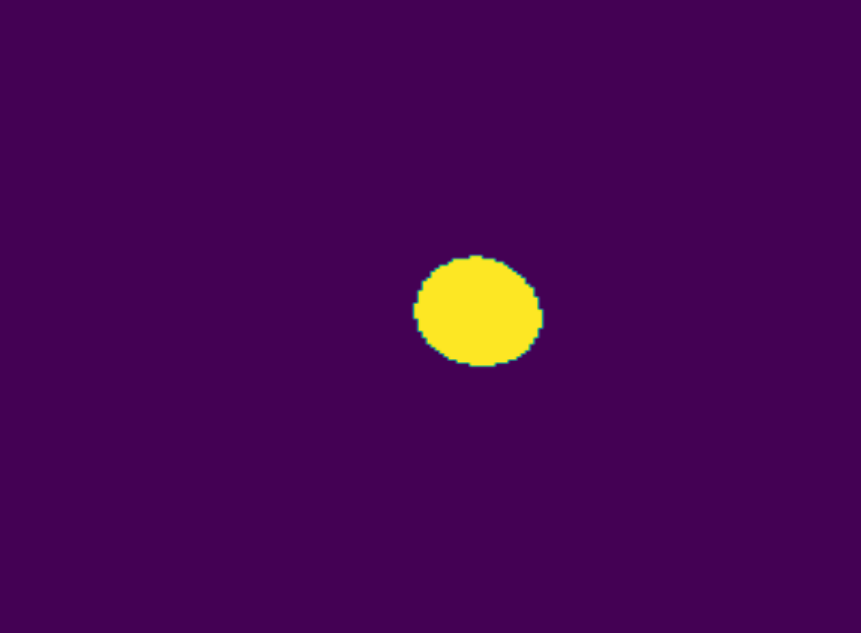};
    %%%%%%%%%%%%%%%%%%
    
    \nextgroupplot[xtick=\empty,ytick=\empty,yticklabel pos = right, xmin=0, xmax=1,ymin=0,ymax=1]
    \addplot graphics[xmin=0,xmax=1,ymin=0,ymax=1] {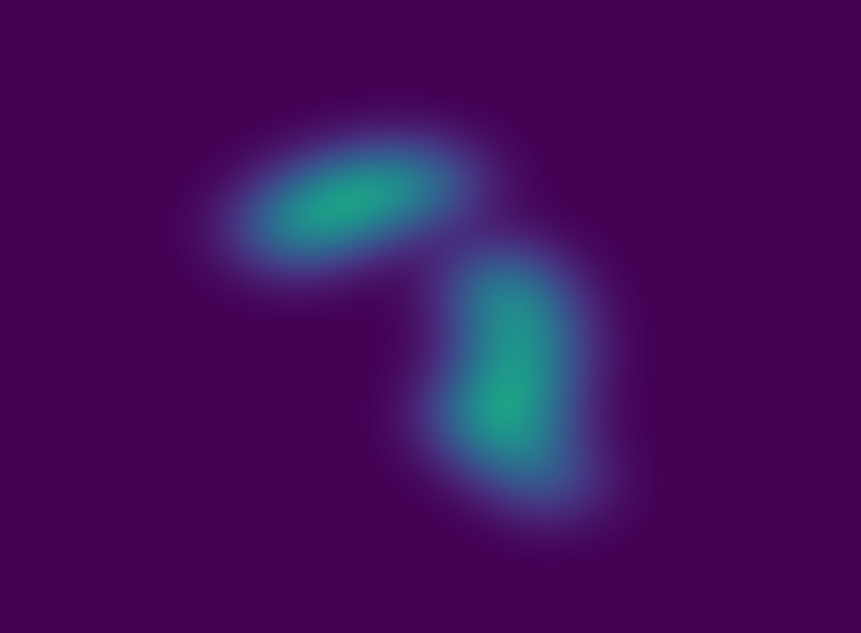};

    \nextgroupplot[xtick=\empty,ytick=\empty,yticklabel pos = right, xmin=0, xmax=1,ymin=0,ymax=1]
    \addplot graphics[xmin=0,xmax=1,ymin=0,ymax=1] {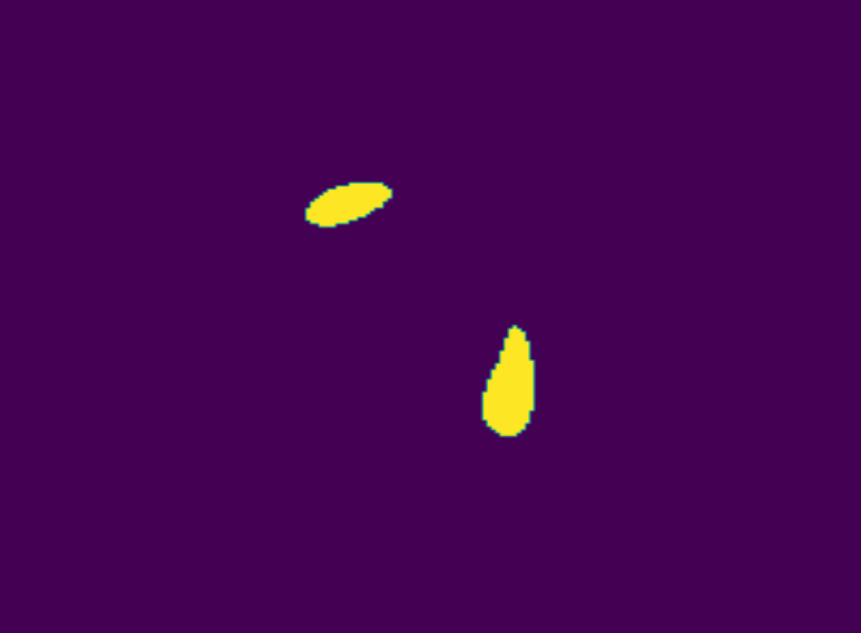};
    
    \nextgroupplot[xtick=\empty,ytick=\empty,yticklabel pos = right, xmin=0, xmax=1,ymin=0,ymax=1]
    \addplot graphics[xmin=0,xmax=1,ymin=0,ymax=1] {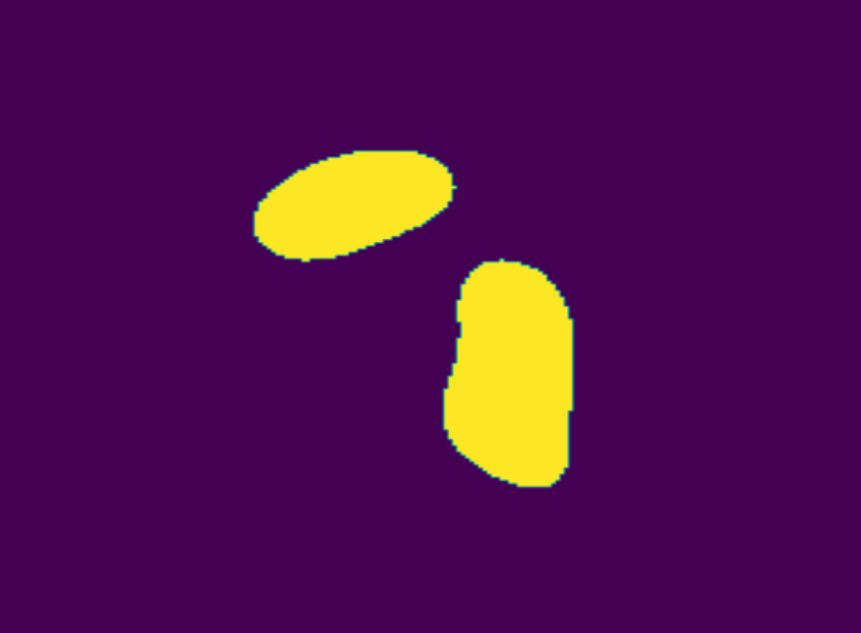};
    
    %%%%%%%%%%%%%%%%%%%%
    
    \nextgroupplot[xtick=\empty,ytick=\empty,yticklabel pos = right, xmin=0, xmax=1,ymin=0,ymax=1]
    \addplot graphics[xmin=0,xmax=1,ymin=0,ymax=1] {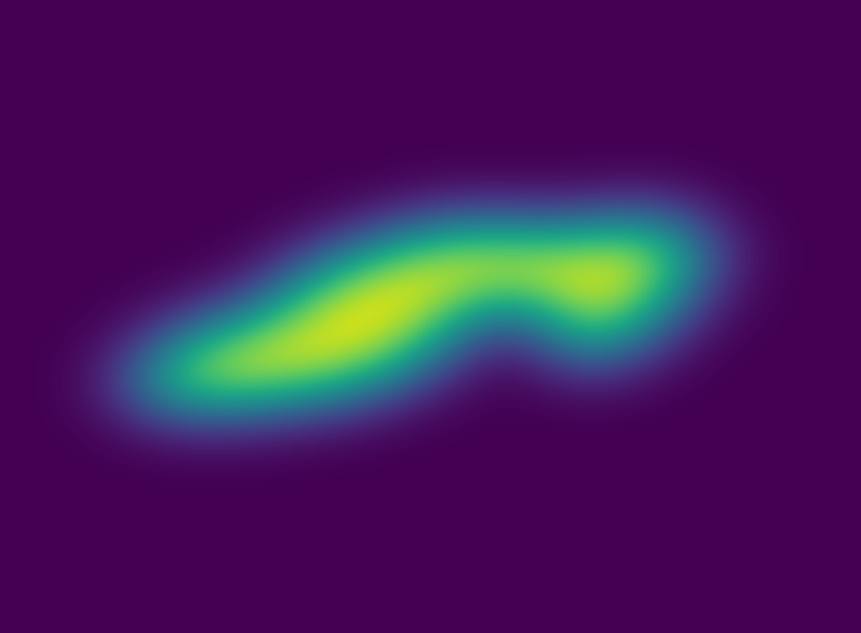};
    
    \nextgroupplot[xtick=\empty,ytick=\empty,yticklabel pos = right, xmin=0, xmax=1,ymin=0,ymax=1]
    \addplot graphics[xmin=0,xmax=1,ymin=0,ymax=1] {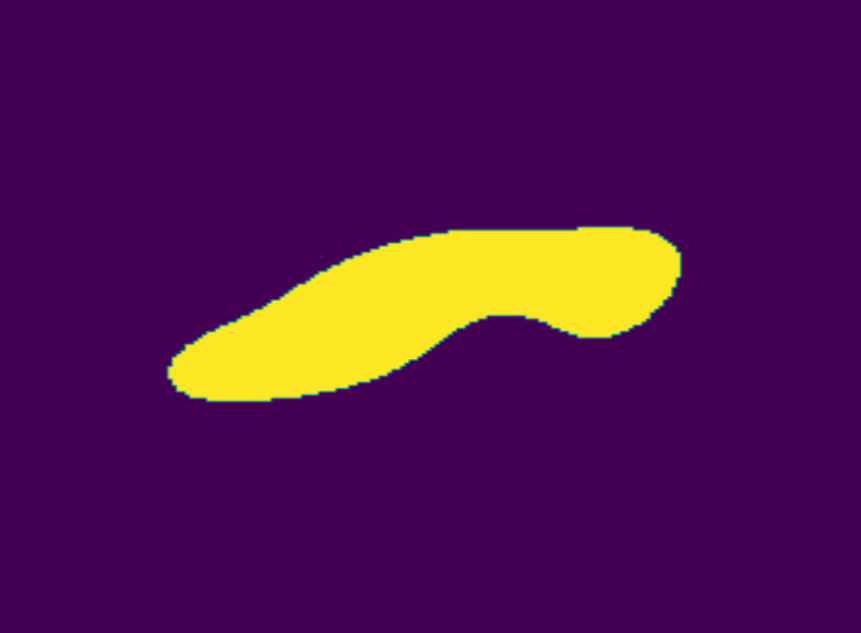};
    
    \nextgroupplot[xtick=\empty,ytick=\empty,yticklabel pos = right, xmin=0, xmax=1,ymin=0,ymax=1]
    \addplot graphics[xmin=0,xmax=1,ymin=0,ymax=1] {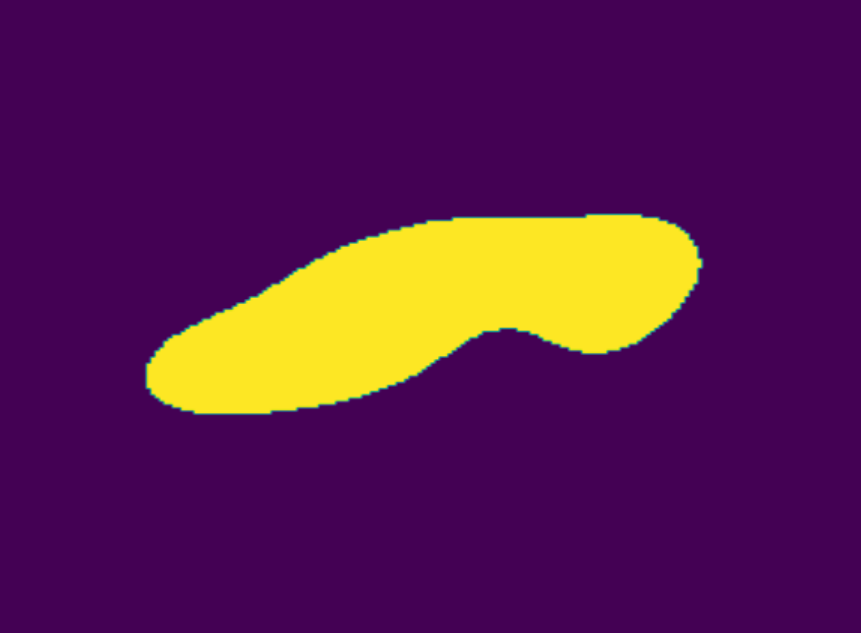};
    % %%%%%%%%%%%%%%%%%%%%
  \end{groupplot}
\end{tikzpicture}
  \caption{
  The leftmost are pixel-wise marginal probabilities (soft labels) for a noisy segmentation. 
  In the middle are segmentations obtained by the standard procedure of minimizing cross-entropy and taking the $1/2$-threshold.
  The rightmost are segmentations obtained by the standard procedure of minimizing soft-Dice and taking a $1/2$-threshold.
  That the volume of a segmentation obtained by $1/2$-thresholding a soft-Dice optimizer is always greater than or equal to that by $1/2$-thresholding cross-entropy optimizer is a consequence of the presented theory.
  }
  \label{burnerfig}
\end{figure} 

When the data is very unbalanced, that is, when foreground is much more common than background or vice versa, cross-entropy  often does not capture the priorities of medical practitioners working in the domain.
This is because the loss encourages improvements in a foreground voxel and a background voxel equally, whereas the preferences amongst practitioners are generally that it is more important to get the details right in the foreground and the region surrounding the foreground as compared to the background.
Since class imbalance is very common in the medical segmentation context, alternative  loss functions thought of as better suited for the situation have been proposed and evaluated.

One such loss function is based on the sigmoid smoothed version the S\"orensen-Dice coefficient, or Dice for short.
This loss function, which is often referred to as soft-Dice or simply the Dice loss, has with the exception of some reported unstable behaviour~\cite{jiang2019two,nordstrom2020calibrated} been shown to yield good performance in many experimental studies~\cite{bertels2019optimizing,eelbode2020optimization,drozdzal2016importance,jiang2019two,sudre2017generalised}.
As the result, it has together with cross-entropy become the most popular loss functions in the medical segmentation field and is used in popular frameworks such as the nnUNet~\cite{isensee2021nnu}.
Beside from soft-Dice, several other loss functions have also been proposed and investigated in the literature.
Examples include the Jaccard loss, the Tversky loss, the weighted cross entropy loss, the TopK loss, and the boundary loss \cite{ma2021loss}.
These losses are however much less commonly found in practice,
which is likely at least partially a consequence of the strong tradition of using the Dice metric for performance evaluation in the field.

Beside from the problem of  class imbalance,
another issue that effects performance of segmentation models is the presence of label noise~\cite{bridge2016intraobserver,nir2018automatic,armato2011lung,nyholm2018mr}.
Label noise is a consequence of the extremely tedious annotating process associated with image segmentation and especially, medical image segmentation, which is often in 3D, may have unclear boundaries, and require medical expertise.
Since label noise is 
pervasive 
in data used for training segmentation models, it is of high importance to understand how it influences the performance and in particular how it interacts with the chosen loss function.

Several works have been devoted to understanding the effect which label noise has on segmentation models trained with the soft-Dice loss.
These works are however almost exclusively of experimental nature and few general principles have been derived.
In the theoretical work that exists, it has been shown that minimizing soft-Dice in contrast to cross-entropy does not yield pixel-wise marginal probabilities of foreground and that the associated volume may be biased~\cite{bertels2021theoretical}.
It has further been conjectured, that soft-Dice is calibrated to Dice~\cite{nordstrom2020calibrated},
meaning that a sequence of soft segmentations converging to optimal soft-Dice when converted to hard segmentations by thresholding, also converges to optimal Dice.
These developments are important because they show that there is a large knowledge gap associated with the behaviour of soft-Dice under label noise, and essentially means that 
% a very large amount of papers 
numerous papers
have been devoted to training increasingly sophisticated models for image segmentation without 
% any 
detailed knowledge of what the models actually are trained to do.

In this paper a complete theoretical picture of the loss function under label noise is presented.
In particular, properties of the optimal solutions to soft-Dice, the volume bias of these solutions, and the calibration of soft-Dice to Dice are studied. 
The results confirm some experimental observations found in practice and also give new insights that have not previously been observed.

\paragraph{Contributions:}
Properties of soft-Dice under the presence of label noise are theoretically studied.
Optimal solutions are characterized, sharp bounds on the volume bias of the associated solutions are provided, and it is shown that soft-Dice is calibrated to Dice.
Finally, experimental results are provided for nine different noisy organ segmentation problems associated with the Gold Atlas project~\cite{nyholm2018mr} and nine noisy segmentation problems that are synthetically generated.

\section{Related work}
The presence of label noise in medical image segmentation is widely known and has been mentioned in many works.
Some examples include~\cite{bridge2016intraobserver, nir2018automatic,armato2011lung,nyholm2018mr}.

Several papers address the impact noise has on the soft-Dice loss.
In~\cite{bertels2021theoretical}, it was shown that optimizing soft-Dice under noisy label conditions in general does not lead to estimates of the pixel-wise probabilities of foreground.
Furthermore, the volume of the resulting estimators was also shown in general not to coincide with the mean volume of the noisy target.
The first issue was further investigated in~\cite{rousseau2021post}, where a method  was proposed for re-calibrating estimates obtained by soft-Dice.
The second issue of volume bias has been further elaborated in~\cite{popordanoska2021relationship}.
It has also been targeted using a different approach based on optimal transport theory~\cite{liu2022deep}.
Another way noise has been studied in the context of noisy labels and soft-Dice is by incorporating the noise as soft labels~\cite{gros2021softseg,kats2019soft}.
This is especially important for this work because it closely relates to the studied problems.

In binary classification, optimal threshold based classifiers for the $\mathrm{F}_1$ metric was discussed in  \cite{zhao2013beyond,lipton2014optimal} and further elaborated by others.
Because of the close relationship between the Dice metric in segmentation and the $\mathrm{F}_1$ metric in binary classification, this idea was later taken to the segmentation context and used to give a characterization of the optimal segmentations with respect to Dice~\cite{nordstrom2022image}.
Furthermore, sharp bounds of the volume of these solutions were provided.
Similar inspiration from work in binary classification \cite{bao2020calibrated,bartlett2006convexity} made authors propose that calibration can be a property of importance for explaining the good performance obtained by soft-Dice when evaluation is done by Dice~\cite{nordstrom2020calibrated}.

Relevant experimental work on noisy segmentation include \cite{heller2018imperfect} where the authors considered three different label noise models and observed that increased noise level caused worse performance.
A similar experimental study also including experiments on biased noise was later described in~\cite{vorontsov2021label}.
Finally, experimental work on soft labels with other loss functions than soft-Dice include
 ~\cite{silva2021using,lemay2022label,li2020superpixel}.
For recent general reviews on work on noisy segmentation and imperfect data, see~\cite{tajbakhsh2020embracing} and \cite{karimi2020deep}.

\section{Preliminaries}
In this work, the analysis will be performed over a continuous image domain rather than a discretized domain.
The reason for this is that some of the results are more clearly stated in this setting.
In general, the discretized setting can be seen as a special case of the continuous setting if appropriate step functions are considered.

\subsection{Notation}
Let $\Omega=[0,1]^n\subset \mathbb{R}^n$ be the unit cube of dimension $n\ge 1$ and $\lambda$ be the associated standard normalized Lebesgue measure such that $\lambda(\Omega)=1$.
Let $\mathcal{S}$ be the space of measurable functions from $\Omega$ to the binary numbers $\{0,1\}$, $\mathcal{M}$ be the space of measurable functions from $\Omega$ to the closed interval $[0,1]$ and $\mathcal{F}$ be the space of bounded measurable functions from $\Omega$ to  $(-\infty,\infty)$.
All of the  function spaces $\mathcal{S}$, $\mathcal{M}$ and $\mathcal{F}$ are equipped with their associated Borel $\sigma$-fields and the standard $L_1$-norm $\lVert \cdot \rVert_1$. 
The letters $s$,$m$,$c$ and $f$ are used to refer to specific type of objects.
The letter $s\in\mathcal{S}$ denotes a segmentation, where $s(\omega) = 1$ indicates foreground of the target structure of interest and $s(\omega) = 0$ indicates background.
The letter $m\in\mathcal{M}$ denotes a marginal probabilty function, that is, $m(\omega)\in[0,1]$ indicates the probability of a noisy label occupying site $\omega\in\Omega$.
% The letter $c\in\mathcal{M}$ denotes a soft label or soft segmentation, that is used for optimization.
The letter, $c\in \mathcal{M}$ denotes the soft segmentation that is obtained by optimizing the chosen loss function prior to any thresholding.
The letter $f\in\mathcal{F}$ denotes a logit function usually associated with a soft segmentation such that $\sigma \circ f(\omega)=c(\omega)$ for $\omega\in\Omega$ a.e., where $\sigma(x)=1/(1+e^{-x})$ is the standard sigmoid function.
The notation $\circ$ is used to denote composition of functions and $I_{A}(x)$ is used to denote the indicator function over the set $A$, that is $I_{A}(x) = 1$ if $x\in A$ and $I_{A}(x) = 0$ if $x\not\in A$.
This will in particular be used in conjunction with some element $c\in \mathcal{M}$ for thresholding a soft segmentation such as $I_{[t,1]}\circ c$ which assigns $1$ to all $\omega\in\Omega$ where $c(\omega) \ge t$ and $0$ to all $\omega\in\Omega$ where $c(\omega) < t$.

\subsection{Dice and soft-Dice}
When label noise is present, there are two different popular ways of defining Dice in the literature.
The first way is to take Dice with respect to the marginal probabilities or estimates thereof and is sometimes referred to as soft-labeling~\cite{kats2019soft,gros2021softseg,silva2021using,lemay2022label,li2020superpixel}.
The second way to define Dice is by taking the expected Dice score.
This turns up naturally in empirical risk minimization and is what is studied in e.g.~\cite{bertels2021theoretical}.
In this work the first convention is followed.
\begin{definition}
For any $m\in\mathcal{M}$, Dice is given by 
\begin{align}
    \mathrm{D}_m(s) &\doteq 
    \frac{2\int_\Omega s(\omega) m(\omega) \lambda(d\omega)}{\lVert s \rVert_1 + \lVert m \rVert_1},\quad s\in\mathcal{S}.
\end{align}
\end{definition}
Soft-Dice, is defined as one added by the negative Dice and the domain is generalized to the space of soft segmentations rather than hard segmentations.
Sometimes the constant is omitted, but since optimization is not affected by constants this does not affect any of the relevant behaviour.
\begin{definition}
For any $m\in\mathcal{M}$, soft-Dice is given by 
\begin{align}
    \mathrm{SD}_m(c) &\doteq 
    1-\frac{2\int_\Omega c(\omega) m(\omega) \lambda(d\omega)}{\lVert c \rVert_1 + \lVert m \rVert_1},\quad c\in\mathcal{M}.
\end{align}
\end{definition}

For a noisy segmentation $L$ taking values in $\mathcal{S}$ with $\mathbb{E}[L(\omega)] = m(\omega),\omega\in\Omega$, the corresponding definition of $\mathrm{D}_m(s),s\in\mathcal{S}$ using the expected Dice can be expressed as $\mathbb{E}[\mathrm{D}_L(s)], s\in\mathcal{S}$.
A brief discussion on the relationship between these definitions 
can be found in~\cite{nordstrom2022image}.
In short, the two definitions are equivalent if the variance of the volume of the noisy label is zero and they are very similar when the variance is small.
Since this is often the case in medical image segmentation, where noise often affects details on the boundary but do not significantly alter the volume, the two definitions will often approximate each other closely.
The same reasoning applies to soft-Dice, that is, in situations when the volume of the noisy labels is not significantly altered by the noise, it follows that $\mathbb{E}[\mathrm{SD}_L(c)] \approx \mathrm{SD}_m(c)$, $c\in\mathcal{M}$.

\subsection{Problem description}
% In this work, two different maximization problems are studied.
When Dice is used as the target evaluation metric, the problem of image segmentation can be seen as finding a segmentation that maximizes Dice, that is
\begin{align}
    \sup_{s\in\mathcal{S}} \mathrm{D}_m(s), \quad m\in\mathcal{M}.
    \label{eq:sq}
\end{align}
Because of the discrete nature of this problem, solving this direct by optimization is in general not feasible.
Instead a loss function, in this case soft-Dice, is minimized over the space of soft segmentations
% One way to do this is to look at the continuous version of the problem 
\begin{align}
    \inf_{c\in\mathcal{M}} \mathrm{SD}_m(c), \quad m\in\mathcal{M},
    \label{eq:sds}
\end{align}
and to avoid inequality bounds, the sigmoid function $\sigma(x)=1/(1+\exp(-x))$ is used
% One of the most popular methods in the medical image segmentation community is to approximate the hard labels with soft-labels as follows
\begin{align}
    \inf_{f\in\mathcal{F}} \mathrm{SD}_m(\sigma\circ f), \quad m\in\mathcal{M}.
\end{align}
This is the form of soft-Dice most commonly presented in the literature.
% In practice $\mathcal{F}$ is usually approximated with the functions expressible with some neural network architecture with elements $f\in\mathcal{F}$ corresponding to the associated functions expressible with some set of associated parameters.
% can be thought of as the a particular set of parameters associated with the neural network.

Once a solution $f \in \mathcal{F}$ is found, it needs to be converted to a binary segmentation, in other words, an element $s\in \mathcal{S}$.
The most common procedure for doing this is by processing the solution with a $1/2$ threshold, that is, to generate
\begin{align}
    s = I_{[1/2,1]} \circ \sigma \circ f.
\end{align}

For the cross-entropy loss, the fact that the optimizers uniquely are given by the marginal estimates and that the loss is calibrated to accuracy is well known~\cite{nordstrom2022image}.
Corresponding results connecting the minimizing of soft-Dice with the maximizing of Dice has to the best of our knowledge not been studied in the literature, and deriving such results is the main objective of this work.

\section{Main results}
In this section the main results are presented.
Associated proofs are found in the Supplemental Material.
The object of the theoretical investigation is two-fold.
Firstly, a detailed picture of optimal solutions to soft-Dice is sought for.
Secondly, assurance for using soft-Dice as a proxy when maximizing Dice is sought for.
To this end, several theoretical results are provided.

\begin{theorem}
For any $m\in\mathcal{M}$, the class $\mathcal{M}^*_m \subset \mathcal{M}$ of elements attaining the infimum $\inf_{c\in\mathcal{M}}\mathrm{SD}_m(c)$ are given by
\begin{align}
    c(\omega) \in
    \begin{cases}
        \{0\} & \text{ if } m(\omega) < \sup_{s'\in\mathcal{S}} \mathrm{D}_m(s')/2,\\  
        [0,1] & \text{ if } m(\omega) = \sup_{s'\in\mathcal{S}} \mathrm{D}_m(s')/2,\\
        \{1\} &\text{ if } m(\omega) > \sup_{s'\in\mathcal{S}} \mathrm{D}_m(s')/2,
    \end{cases}
\end{align}
for $\omega\in\Omega$, $\lambda$-a.e.
\label{theorem1}
\end{theorem}
In~Theorem \ref{theorem1}, solutions that minimize soft-Dice are characterized.
Note that with the exception of some potential corner cases, the set of optimal solutions assign $0$ or $1$ to almost all of the domain.
This implies, that exact solutions in theory can never be reached when using sigmoids as in~\eqref{eq:sds}.
It is, however, possible to get arbitrarily close.

\begin{theorem}
For any $m\in\mathcal{M}$, the class $\mathcal{M}^*_m \subset \mathcal{M}$ of elements attaining the infimum $\inf_{c\in\mathcal{M}}\mathrm{SD}_m(c)$ satisfy the following bounds
\begin{align}
[\inf_{c\in\mathcal{M}_m^*}\lVert  c\rVert_1,\sup_{c\in\mathcal{M}_m^*} \lVert c \rVert_1] \subseteq [\lVert m \rVert_1^2,1].
\end{align}
Moreover, the bounds are sharp in the sense that there for any $v\in(0,1]$ exist $m_0,m_1\in\mathcal{M}$ such that $\lVert m_0 \rVert_1 = \lVert m_1 \rVert_1 = v$ and
\begin{align}
\inf_{c\in\mathcal{M}_{m_0}^*} \lVert c \rVert_1 = \lVert m_0 \rVert_1^2, \quad \sup_{c\in\mathcal{M}_{m_1}^*}\lVert c \rVert_1 =1.
\end{align}
\label{theorem2}
\end{theorem}
In Theorem~\ref{theorem2}, sharp bounds for the volume bias associated with soft-Dice are presented.
Firstly, this means that bounds for the volume bias are described.
Secondly, this means that that there are situations when the extreme cases are obtained, and consequently, that there exist no better bounds.
\begin{theorem}
For any $m\in\mathcal{M}$, threshold $a\in(0,1)$ and relatively compact sequence $\{c_l\}\subset \mathcal{M}$, it follows that if
\begin{align}
\lim_{l\rightarrow\infty}\mathrm{SD}_m(c_l) = \inf_{c\in\mathcal{M}}\mathrm{SD}_m(c)
\end{align}
then
\begin{align}
\lim_{l\rightarrow \infty}\mathrm{D}_m(I_{[a,1]}\circ c_l)= \sup_{s\in\mathcal{S}}\mathrm{D}_m(s).
\end{align}
   \label{theorem3}
\end{theorem}
In Theorem~\ref{theorem3}, a \emph{calibration result}, connecting soft-Dice with Dice is presented.
Informally, the result says that a sequence of soft segmentations $\{c_l\}_{l\ge 1}$ that converges to optimal soft-Dice converges to optimal Dice when thresholded appropriately by some constant $a\in(0,1)$.
This property is important because it gives assurance that training a model with respect to soft-Dice is motivated if the target is to maximize Dice.
Note that the results also hold when using sigmoid functions to bound the interval.
Furthermore, the condition of relative compactness is only necessary when dealing with continuous domains as any subset of functions associated with a finite voxelization is always relatively compact.

\section{Implications}

The proofs are partially built on work presented in~\cite{nordstrom2022image}, where the analysis of the optimal solutions to Dice over the space $\mathcal{S}$, that is $\sup_{s\in\mathcal{S}}\mathrm{D}_m(s)$, is studied.
The authors show that the class of optimal segmentations to this problem 
$\mathcal{S}^*_m \subset \mathcal{S}$ is given by $s\in\mathcal{S}$
such that
\begin{align}
    s(\omega) \in
    \begin{cases}
        \{ 0 \} &\text{ if } m(\omega) < \sup_{s'\in\mathcal{S}} \mathrm{D}_m(s')/2, \\
        \{ 0,1 \} &\text{ if } m(\omega)= \sup_{s'\in\mathcal{S}} \mathrm{D}_m(s')/2, \\
        \{ 1 \} &\text{ if } m(\omega) > \sup_{s'\in\mathcal{S}} \mathrm{D}_m(s')/2,
    \end{cases}
    \label{diceopt}
\end{align}
for $\omega\in\Omega$, $\lambda$-a.e.
Note that the optimal solutions are almost exactly the same as $\mathcal{M}_m^*$ presented in Theorem~\ref{theorem1}.
The only difference is that the domain with marginal $m(\omega) = \sup_{s\in\mathcal{S}} \mathrm{D}_m(s)/2, \omega\in\Omega$ can take values over the whole interval $[0,1]$ instead of the binary values $\{0,1\}$.

\begin{figure}[t!]
  \centering

\begin{tikzpicture}
  \begin{axis}[height=7cm,width=9cm,xmin=-1,xmax=1,ymin=0,ymax=1,xtick=\empty,ytick={0,0.25,0.5,0.75,1.0},xlabel=$c\in\mathcal{C}$,legend columns=-1, legend cell align={left},
  legend style={at={(0.5,1.1)},column sep=0.1cm,anchor=south,legend cell align=left}
  ]
  
    \addplot[color=blue,dashed,samples=1000] {max(1-exp(-3*x^2)*0.9,0.25)};
    \addlegendentry{$\mathrm{SD}_m(c)$}
    
    \addplot[no marks,color=blue] coordinates {(-0.2,0.75) (0.2,0.75)};
    \addlegendentry{$\mathrm{D}_m(I_{[a,1]}\circ c)$}

    \addplot[no marks,color=blue] coordinates {(-0.6,0.5) (-0.2,0.5)};
    \addplot[no marks,color=blue] coordinates {(-1.0,0.25) (-0.6,0.25)};
    
    \addplot[no marks,color=blue] coordinates {(0.2,0.5) (0.6,0.5)};
    \addplot[no marks,color=blue] coordinates {(0.6,0.25) (1.0,0.25)};

    \addplot[only marks,mark=*,color=blue,fill=blue] coordinates{(-0.2,0.75)};
    \addplot[only marks,mark=*,color=blue,fill=blue] coordinates{(0.2,0.75)};
    
    \addplot[only marks,mark=*,color=blue,fill=blue] coordinates{(-0.6,0.5)};
    \addplot[only marks,mark=*,color=blue,fill=white] coordinates{(-0.2,0.5)};
    \addplot[only marks,mark=*,color=blue,fill=white] coordinates{(-0.6,0.25)};
    
    \addplot[only marks,mark=*,color=blue,fill=blue] coordinates{(0.6,0.5)};
    \addplot[only marks,mark=*,color=blue,fill=white] coordinates{(0.2,0.5)};
    \addplot[only marks,mark=*,color=blue,fill=white] coordinates{(0.6,0.25)};
    
  \end{axis}
\end{tikzpicture}

\caption{
  Illustration of soft-Dice (dashed line) with respect to a soft prediction $c\in\mathcal{M}$ for some $m\in\mathcal{M}$ and Dice (solid line) with respect to the associated thresholded solutions $I_{[a,1]}\circ c$, for some $a\in (0,1)$.
  The x-axis is a schematic illustration of a function space, it is \emph{not} one-dimensional.
  }
  \label{calfig}
\end{figure}
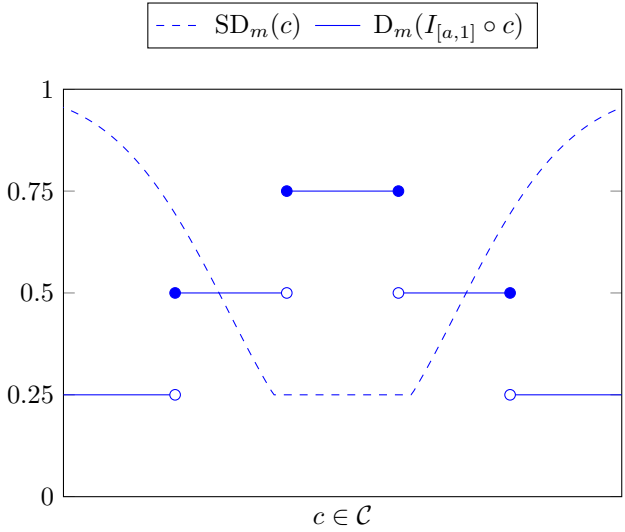

An obvious consequence of this is that if $c\in\mathcal{M}_m^*$, then $I_{[a,1]}\circ c\in\mathcal{S}_m^*$ for any $a\in(0,1)$.
Or informally, that thresholded solutions of optimizers to soft-Dice are optimizers to Dice and therefore carry the associated properties.
A schematic illustration of this relationship between soft-Dice and Dice is depicted in Figure~\ref{calfig}.

Another important related property that follows from this connection to~\cite{nordstrom2022image} is that the volume of thresholded minimizers to soft-Dice always is greater than or equal to the volume of thresholded minimizers to cross-entropy.
Since $I_{[1/2,1]}\circ m$ is one such maximizer, it follows that
\begin{align}
    \inf_{s\in\mathcal{S}_m^*}\lVert s \rVert_1 \ge \lVert I_{[1/2,1]}\circ m \rVert_1.
\end{align}
This property theoretically motivates the illustration in Figure~\ref{burnerfig} and implies that we can never get a reversed situation where the volume of a $1/2$-thresholded minimizer to cross-entropy solution is greater than the volume of a $1/2$-thresholded maximizer of soft-Dice.
Before transforming the soft segmentations to hard segmentations with thresholding, however, the optimizer to cross-entropy will always have volume $\lVert m\rVert_1$ and 
by Theorem~\ref{theorem2}, there exist situations when the optimizers to soft-Dice have volume $\lVert m \rVert_1^2$.
Recall that since $\lambda(\Omega) = 1$, it follows that $\lVert m \rVert_1 \le 1$ and consequently that $\lVert m\rVert_1^2 \le \lVert m\rVert_1$.

\section{Experiments}
To verify that the theoretical results are relevant in practice, a series of experiments are conducted.
In particular, it is investigated if the results obtained by optimizing soft-Dice
\begin{align}
    \inf_{f\in\mathcal{F}} \mathrm{SD}_m(\sigma\circ f),
\end{align}
in various situations with a simple optimization scheme lead to solutions that are close to those described in Theorem~\ref{theorem1}.
Since it is unlikely that a significant part of the image domain has marginal value exactly equal to $\sup_{s\in\mathcal{S}} \mathrm{D}_m(s)/2$, or equivalently, that
\begin{align}
 \lambda(\{\omega\in \Omega: m(\omega)=\sup_{s\in\mathcal{S}}\mathrm{D}_m(s)/2 \}) > 0,
\end{align}
the task is simplified to only compare numerical solutions to the single segmentation given by
\begin{align}
    s(\omega) = I_{[\sup_{s'\in\mathcal{S}} \mathrm{D}_m(s')/2,1] }\circ m(\omega), \quad \omega \in \Omega.
\end{align}
This segmentation is easy to compute efficiently for the numerical cases in the experiments.

\begin{table*}[t!]
\centering
% \begin{tabular}{ccccccccc}
% \begin{tabular}{lllll}
\begin{tabular}{@{}lc@{ \hspace{0.5cm} }c@{ \hspace{0.1cm}}c@{ \hspace{0.5cm} }c@{ \hspace{0.1cm} }c@{ \hspace{0.5cm} }c@{ \hspace{0.1cm} }c@{ \hspace{0.5cm} }c@{ \hspace{0.1cm} }c@{\hspace{0.5cm} }c@{ \hspace{0.1cm} }c@{}}
% \begin{tabular}{lcccccccccc}
\toprule
ROI & samples &  $\bar e_{0,1}$ & $\bar e_{1,1}$ & $\bar e_{0,10}$ & $\bar e_{1,10}$ & $\bar e_{0,20}$ & $\bar e_{1,20}$ & $\bar e_{0,100}$ & $\bar e_{1,100}$ & $\bar e_{0,200}$ & $\bar e_{1,200}$ \\
\midrule
(G) Urinary bladder & 19
 & 0.500 & 0.500 & 0.004 & 0.002 & 0.001 & 0.000 & 0.000 & 0.000 & 0.000 & 0.000 \\
(G) Rectum & 19
 & 0.500 & 0.500 & 0.040 & 0.011 & 0.003 & 0.002 & 0.000 & 0.000 & 0.000 & 0.000 \\
(G) Anal canal & 19
 & 0.500 & 0.500 & 0.408 & 0.307 & 0.236 & 0.087 & 0.000 & 0.000 & 0.000 & 0.000 \\
(G) Penile bulb & 19
 & 0.500 & 0.500 & 0.466 & 0.425 & 0.417 & 0.329 & 0.044 & 0.014 & 0.000 & 0.000 \\
(G) Neurovascular b. & 19
 & 0.500 & 0.500 & 0.390 & 0.286 & 0.210 & 0.095 & 0.000 & 0.000 & 0.000 & 0.000 \\
(G) Femoral head R & 19
 & 0.500 & 0.500 & 0.028 & 0.005 & 0.001 & 0.000 & 0.000 & 0.000 & 0.000 & 0.000 \\
(G) Femoral head L & 19
 & 0.500 & 0.500 & 0.031 & 0.006 & 0.001 & 0.000 & 0.000 & 0.000 & 0.000 & 0.000 \\
(G) Prostate & 19
 & 0.500 & 0.500 & 0.166 & 0.060 & 0.013 & 0.004 & 0.000 & 0.000 & 0.000 & 0.000 \\
(G) Seminal vesicles & 19
 & 0.500 & 0.500 & 0.413 & 0.322 & 0.257 & 0.136 & 0.000 & 0.000 & 0.000 & 0.000 \\
\midrule
(S) $\rho=0.01$ & 1000
 & 0.500 & 0.500 & 0.022 & 0.008 & 0.001 & 0.001 & 0.000 & 0.000 & 0.000 & 0.000 \\
(S) $\rho=0.02$ & 1000
 & 0.500 & 0.500 & 0.026 & 0.011 & 0.002 & 0.002 & 0.000 & 0.000 & 0.000 & 0.000 \\
(S) $\rho=0.03$ & 1000
 & 0.500 & 0.500 & 0.027 & 0.015 & 0.004 & 0.003 & 0.001 & 0.000 & 0.000 & 0.000 \\
(S) $\rho=0.04$ & 1000
 & 0.500 & 0.500 & 0.034 & 0.021 & 0.005 & 0.005 & 0.001 & 0.001 & 0.000 & 0.000 \\
(S) $\rho=0.05$ & 1000
 & 0.500 & 0.500 & 0.039 & 0.025 & 0.007 & 0.006 & 0.001 & 0.001 & 0.000 & 0.000 \\
(S) $\rho=0.06$ & 1000
 & 0.500 & 0.500 & 0.051 & 0.034 & 0.010 & 0.009 & 0.001 & 0.001 & 0.001 & 0.001 \\
(S) $\rho=0.07$ & 1000
 & 0.500 & 0.500 & 0.057 & 0.040 & 0.013 & 0.011 & 0.002 & 0.001 & 0.001 & 0.001 \\
(S) $\rho=0.08$ & 1000
 & 0.500 & 0.500 & 0.067 & 0.048 & 0.017 & 0.015 & 0.002 & 0.002 & 0.001 & 0.001 \\
(S) $\rho=0.09$ & 1000
 & 0.500 & 0.500 & 0.080 & 0.057 & 0.022 & 0.019 & 0.003 & 0.003 & 0.001 & 0.001 \\
\bottomrule
\end{tabular}
\vspace{0.2cm}
\caption{
Results of experiments with respect to the  pelvic data in the the Gold Atlas project (G) and the
synthetic data (S).
The columns show the name of the ROI, the number of samples used for computation and the average absolute differences $\bar e_{0,l}$ with respect to the soft labels $e_{0,l} = \lVert \sigma\circ f_l - I_{[\sup_{s'\in\mathcal{S}}\mathrm{D}_m(s')/2,1]} \circ m\rVert_1$ and the average absolute difference $\bar e_{1,l}$ with respect to the hard labels
$e_{1,l} = \lVert I_{[1/2,1]}\circ \sigma\circ f_l - I_{[\sup_{s'\in\mathcal{S}}\mathrm{D}_m(s')/2,1]} \circ m\rVert_1$
for the gradient descent iterations $l=1,10,20,100,200$.
}

\label{results_table}
\end{table*}

% \FloatBarrier
\subsection{Setup}
The experiments are conducted with respect to voxelizations of the image domain into $N\ge 1$ voxels with equal volume.
That is, $\Omega$ is partitioned into $\{\Omega_i\}_{1\le i\le N}$ of such that $\Omega = \Omega_1\cup\dots\cup\Omega_N$ with $\Omega_i \cap\Omega_j=\emptyset$ when $i\not=j$ and $\lambda(\Omega_i) = \lambda(\Omega_j)$ for all $ 1\le i,j \le N$.
Note that different images in the data sets may have a different number of voxels $N$.
For a particular marginal function $m$, an initial element $f_1\in\mathcal{F}$  is initialized so that each of the voxel values is assigned the values of an independently drawn isotropic Gaussian random variable. 
A sequence of updates is then generated by a simple normalized gradient descent scheme with a fixed learning rate $\gamma=10N$.
The learning rate will affect the convergence rate of the optimization. 
However, since the main object of interest is the solution to which the method converge, this is not a concern as long as the methods converge reasonably well during the the steps that we run.
Consequently, the learning rate was chosen after some minor fine-tuning.

To measure how close the computed solution is to the theoretical solution, two metrics are considered.
The first metric is defined as
\begin{align}
   e_{0,l} = \lVert \sigma\circ f_l - I_{[\sup_{s'\in\mathcal{S}}\mathrm{D}_m(s')/2,1]} \circ m\rVert_1
\end{align}
and measures the average voxel-wise distance from the soft values and the theoretical value.
The second metric is defined as
\begin{align}
   e_{1,l} = \lVert I_{[1/2,1]}\circ\sigma\circ f_l - I_{[\sup_{s'\in\mathcal{S}}\mathrm{D}_m(s')/2,1]} \circ m\rVert_1
\end{align}
and measures the same average voxel-wise distance but with values that are processed with a $1/2$-threshold.
These two metrics are reported after running $l=1,10,20,100,200$ iterations.
Details on the experiment and code are available in the Supplementary Material.

\subsection{Data}
The experiments are conducted on two different data sets.
One real world (G) and one synthetically generated (S).
Instructions on how to access and process the data are available in the Supplementary Material.

\paragraph{(G):}
The first data set contains nine structures in the pelvic area and is part of the Gold Atlas project~\cite{nyholm2018mr}.
It %data 
includes $19$ cases, is in 3D, and slices are of resolution $512\times 512$ pixels.
Each of the structures has been delineated by five clinical experts and  marginals are formed by taking the pixel-wise average of the various labels.
Patches centered in the each of the structures of size $128\times 128$ on the original resolution are extracted.
To extract the data the \emph{Plastimatch}~\cite{sharp2010plastimatch} software is used.

\paragraph{(S):}
The second data set is synthetic and generated by three steps.
Firstly, a ball with radius $0.2$ is constructed on a discretized domain in 2D containing $200\times 200$ pixels.
Secondly, this ball is convolved with a Gaussian filter with variance $\rho$ for a set of different values.
Each value corresponds to a particular noise level, where higher $\rho$ is higher noise.
Thirdly, the image is deformed by a random Gaussian field to get different shapes.
The idea is that the larger $\rho$, the more noise is present in the image.

\subsection{Results}
In Table~\ref{results_table} the results of the experiments are listed.
This includes, for each region of interest, the average errors as measured by $e_{0,l}$ and $e_{1,l}$ for various iterations $l=1,10,20,100,200$.
It is clear that the optimization scheme on average converges to the optimizers described in Theorem~\ref{theorem1}.
There also seems to be some indication that the amount of noise as measured by $\rho$ has some effect on the rate of convergence.
That is, higher values of $\rho$ seem to cause the convergence to be a bit slower.
Figure~\ref{seqGfig} in Appendix A illustrates different samples from the set of experiments (G) together with $\sigma\circ f_l$ for various iterations and the theoretically optimal solution $s$.
Similarly, Figure~\ref{seqSfig} in Appendix B illustrates different samples from the set of experiments (S) together with $\sigma\circ f_l$ for various iterations $l$ and the theoretically optimal solution $s$.

\section{Conclusion}
In this work the optimal solutions 
have been studied
with respect to soft-Dice under when label noise is incorporated using soft labels. 
%has been studied.
The optimal solutions have been characterized and sharp bounds on the volume bias for these solutions have been provided.
Furthermore, it has been shown that soft-Dice is calibrated to Dice, in the sense that any sequence that converges to optimal soft-Dice, when thresholded appropriately, converges to optimal Dice.
Finally, the relevance of the characterization in practice has been illustrated 
% by an experimental study 
through experiments 
on one synthetic data set and data from the Gold Atlas project~\cite{nyholm2018mr}.
All the results support the theoretical results.
The results presented give important insight into how noise affects one of the most common training setups used in medical image segmentation.

\paragraph{Limitations:}
(1) In a numerical setting it might not be possible to exactly construct the cases necessary for the sharpness part in Theorem~\ref{theorem2}.
The numerical error is however of size $1/N$ where $N$ is the number of voxels which is usually large in practice.
(2) The experiments are with respect to an unconstrained function space rather than a space constrained by some particular neural network architecture.
This is purposely done in order to isolate the behaviour of the objective as much as possible.
% from external influences from a particular neural network model.
% To see if the same behaviour is present for a particular artificial neural network, experiments with that particular network can be conducted.

% \paragraph{Acknowledgement:}
% The work was funded by RaySearch Laboratories AB.

\begin{appendices}

\begin{figure*}[htb!]
\section{Samples from experiments (G)}
  \centering
\begin{tikzpicture}
  \begin{groupplot}[group style={group size=7 by 9, horizontal sep=0.05cm, vertical sep=0.05cm},height=3.70cm,width=3.70cm,xmin=0,xmax=1,ymin=0,ymax=1, xticklabels=\empty, yticklabels=\empty]
  
    \nextgroupplot[title={$m$}, ylabel={Urinary b.}]
    \addplot graphics[xmin=0,xmax=1,ymin=0,ymax=1] {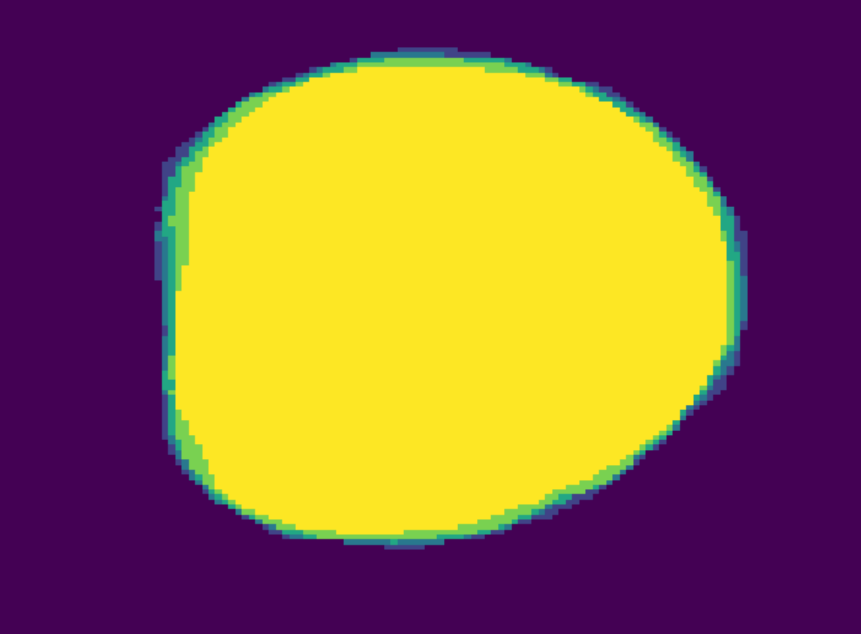};
    
    \nextgroupplot[title={$\sigma\circ f_1$}]
    \addplot graphics[xmin=0,xmax=1,ymin=0,ymax=1] {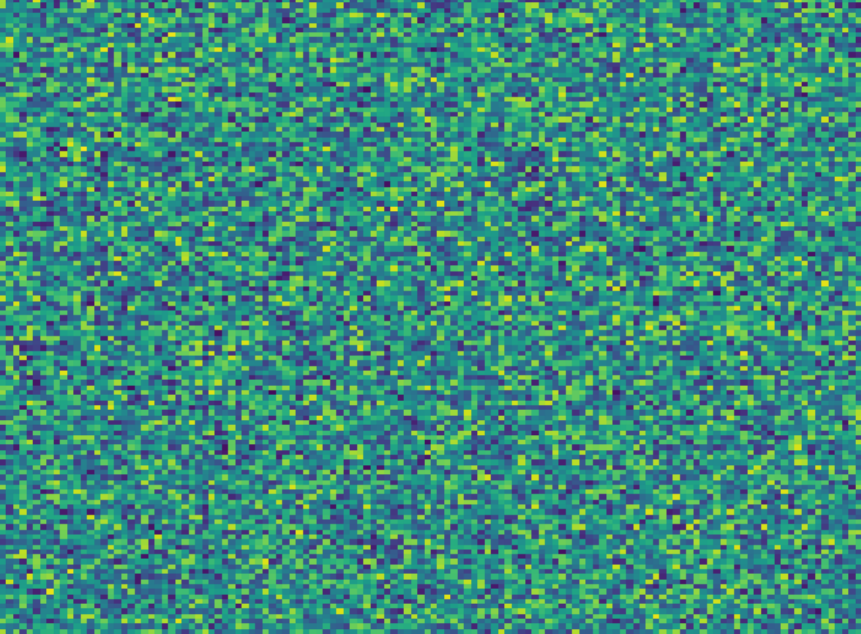};
    
    \nextgroupplot[title=$\sigma\circ f_{10}$]
    \addplot graphics[xmin=0,xmax=1,ymin=0,ymax=1] {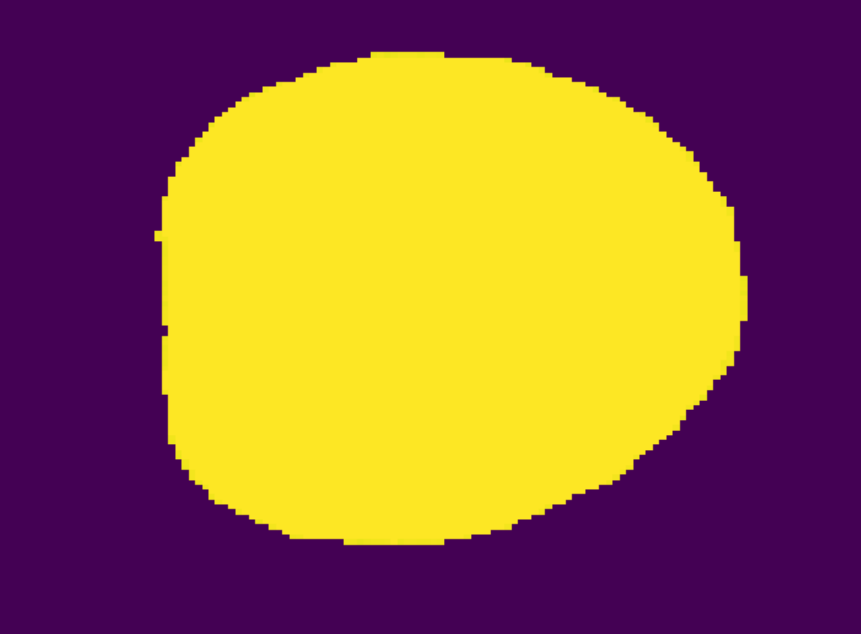};
  
    \nextgroupplot[title=$\sigma\circ f_{20}$]
    \addplot graphics[xmin=0,xmax=1,ymin=0,ymax=1] {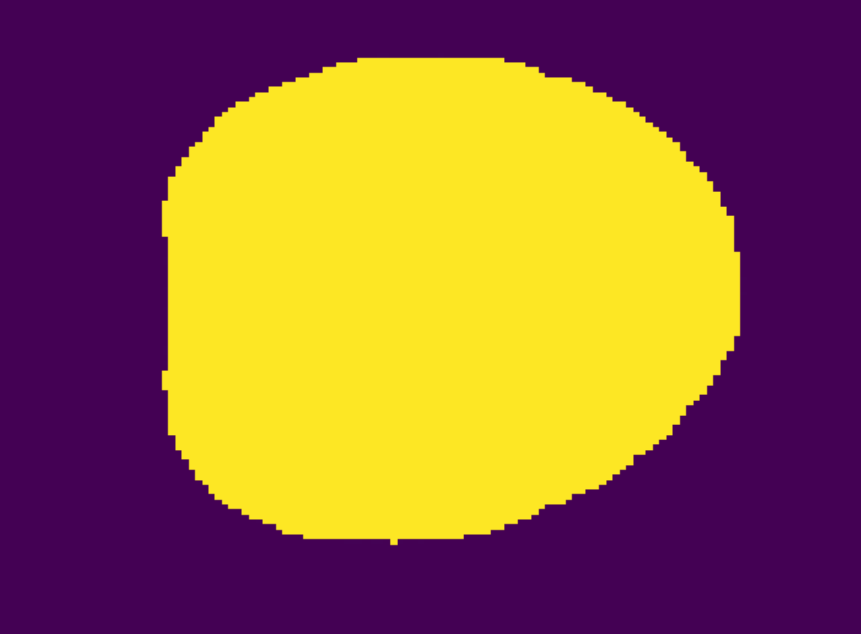};
 
    \nextgroupplot[title=$\sigma\circ f_{100}$]
    \addplot graphics[xmin=0,xmax=1,ymin=0,ymax=1] {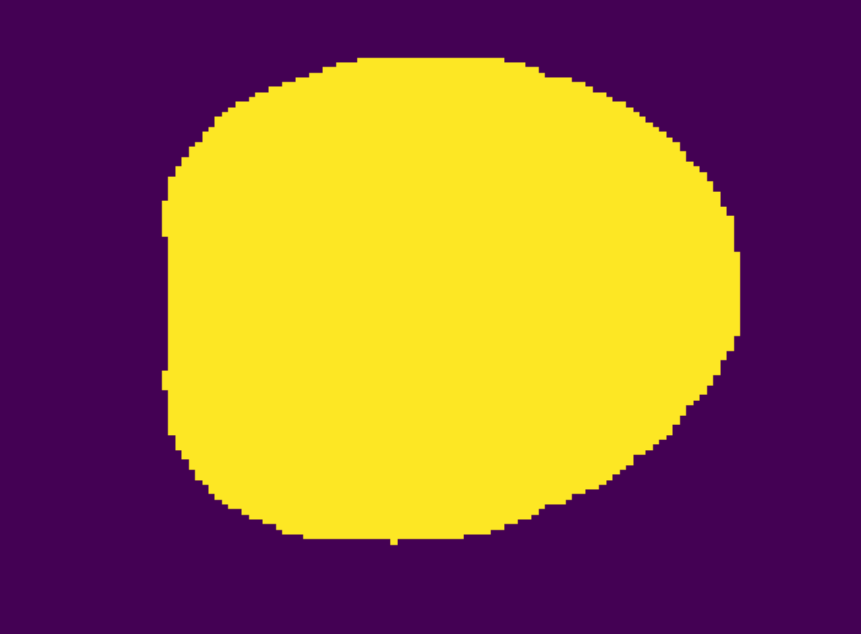};
    
    \nextgroupplot[title=$\sigma\circ f_{200}$]
    \addplot graphics[xmin=0,xmax=1,ymin=0,ymax=1] {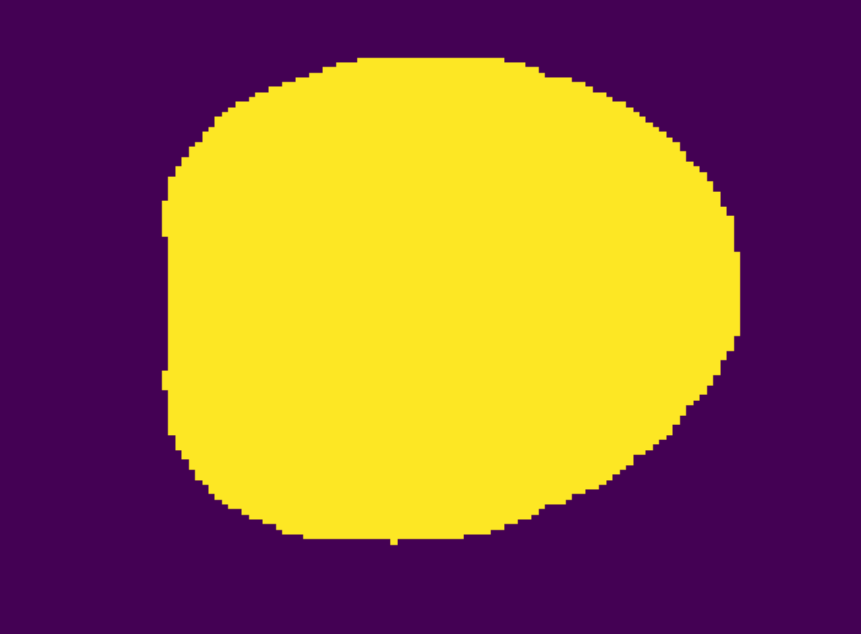};

    \nextgroupplot[title=$s$]
    \addplot graphics[xmin=0,xmax=1,ymin=0,ymax=1] {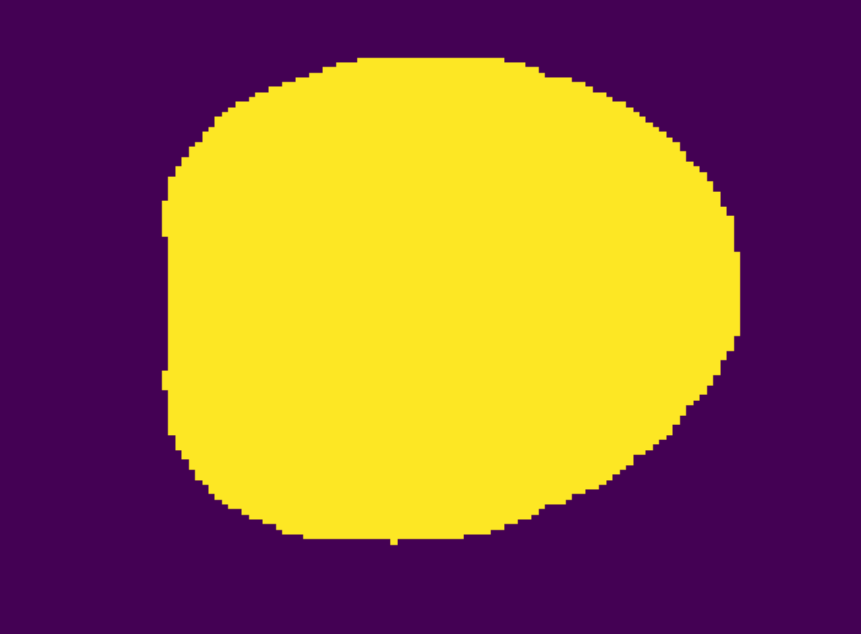};
    %%%%%%%%%%%%%%%%%%
    
    \nextgroupplot[ylabel={Rectum} ]
    \addplot graphics[xmin=0,xmax=1,ymin=0,ymax=1] {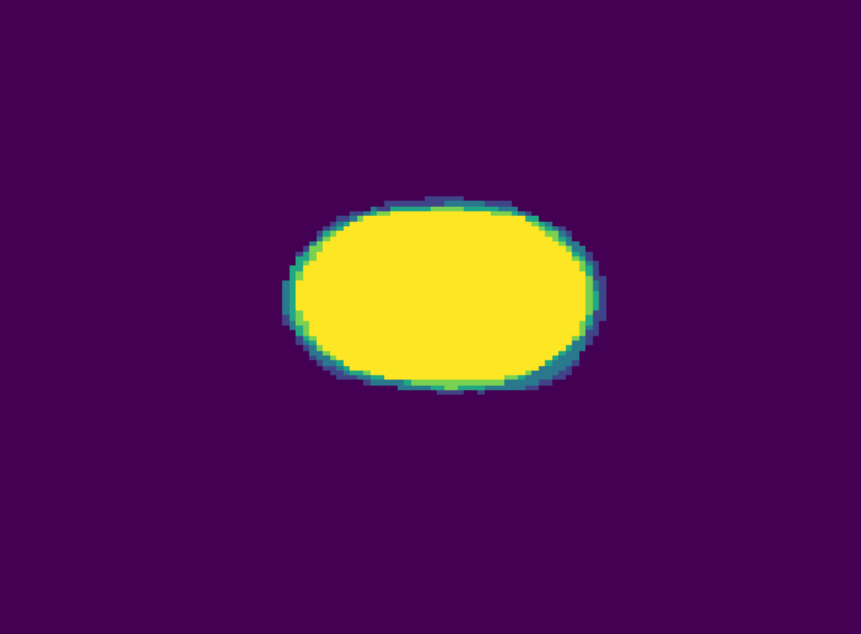};
    
    \nextgroupplot
    \addplot graphics[xmin=0,xmax=1,ymin=0,ymax=1] {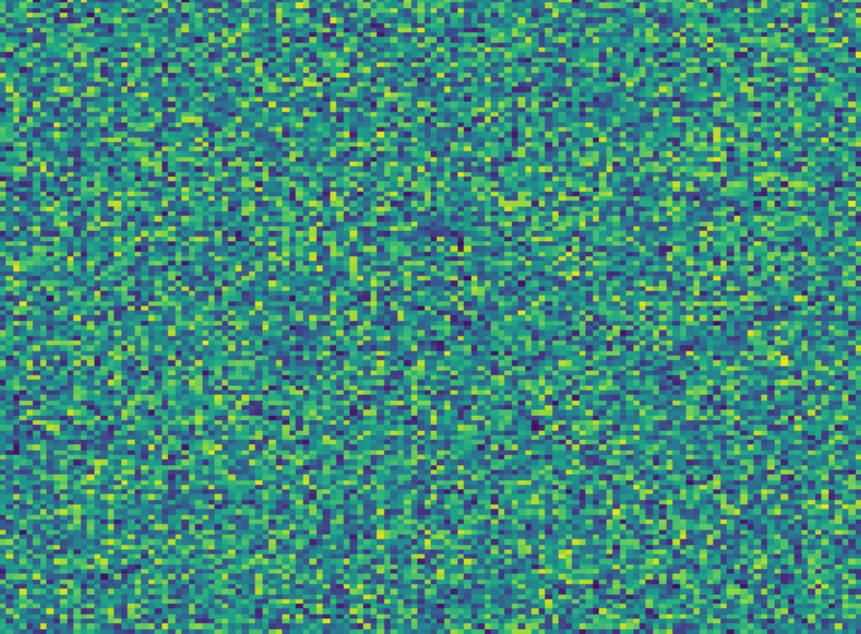};
    
    \nextgroupplot
    \addplot graphics[xmin=0,xmax=1,ymin=0,ymax=1] {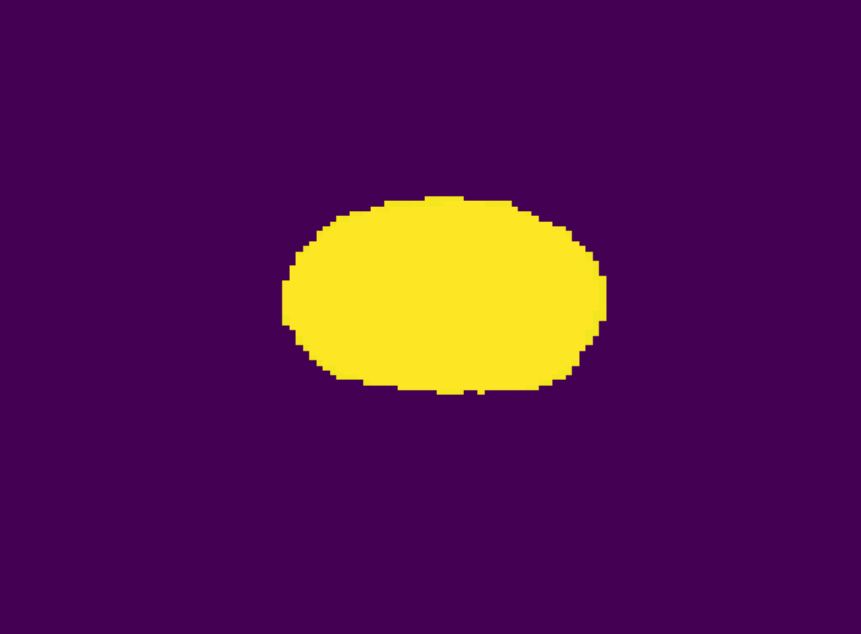};
  
    \nextgroupplot
    \addplot graphics[xmin=0,xmax=1,ymin=0,ymax=1] {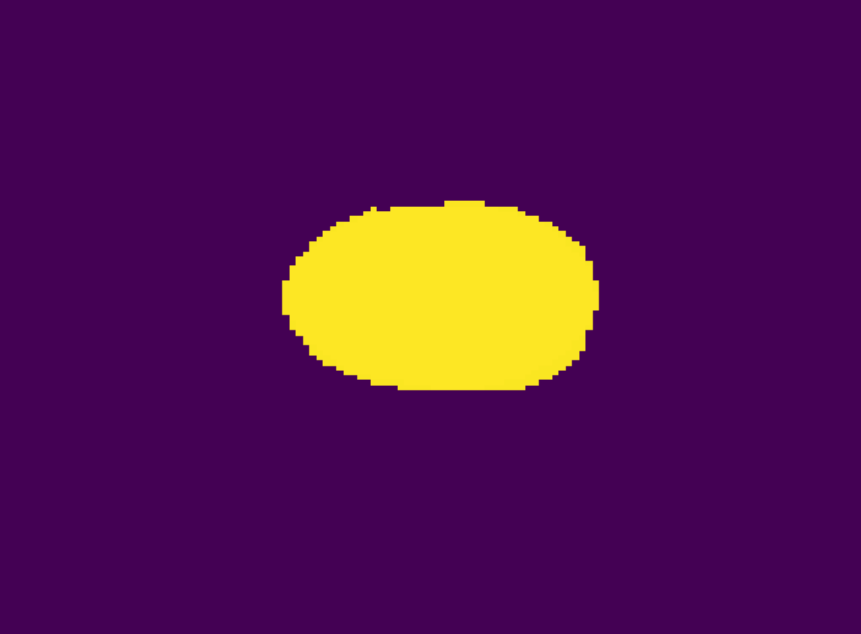};
 
    \nextgroupplot
    \addplot graphics[xmin=0,xmax=1,ymin=0,ymax=1] {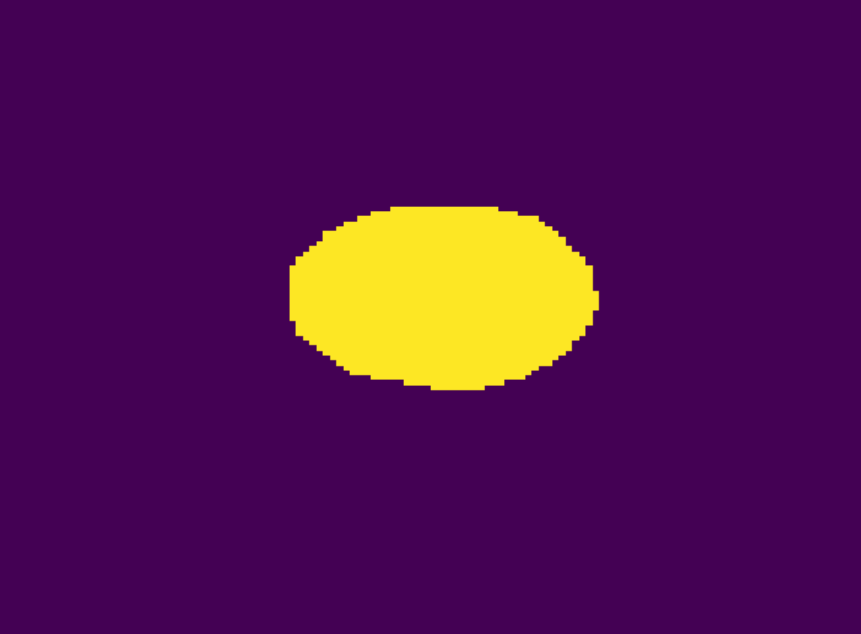};
    
    \nextgroupplot
    \addplot graphics[xmin=0,xmax=1,ymin=0,ymax=1] {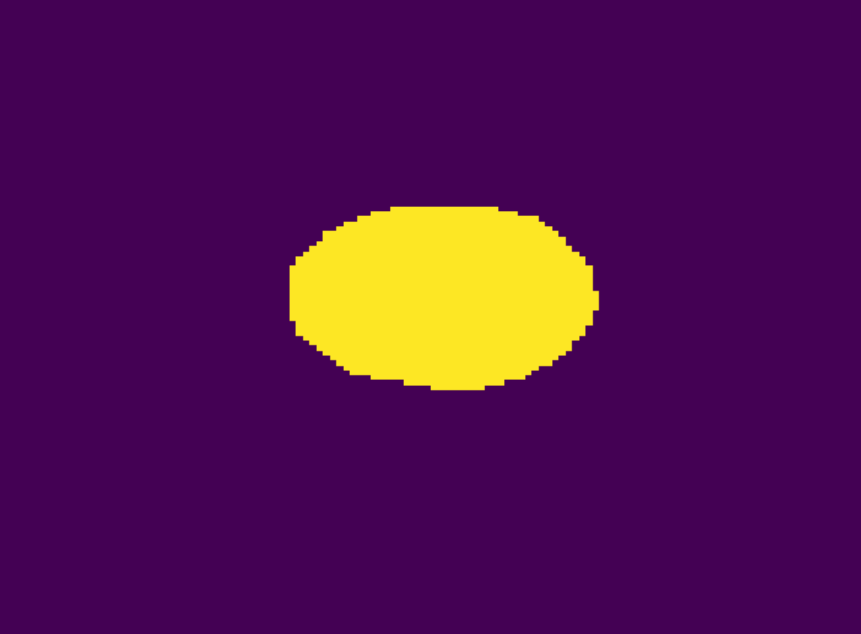};
    
    \nextgroupplot
    \addplot graphics[xmin=0,xmax=1,ymin=0,ymax=1] {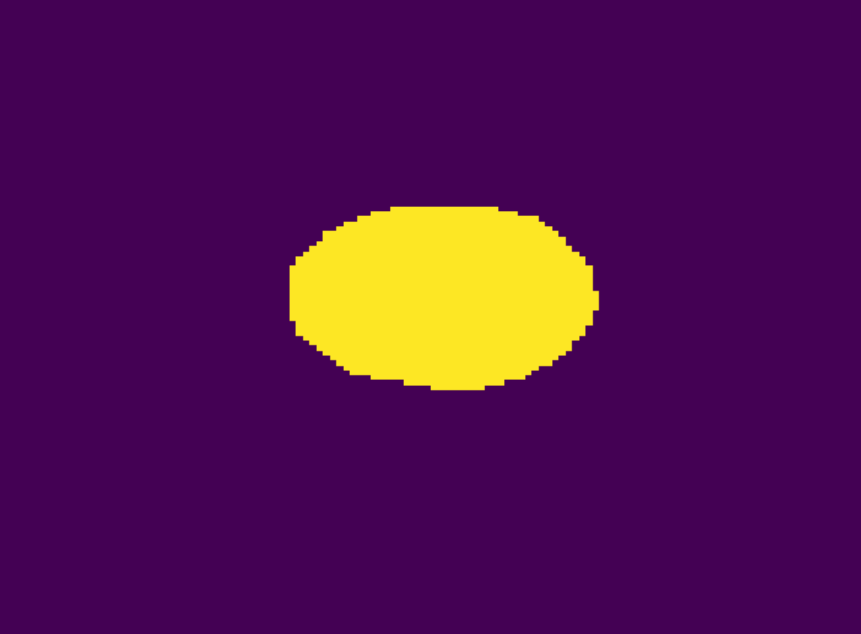};
    %%%%%%%%%%%%%%%%%%%%
    
    \nextgroupplot[ylabel={Anal canal}]
    \addplot graphics[xmin=0,xmax=1,ymin=0,ymax=1] {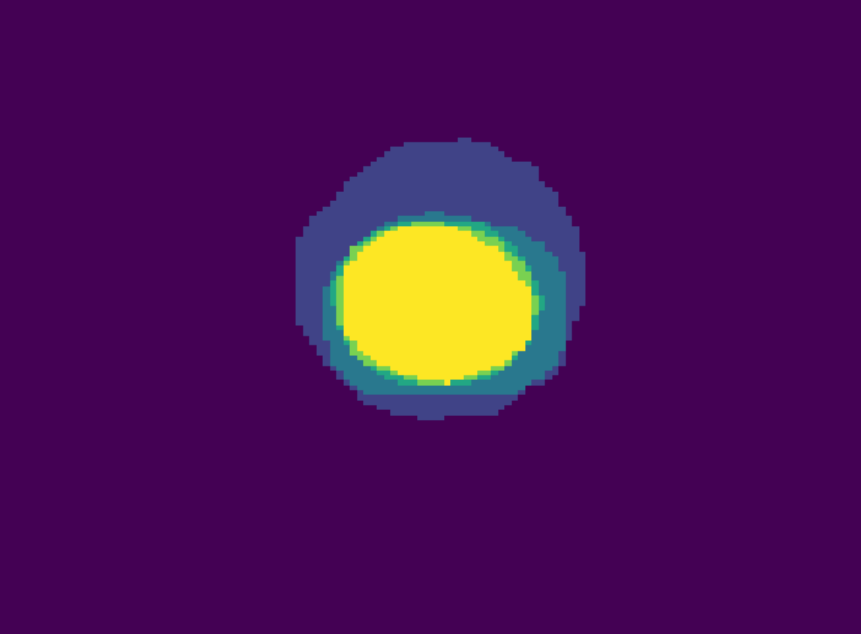};
    
    \nextgroupplot
    \addplot graphics[xmin=0,xmax=1,ymin=0,ymax=1] {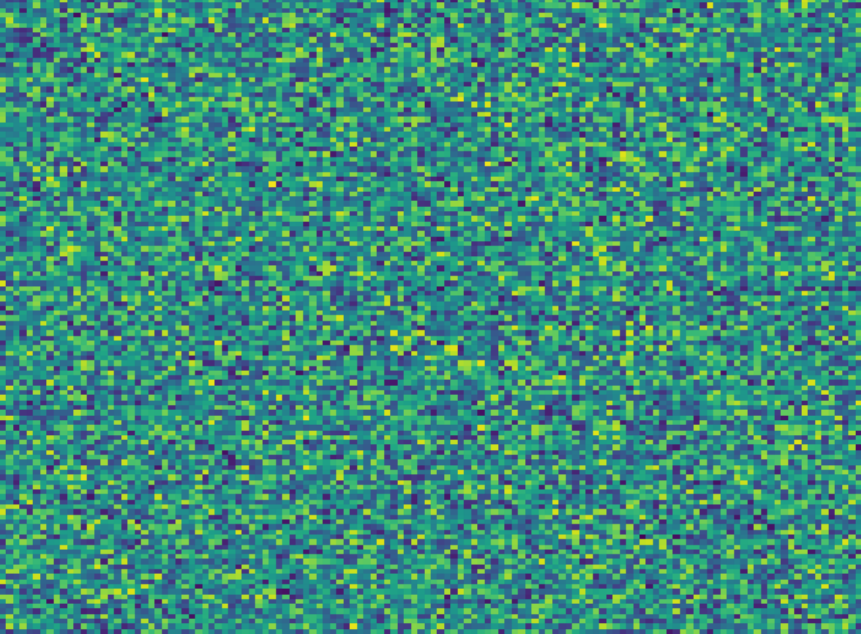};
    
    \nextgroupplot
    \addplot graphics[xmin=0,xmax=1,ymin=0,ymax=1] {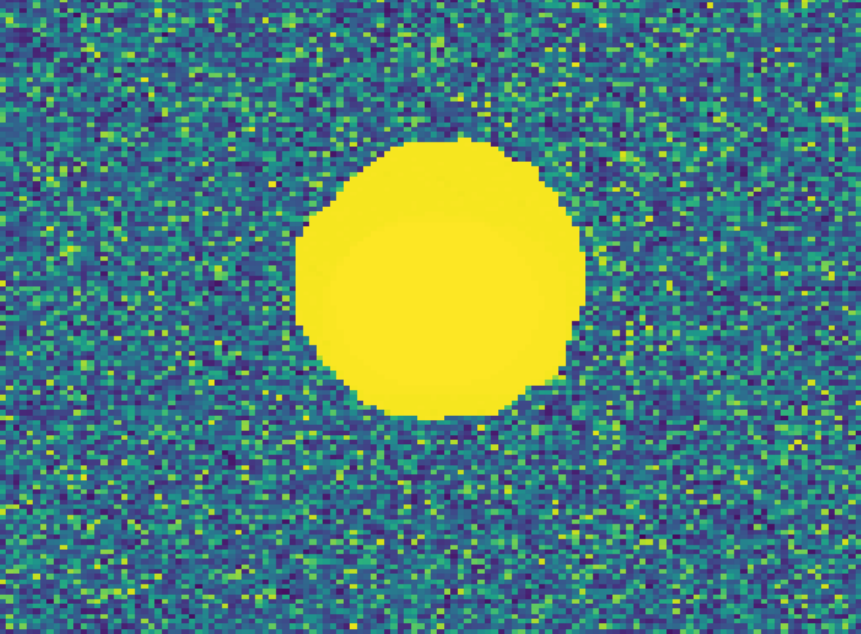};
  
    \nextgroupplot
    \addplot graphics[xmin=0,xmax=1,ymin=0,ymax=1] {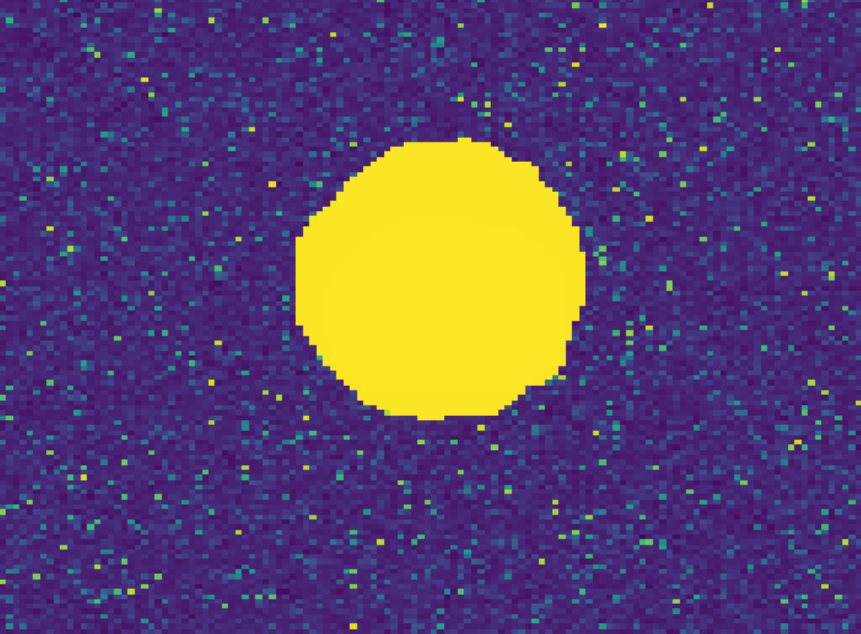};
 
    \nextgroupplot
    \addplot graphics[xmin=0,xmax=1,ymin=0,ymax=1] {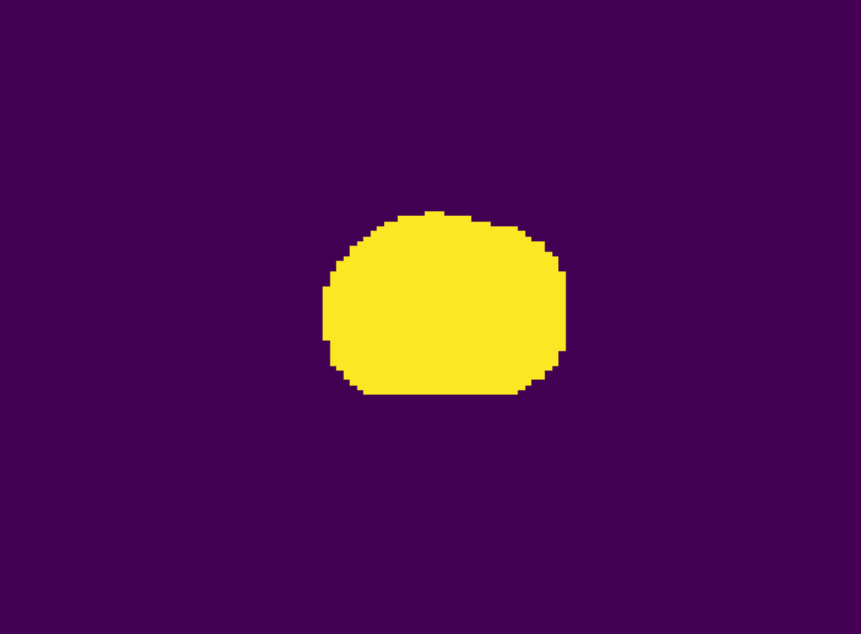};
    
    \nextgroupplot
    \addplot graphics[xmin=0,xmax=1,ymin=0,ymax=1] {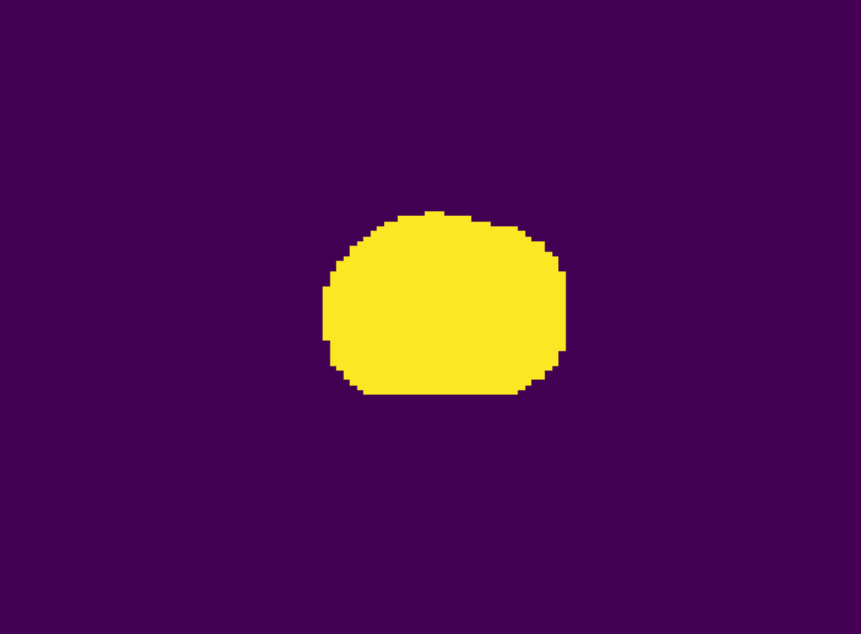};
    
    \nextgroupplot
    \addplot graphics[xmin=0,xmax=1,ymin=0,ymax=1] {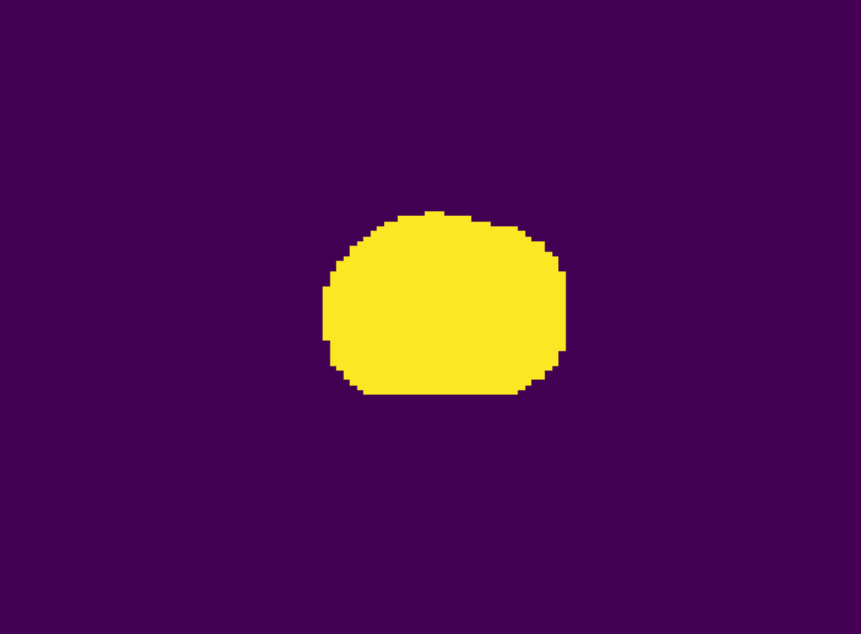};
    
    %%%%%%%%%%%%%%%%%%%%
    
    \nextgroupplot[ylabel={Penile bulb}]
    \addplot graphics[xmin=0,xmax=1,ymin=0,ymax=1] {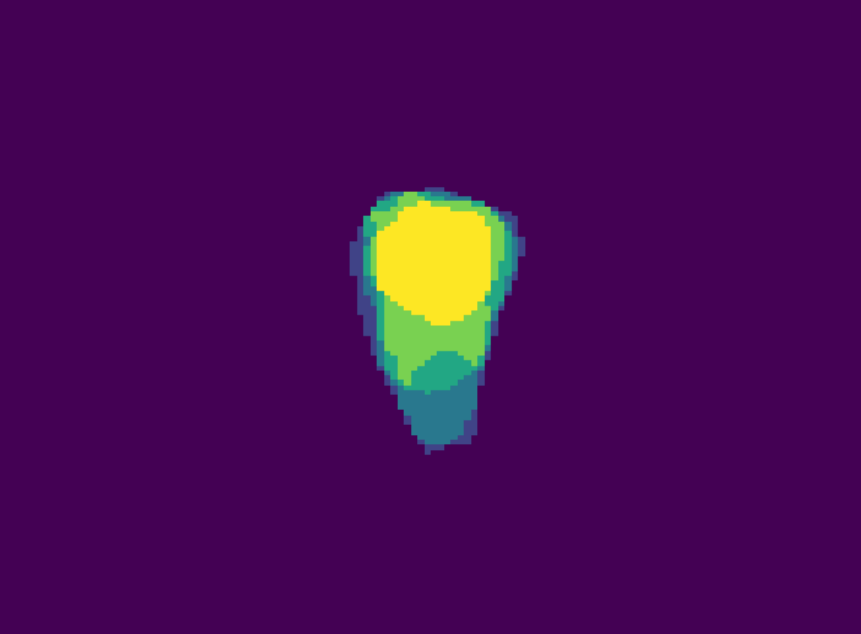};
    
    \nextgroupplot
    \addplot graphics[xmin=0,xmax=1,ymin=0,ymax=1] {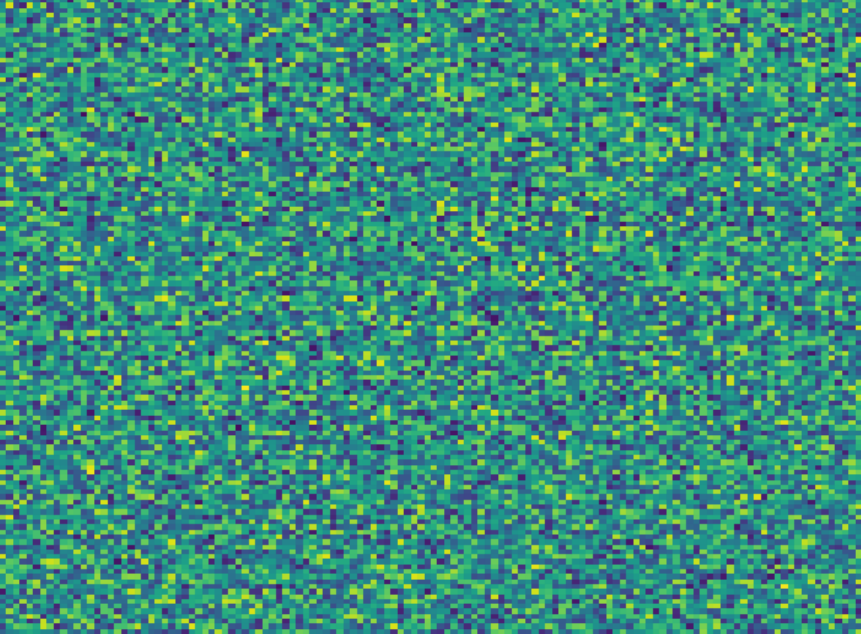};
    
    \nextgroupplot
    \addplot graphics[xmin=0,xmax=1,ymin=0,ymax=1] {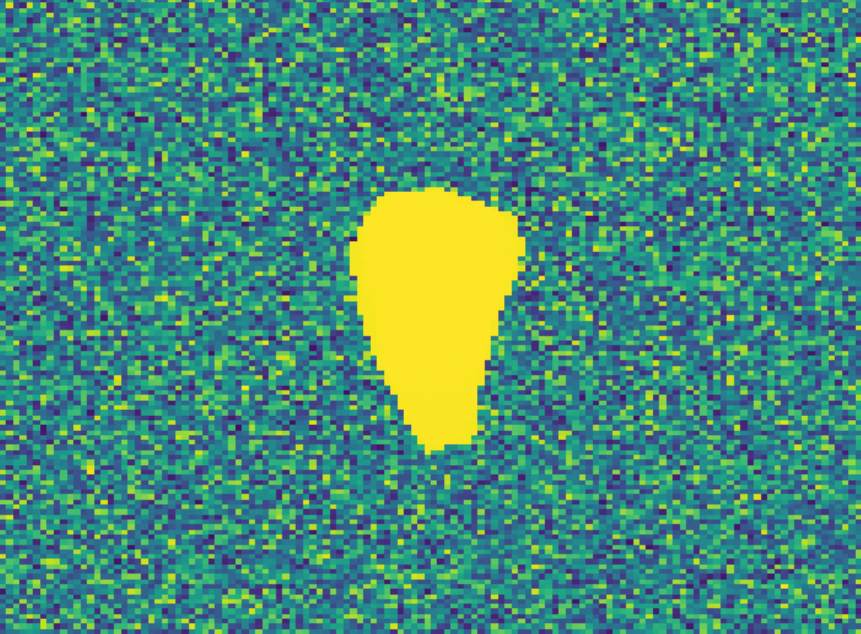};
  
    \nextgroupplot
    \addplot graphics[xmin=0,xmax=1,ymin=0,ymax=1] {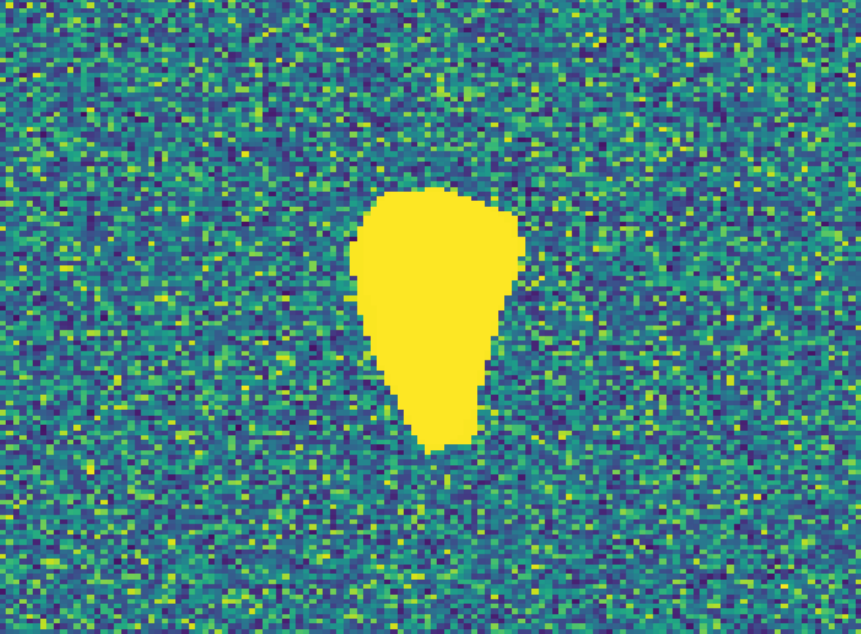};
 
    \nextgroupplot
    \addplot graphics[xmin=0,xmax=1,ymin=0,ymax=1] {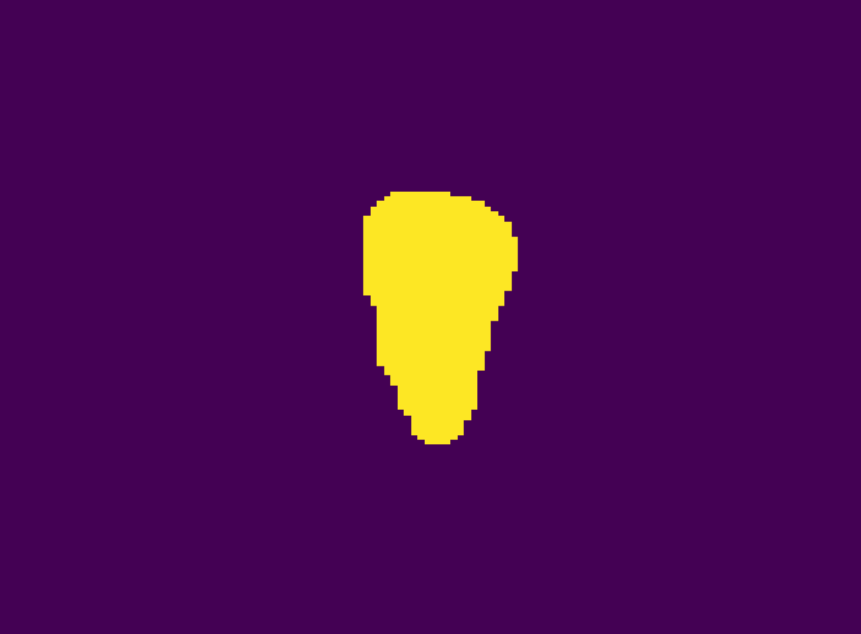};
    
    \nextgroupplot
    \addplot graphics[xmin=0,xmax=1,ymin=0,ymax=1] {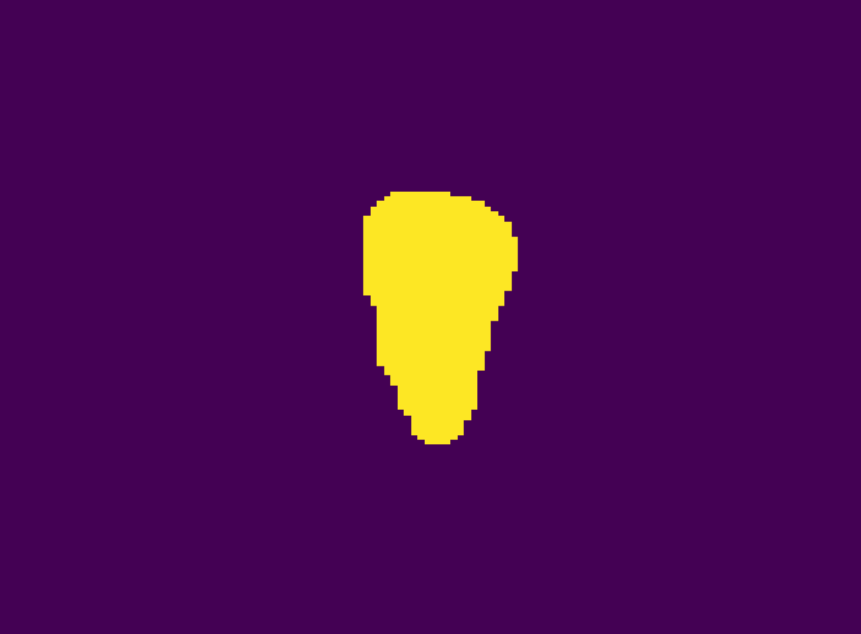};
    
    \nextgroupplot
    \addplot graphics[xmin=0,xmax=1,ymin=0,ymax=1] {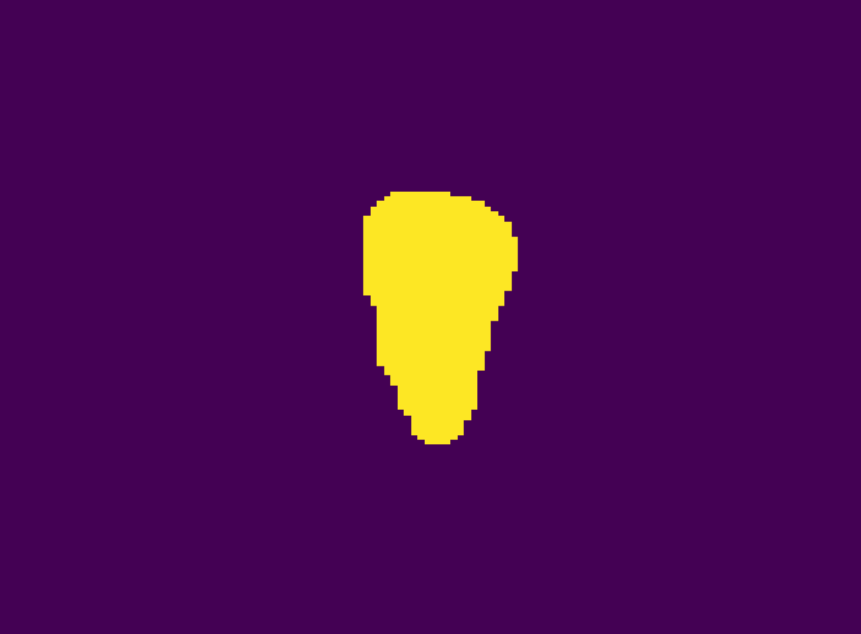};
    
    %%%%%%%%%%%%%%%%%%%%
    
    \nextgroupplot[ylabel={Neurov. b.}]
    \addplot graphics[xmin=0,xmax=1,ymin=0,ymax=1] {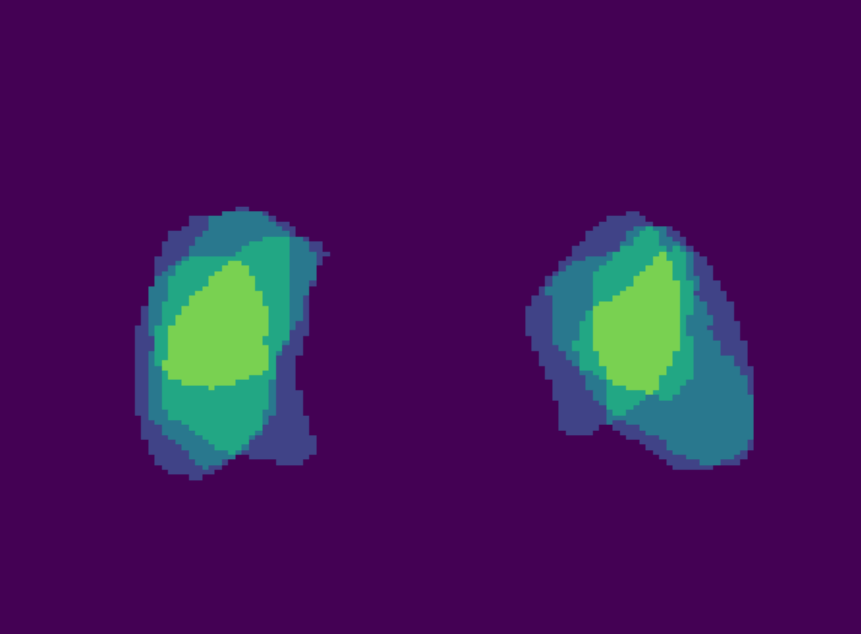};
    
    \nextgroupplot
    \addplot graphics[xmin=0,xmax=1,ymin=0,ymax=1] {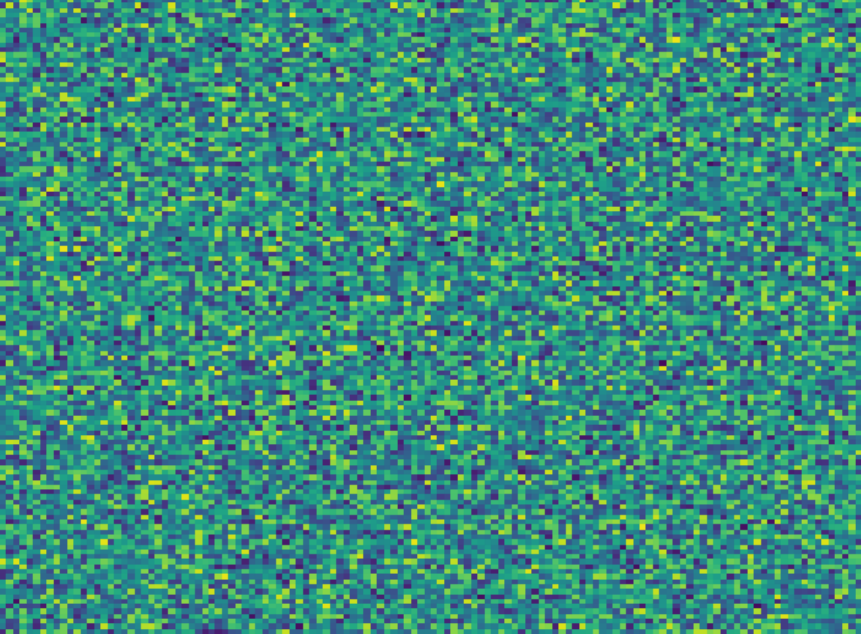};
    
    \nextgroupplot
    \addplot graphics[xmin=0,xmax=1,ymin=0,ymax=1] {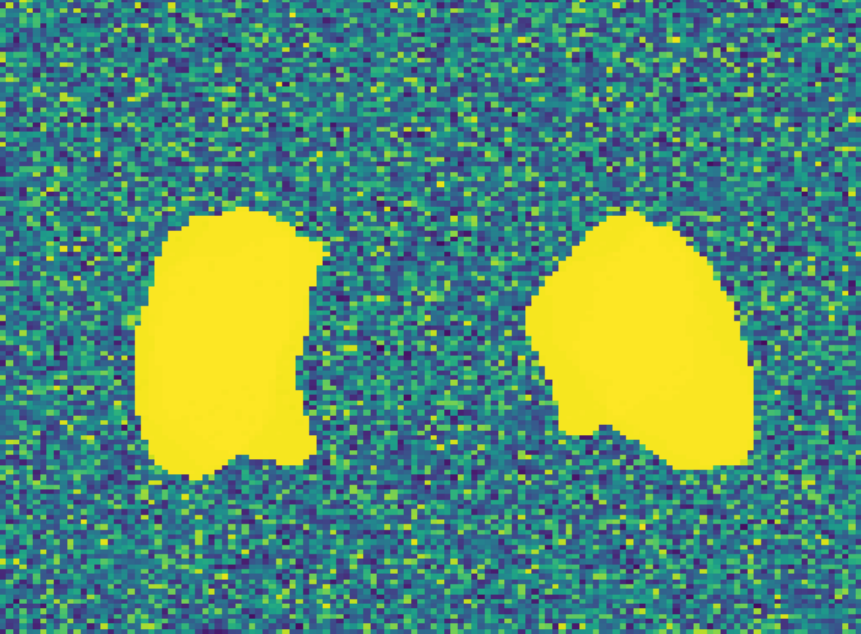};
  
    \nextgroupplot
    \addplot graphics[xmin=0,xmax=1,ymin=0,ymax=1] {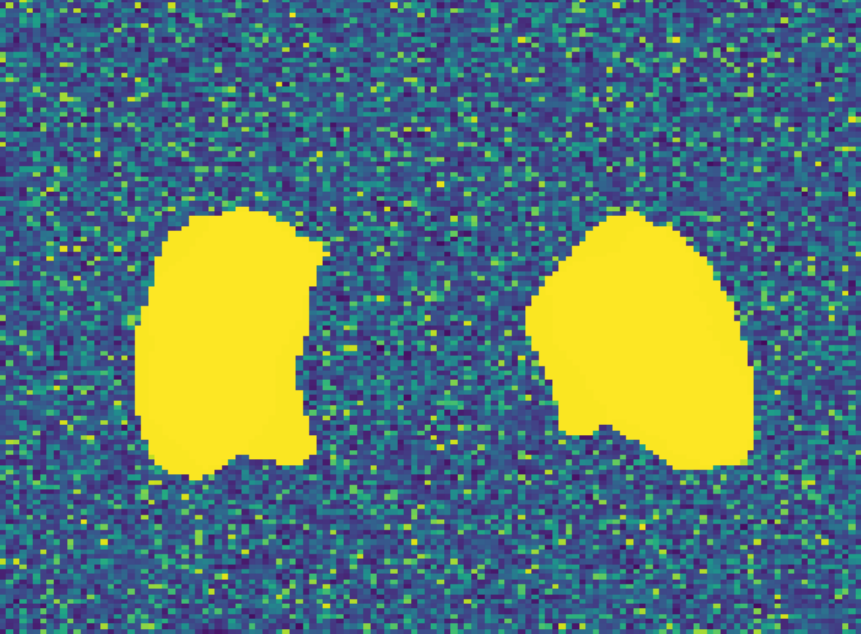};
 
    \nextgroupplot
    \addplot graphics[xmin=0,xmax=1,ymin=0,ymax=1] {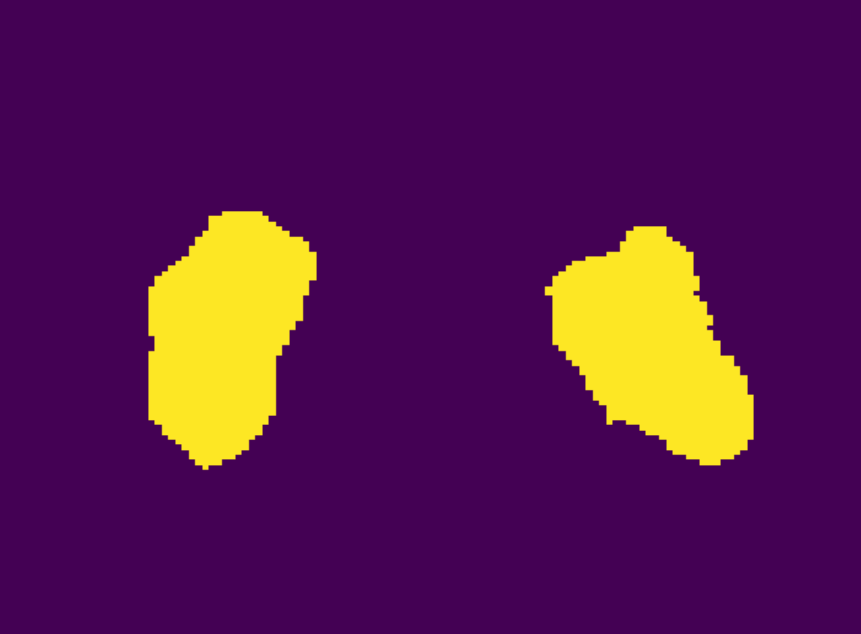};
    
    \nextgroupplot
    \addplot graphics[xmin=0,xmax=1,ymin=0,ymax=1] {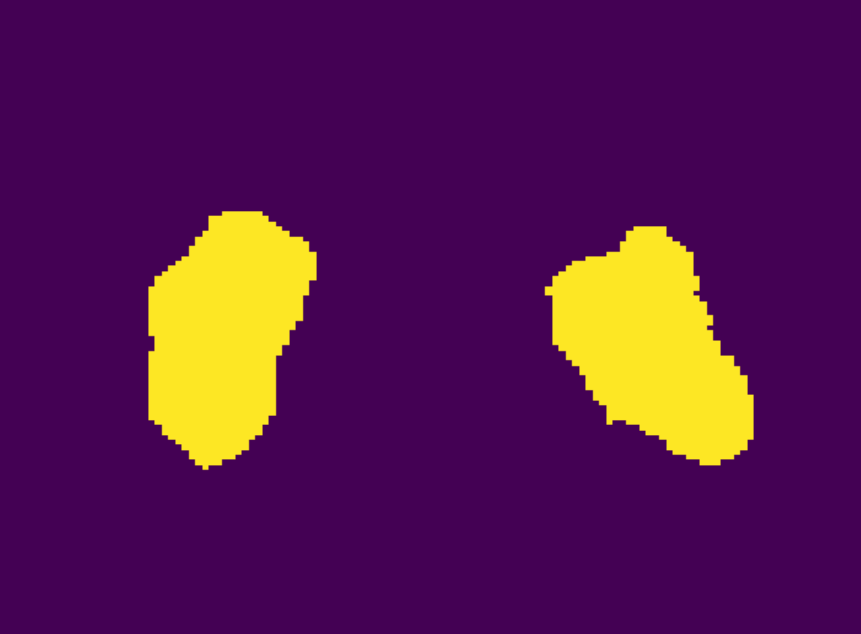};
    
    \nextgroupplot
    \addplot graphics[xmin=0,xmax=1,ymin=0,ymax=1] {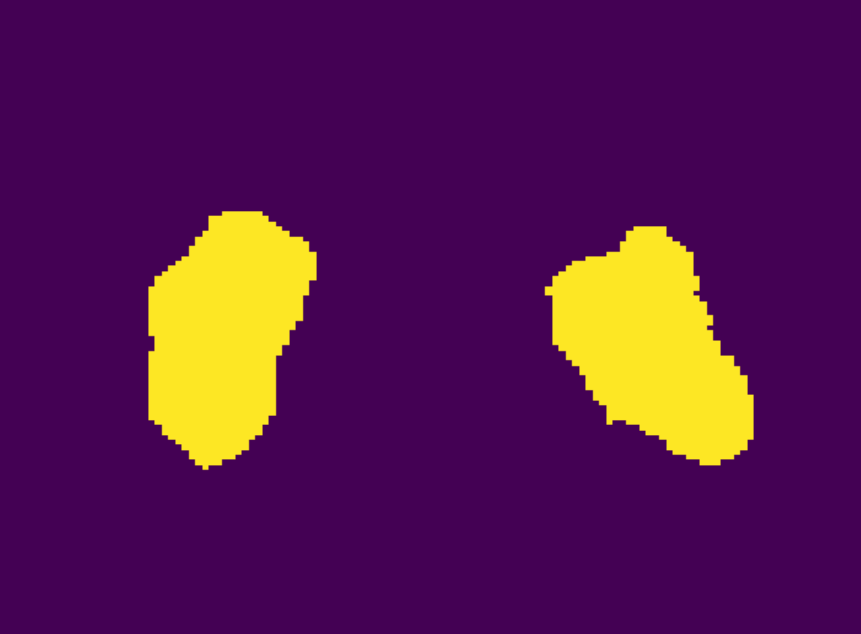};
    
    %%%%%%%%%%%%%%%%%%%%
    
    \nextgroupplot[ylabel={Femoral h. R} ]
    \addplot graphics[xmin=0,xmax=1,ymin=0,ymax=1] {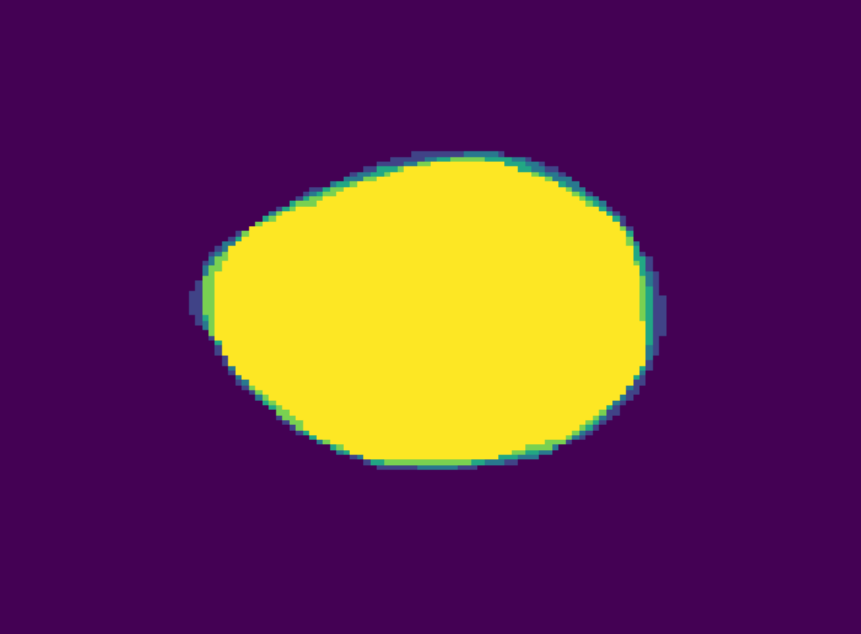};
    
    \nextgroupplot
    \addplot graphics[xmin=0,xmax=1,ymin=0,ymax=1] {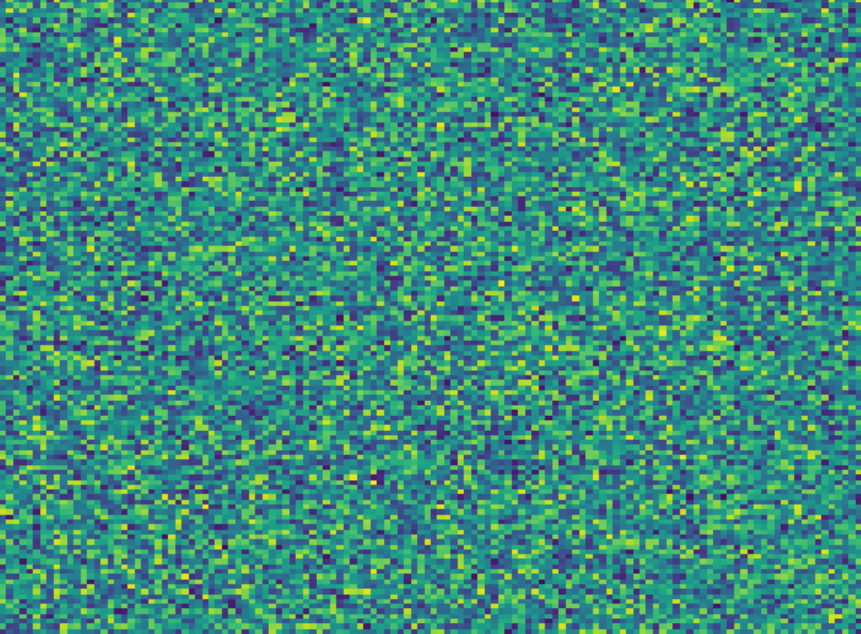};
    
    \nextgroupplot
    \addplot graphics[xmin=0,xmax=1,ymin=0,ymax=1] {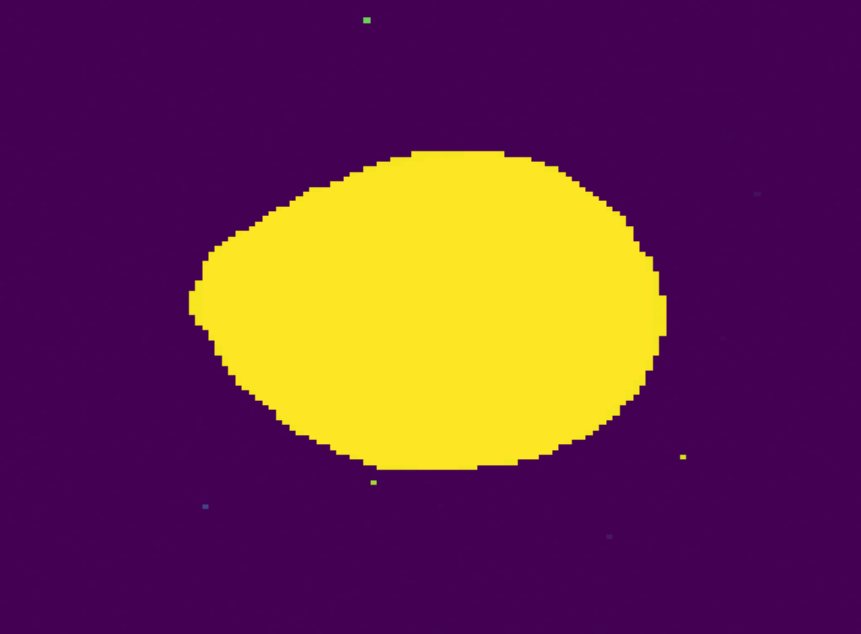};
  
    \nextgroupplot
    \addplot graphics[xmin=0,xmax=1,ymin=0,ymax=1] {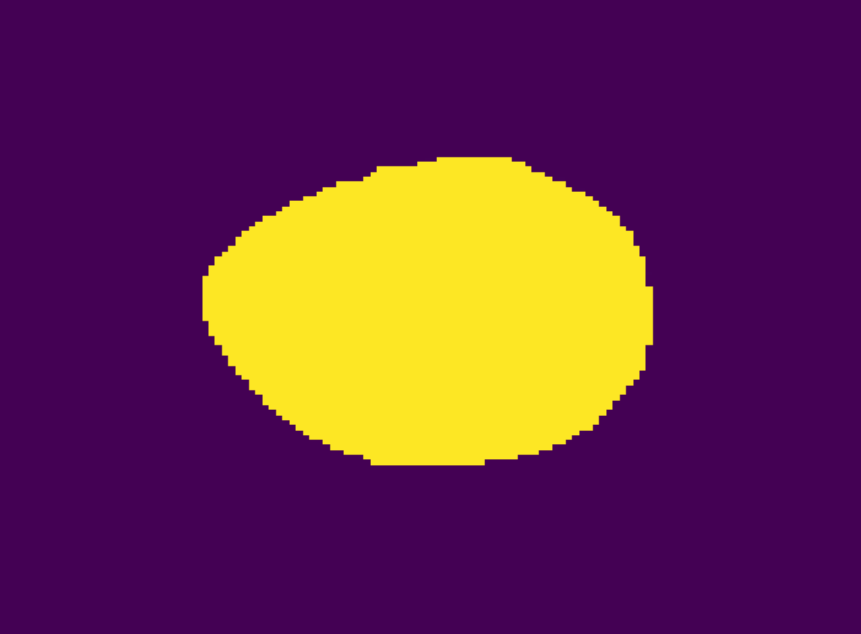};
 
    \nextgroupplot
    \addplot graphics[xmin=0,xmax=1,ymin=0,ymax=1] {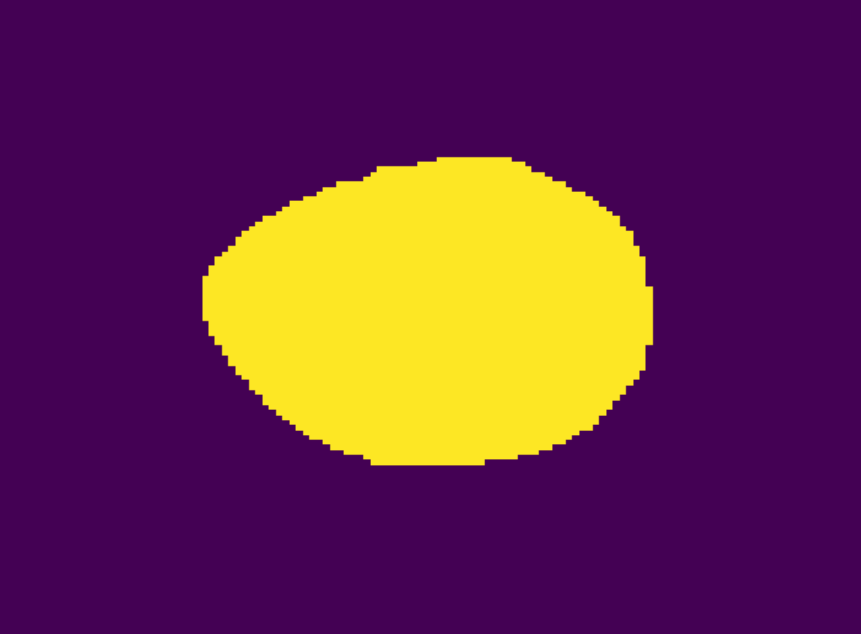};
    
    \nextgroupplot
    \addplot graphics[xmin=0,xmax=1,ymin=0,ymax=1] {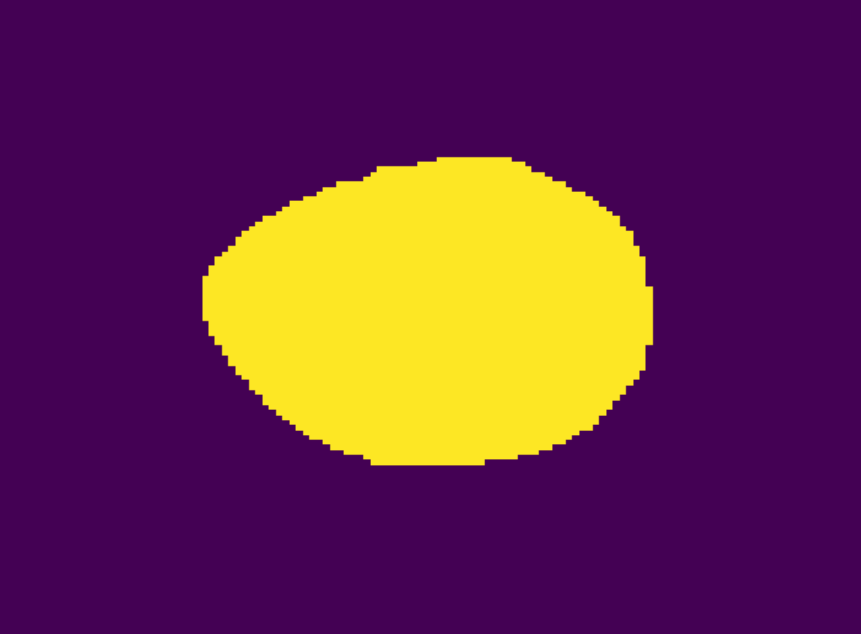};
    
    \nextgroupplot
    \addplot graphics[xmin=0,xmax=1,ymin=0,ymax=1] {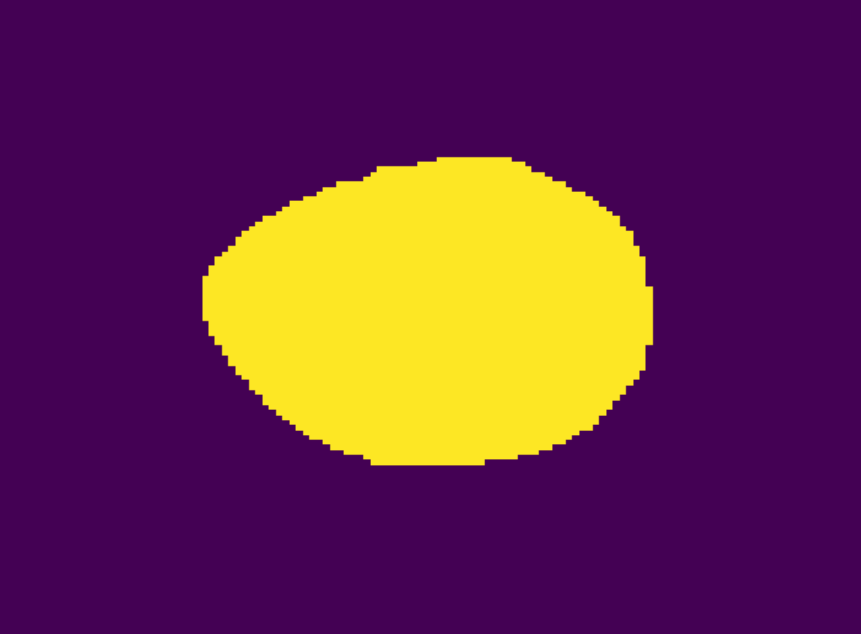};
    
    %%%%%%%%%%%%%%%%%%%%
    
    \nextgroupplot[ylabel={Femoral h. L} ]
    \addplot graphics[xmin=0,xmax=1,ymin=0,ymax=1] {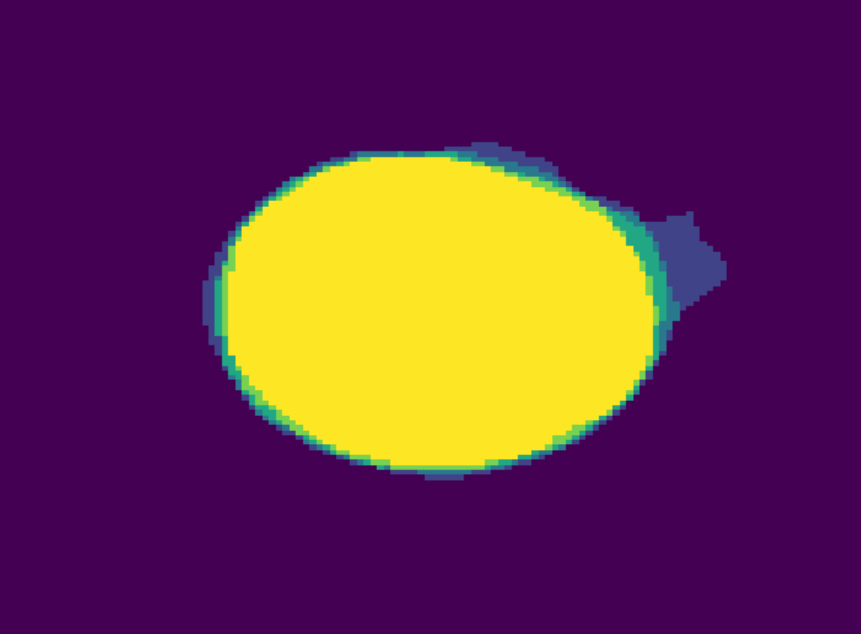};
    
    \nextgroupplot
    \addplot graphics[xmin=0,xmax=1,ymin=0,ymax=1] {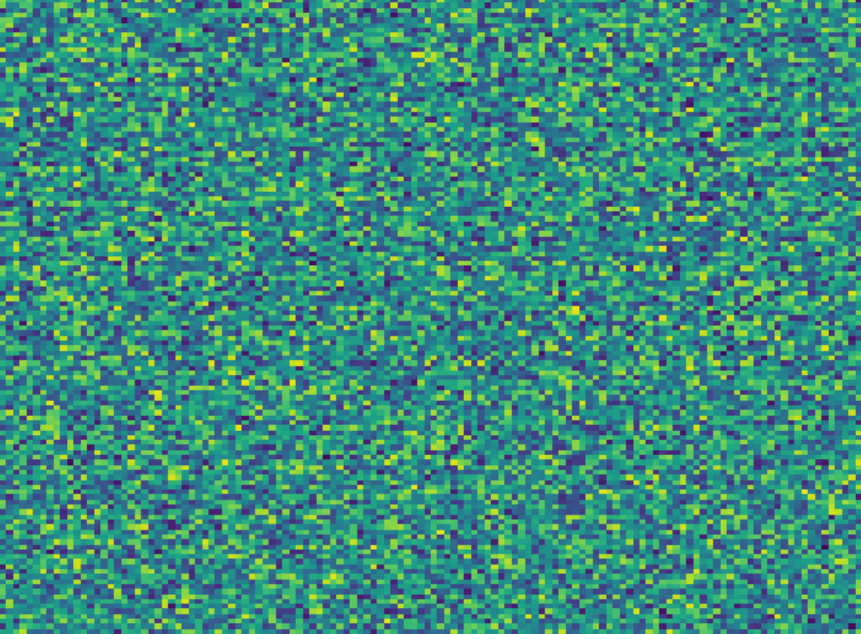};
    
    \nextgroupplot
    \addplot graphics[xmin=0,xmax=1,ymin=0,ymax=1] {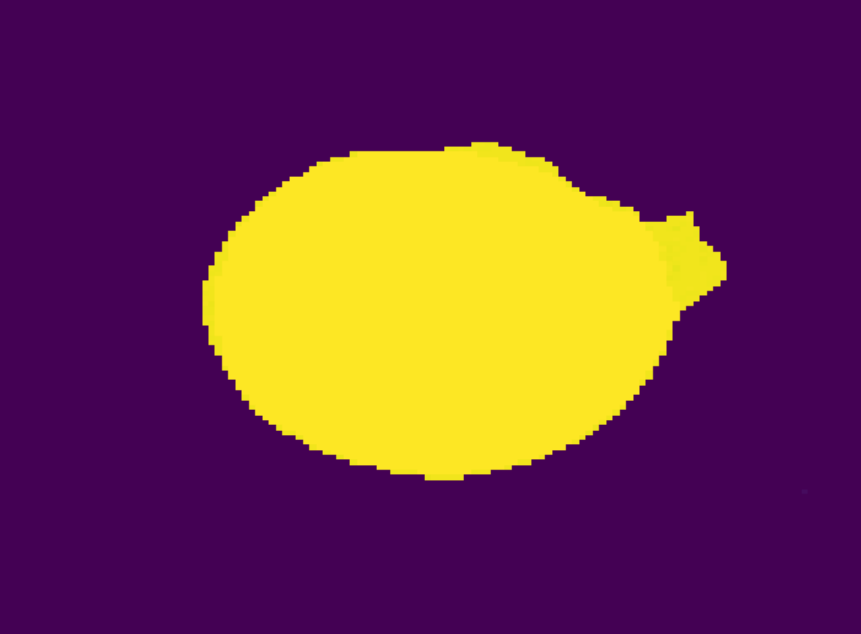};
  
    \nextgroupplot
    \addplot graphics[xmin=0,xmax=1,ymin=0,ymax=1] {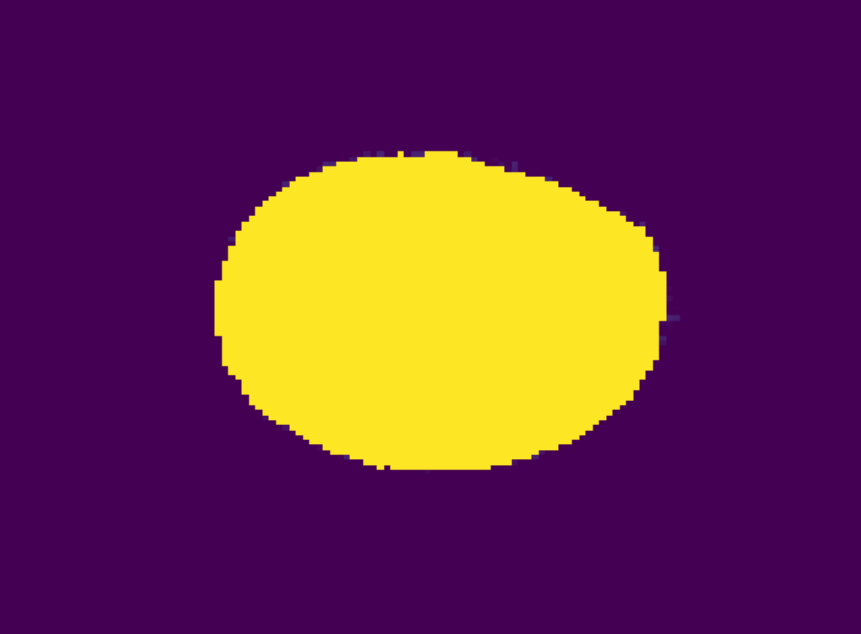};
 
    \nextgroupplot
    \addplot graphics[xmin=0,xmax=1,ymin=0,ymax=1] {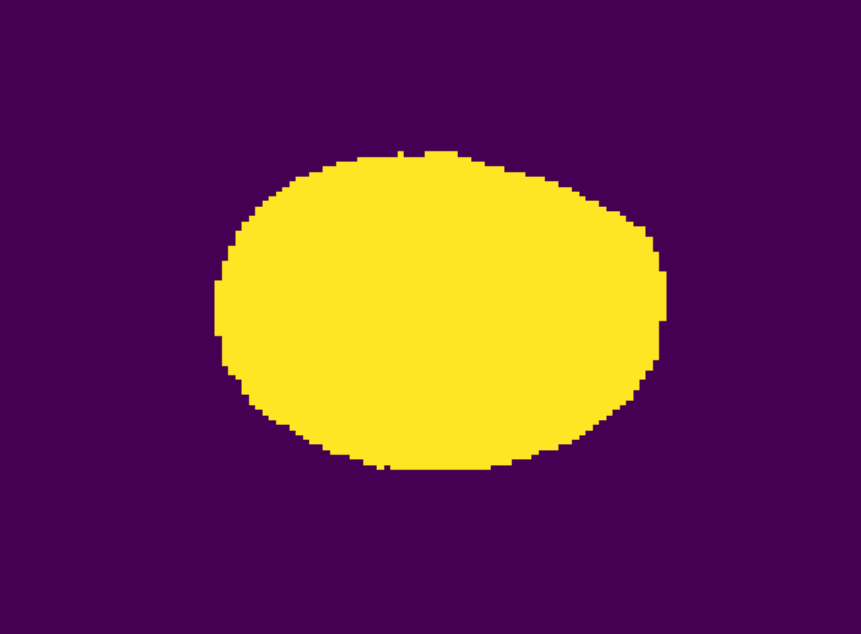};
    
    \nextgroupplot
    \addplot graphics[xmin=0,xmax=1,ymin=0,ymax=1] {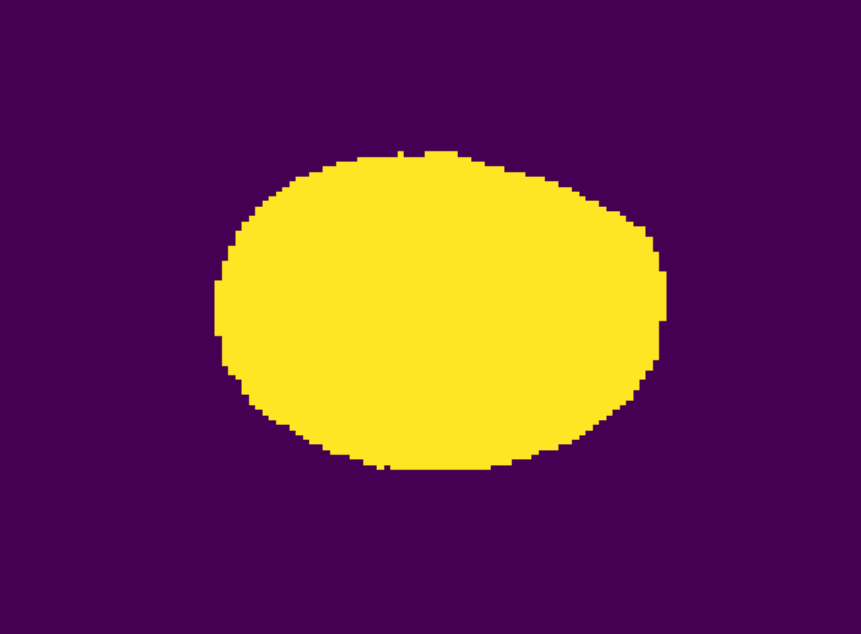};
    
    \nextgroupplot
    \addplot graphics[xmin=0,xmax=1,ymin=0,ymax=1] {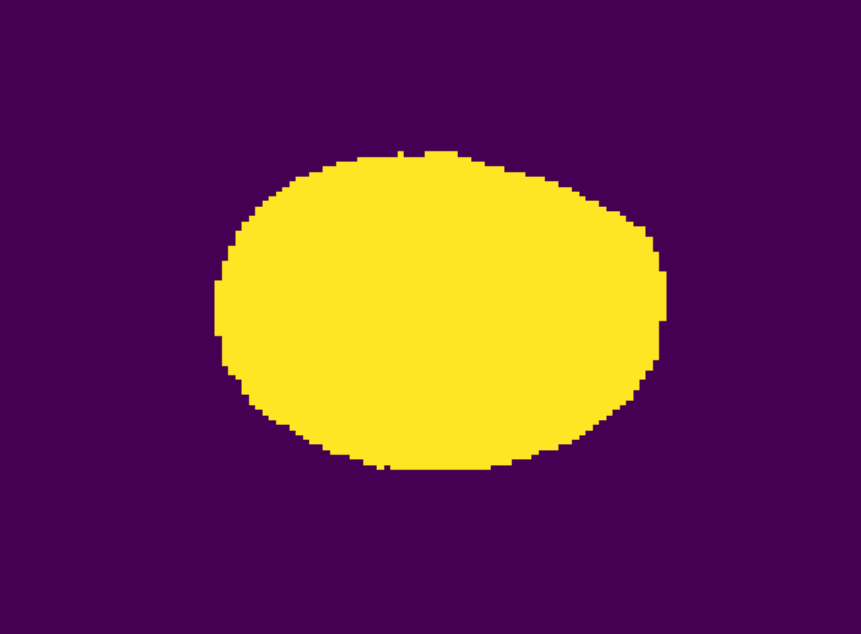};
    
    %%%%%%%%%%%%%%%%%%%%
    
    \nextgroupplot[ylabel={Prostate} ]
    \addplot graphics[xmin=0,xmax=1,ymin=0,ymax=1] {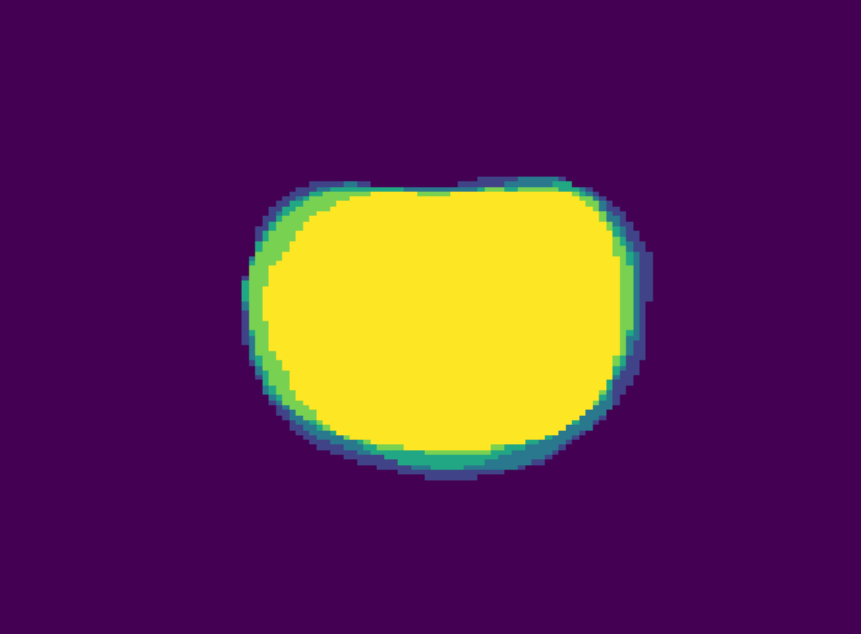};
    
    \nextgroupplot
    \addplot graphics[xmin=0,xmax=1,ymin=0,ymax=1] {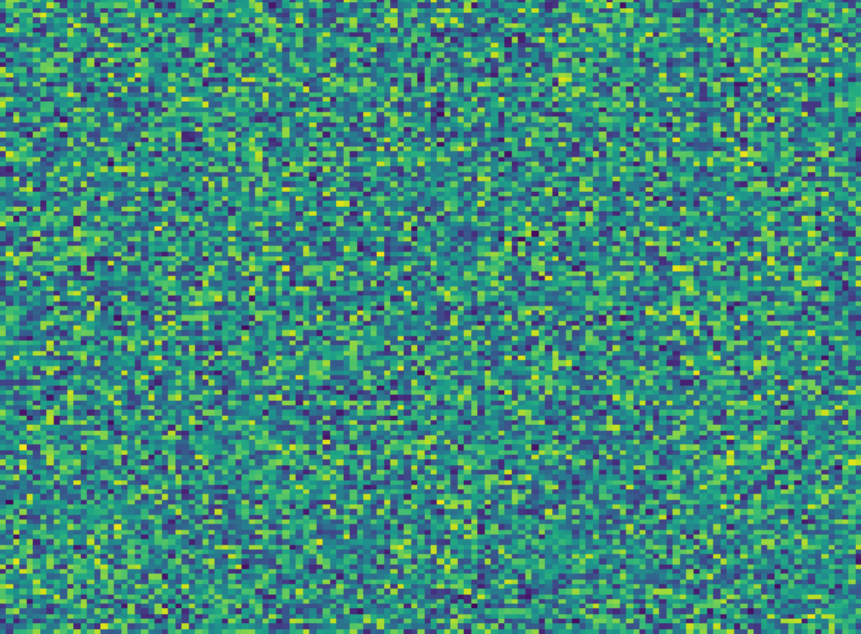};
    
    \nextgroupplot
    \addplot graphics[xmin=0,xmax=1,ymin=0,ymax=1] {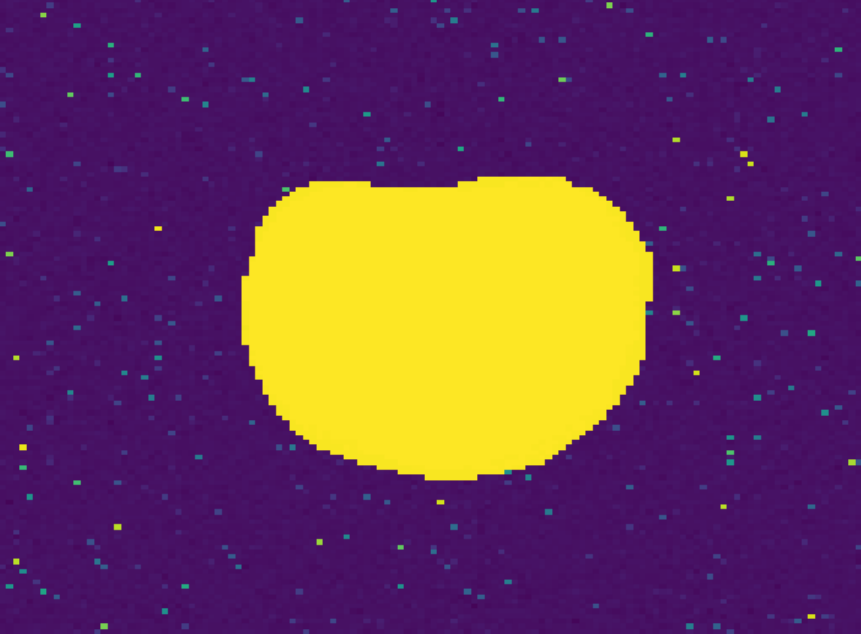};
  
    \nextgroupplot
    \addplot graphics[xmin=0,xmax=1,ymin=0,ymax=1] {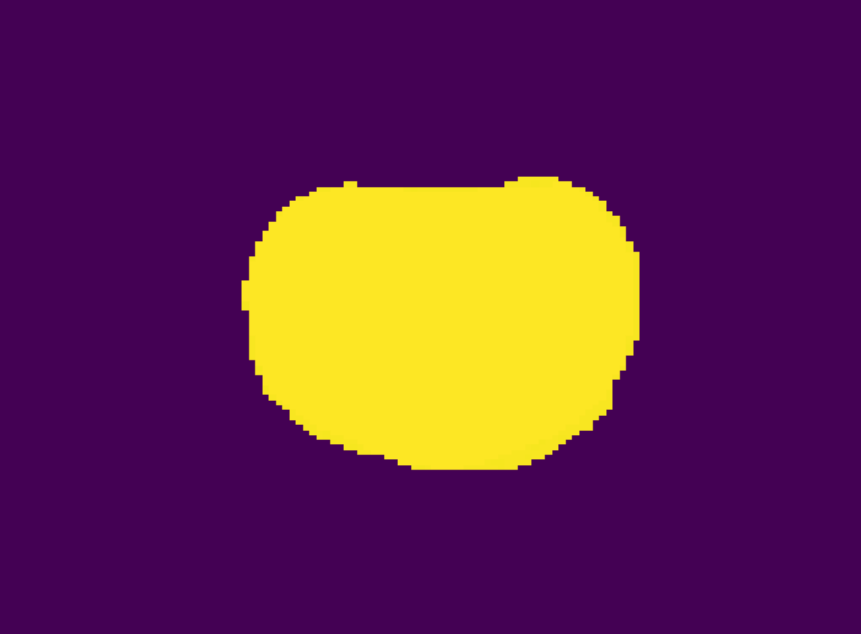};
 
    \nextgroupplot
    \addplot graphics[xmin=0,xmax=1,ymin=0,ymax=1] {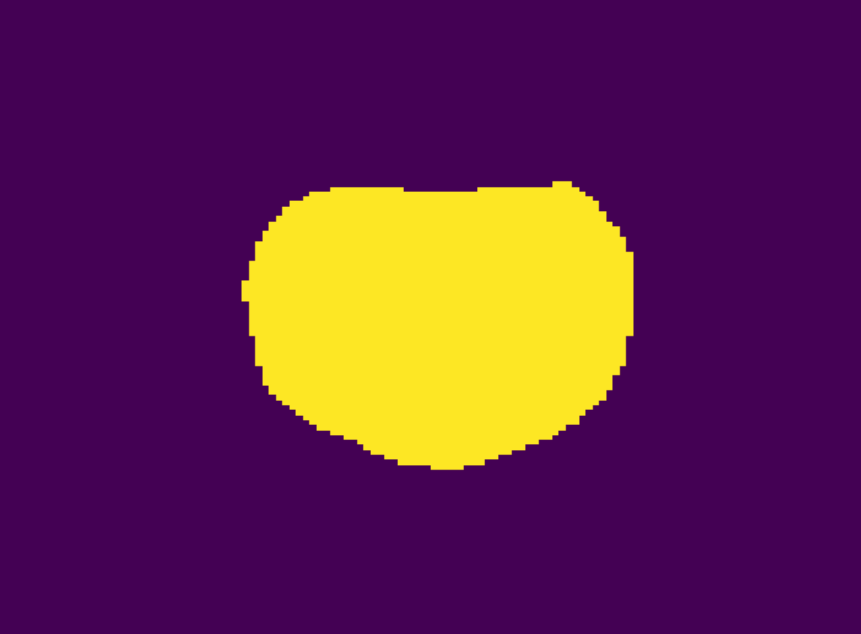};
    
    \nextgroupplot
    \addplot graphics[xmin=0,xmax=1,ymin=0,ymax=1] {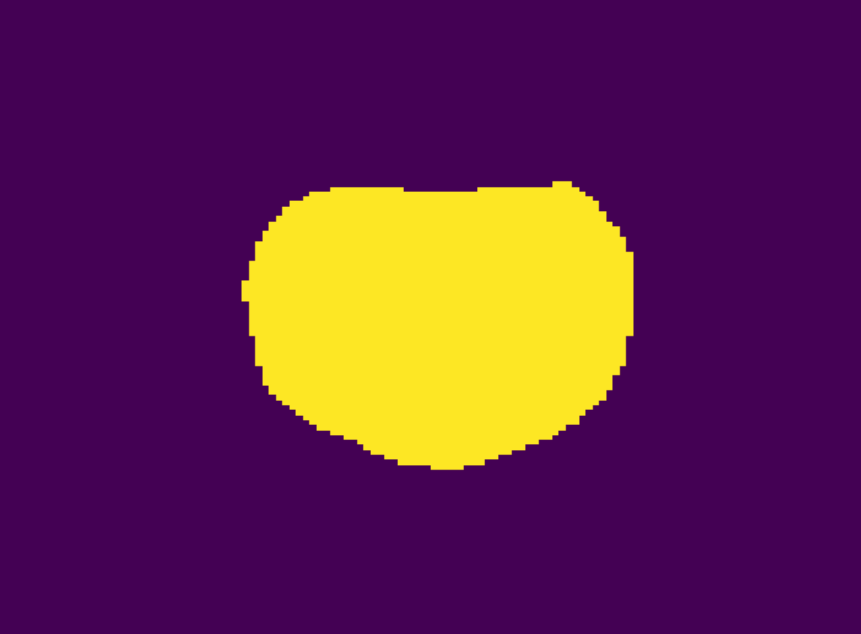};
    
    \nextgroupplot
    \addplot graphics[xmin=0,xmax=1,ymin=0,ymax=1] {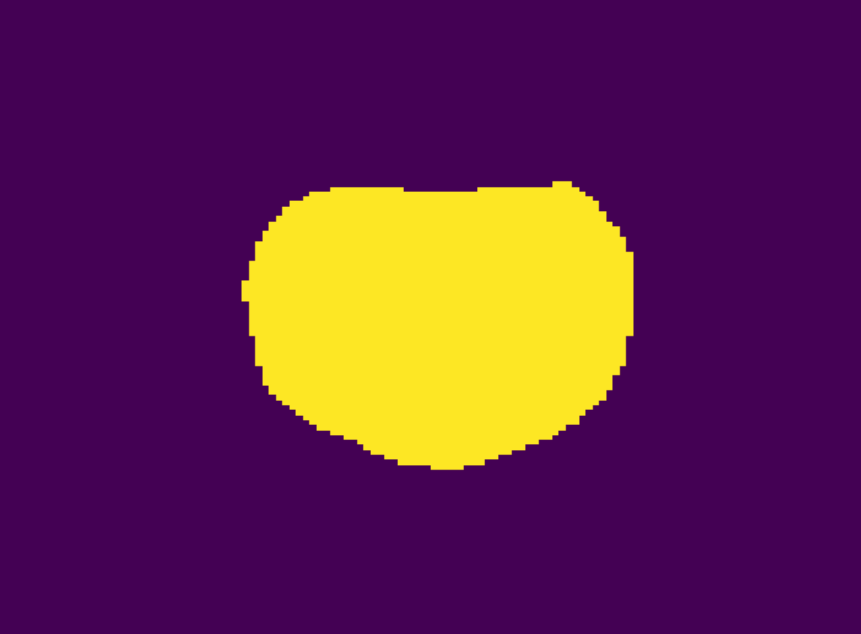};

    %%%%%%%%%%%%%%%%%%%%
    
    \nextgroupplot[ylabel={Seminal v.} ]
    \addplot graphics[xmin=0,xmax=1,ymin=0,ymax=1] {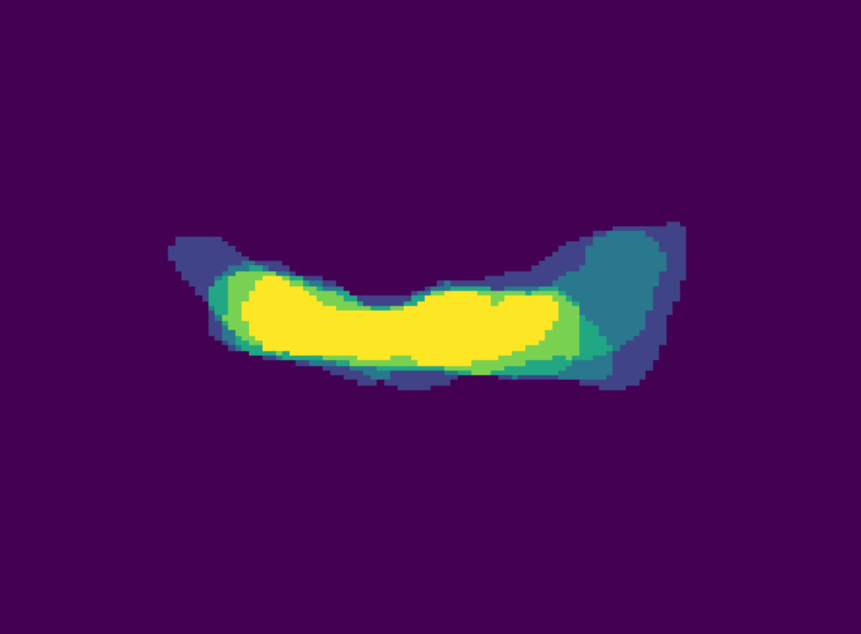};
    
    \nextgroupplot
    \addplot graphics[xmin=0,xmax=1,ymin=0,ymax=1] {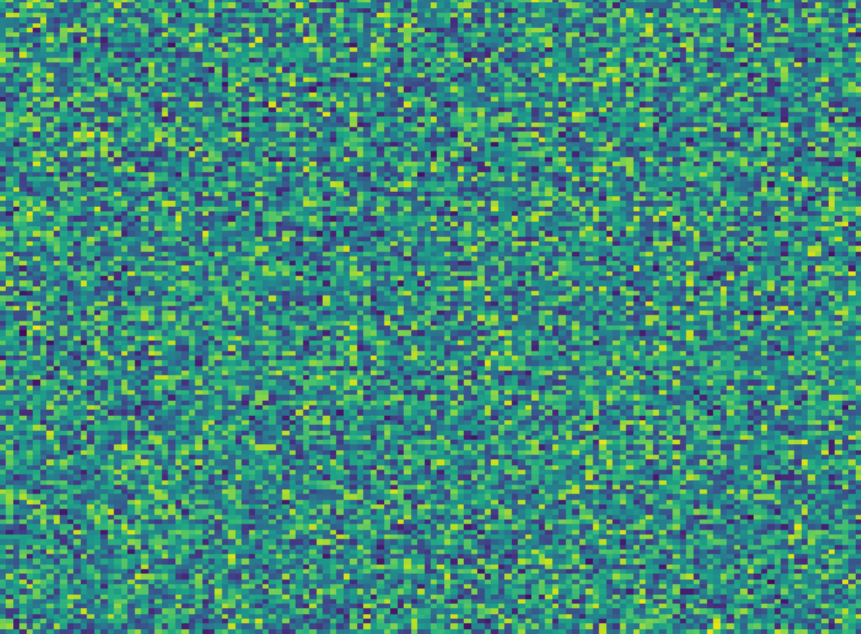};
    
    \nextgroupplot
    \addplot graphics[xmin=0,xmax=1,ymin=0,ymax=1] {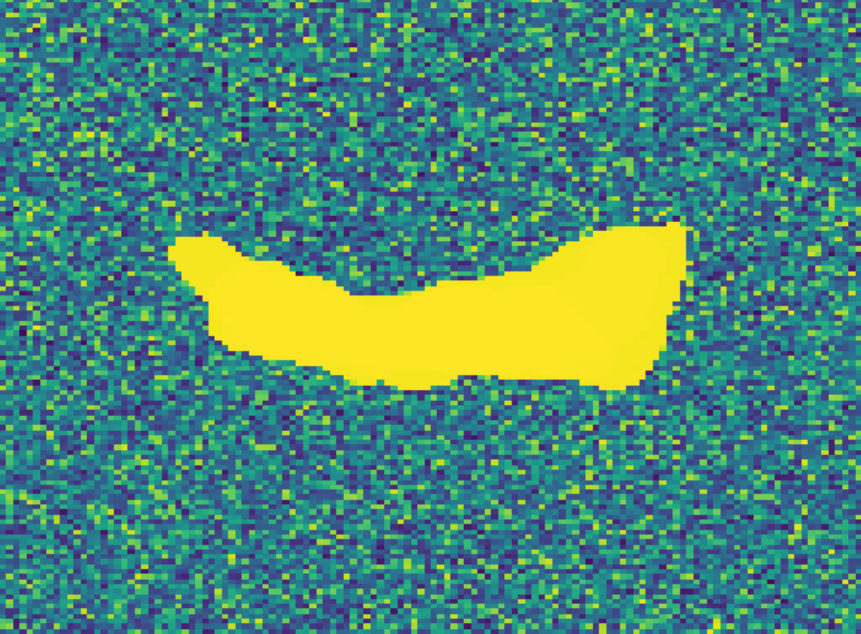};
  
    \nextgroupplot
    \addplot graphics[xmin=0,xmax=1,ymin=0,ymax=1] {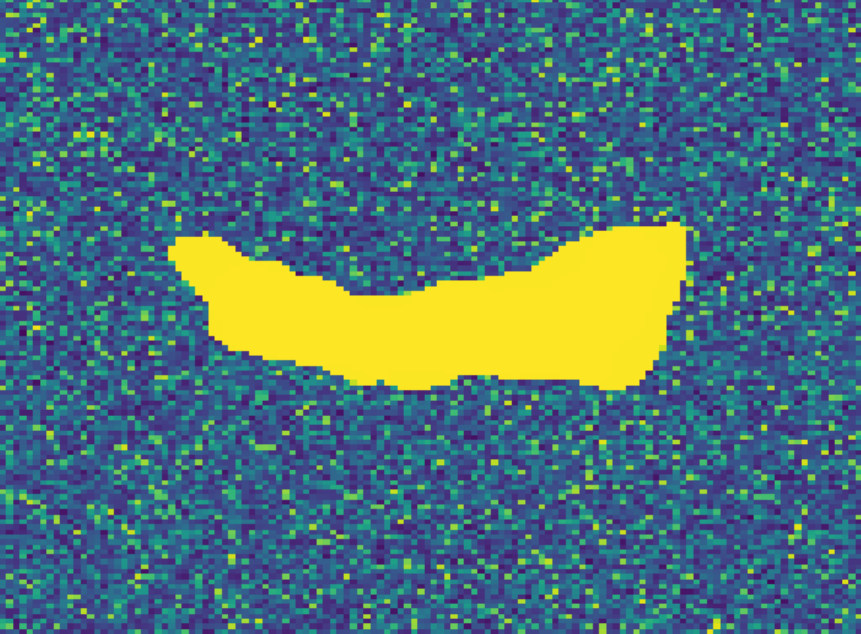};
 
    \nextgroupplot
    \addplot graphics[xmin=0,xmax=1,ymin=0,ymax=1] {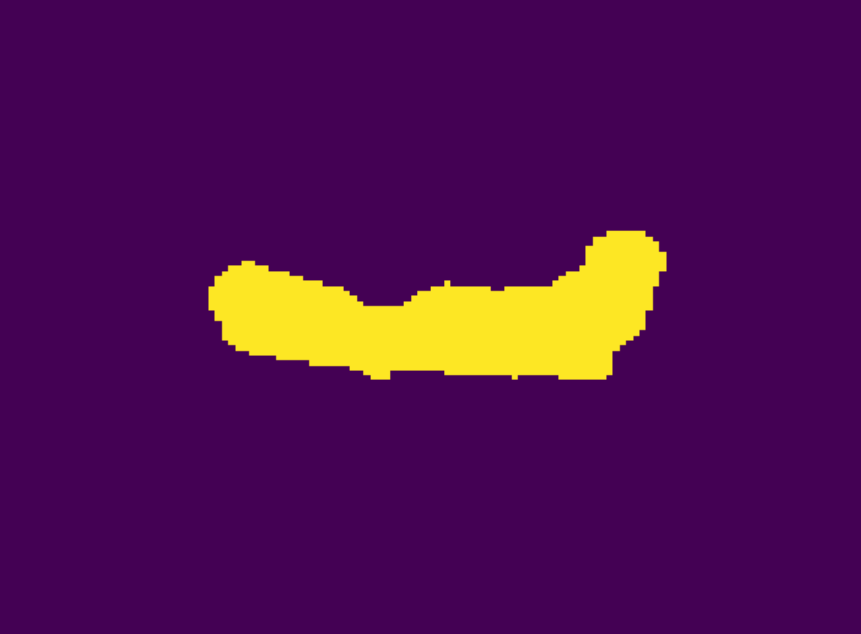};
    
    \nextgroupplot
    \addplot graphics[xmin=0,xmax=1,ymin=0,ymax=1] {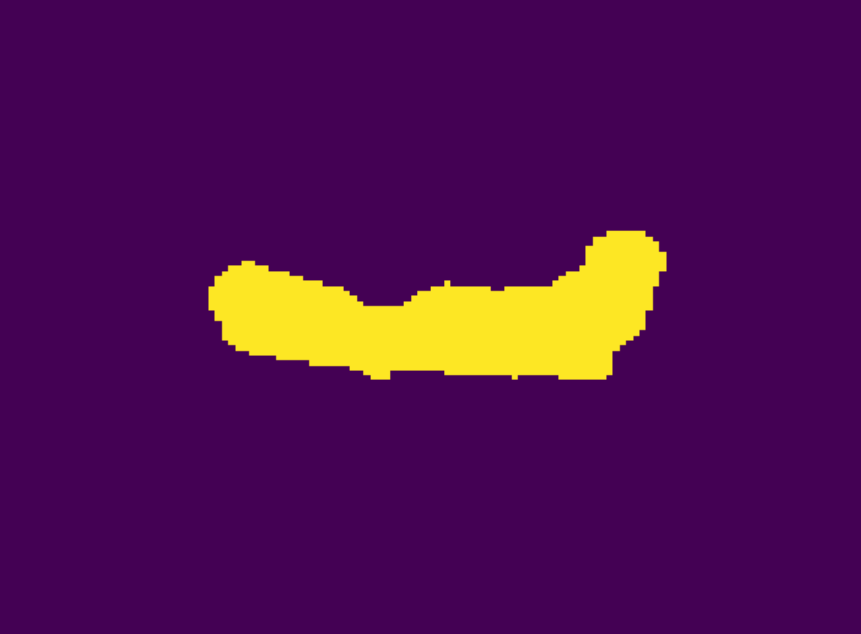};
    
    \nextgroupplot
    \addplot graphics[xmin=0,xmax=1,ymin=0,ymax=1] {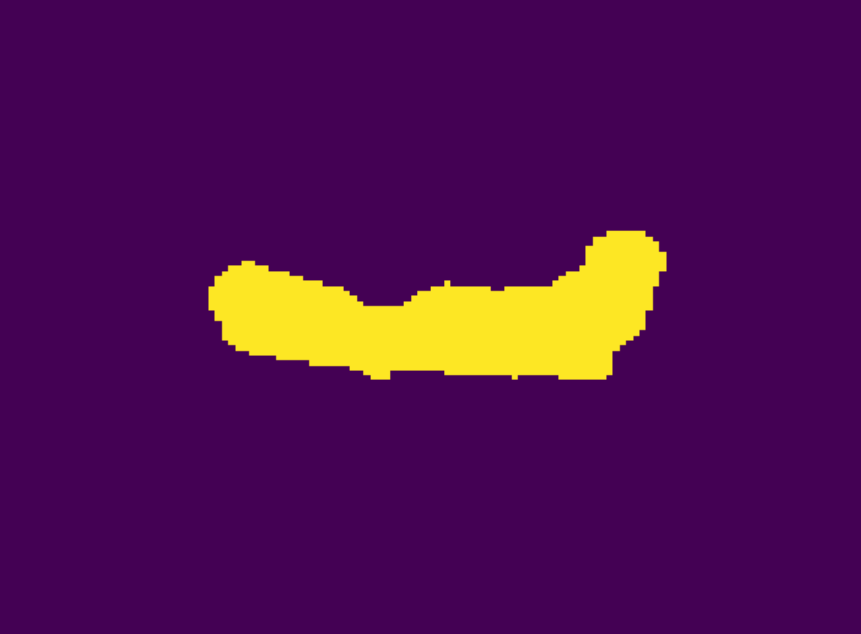};
    
  \end{groupplot}
\end{tikzpicture}

  \caption{
Each row corresponds to a two dimensional slice associated with an example from one of the experiments in (G).
The first column depicts a marginal function $m$.
In column 2-6 various iterations from gradient descent minimization of soft-Dice $\mathrm{SD}_m(\sigma\circ f)$ and a random starting point is depicted. 
The last column depicts the theoretical optimal solutions $s=I_{[\sup_{s'\in\mathcal{S}}\mathrm{D}_m(s')/2,1]} \circ m$ described in Theorem~\ref{theorem1}.
% The rows are generated using the synthetic model parameters $\sigma=0.01,0.02,0.03,0.04,0.05,0.06,0.07,0.08$.
}
  \label{seqGfig}
\end{figure*}

\begin{figure*}[htb!]
\section{Samples from experiments (S)}
  \centering
\begin{tikzpicture}
  \begin{groupplot}[group style={group size=7 by 9, horizontal sep=0.05cm, vertical sep=0.05cm},height=3.70cm,width=3.70cm,xmin=0,xmax=1,ymin=0,ymax=1, xticklabels=\empty, yticklabels=\empty]
  
    \nextgroupplot[title={$m$}, ylabel={$\rho=0.01$}]
    \addplot graphics[xmin=0,xmax=1,ymin=0,ymax=1] {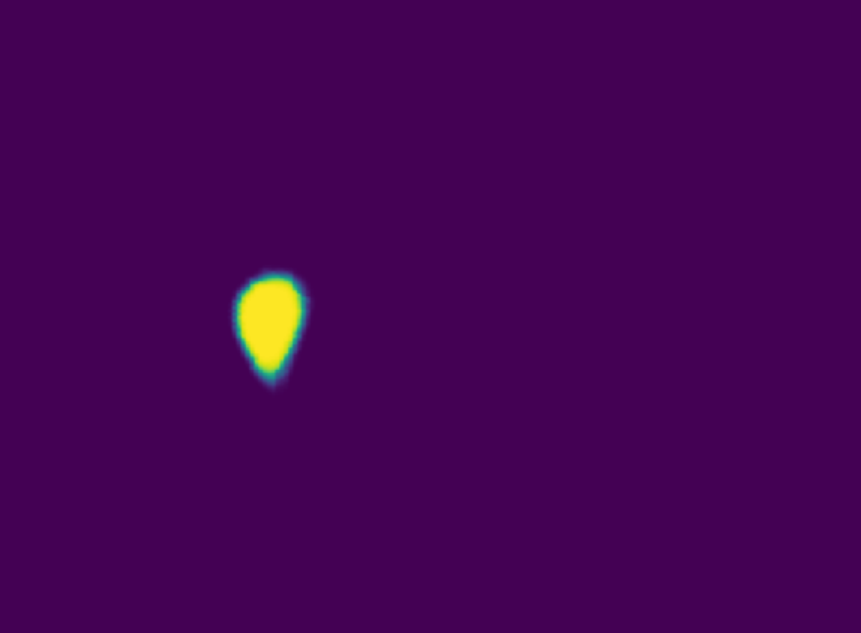};
    
    \nextgroupplot[title={$\sigma\circ f_1$}]
    \addplot graphics[xmin=0,xmax=1,ymin=0,ymax=1] {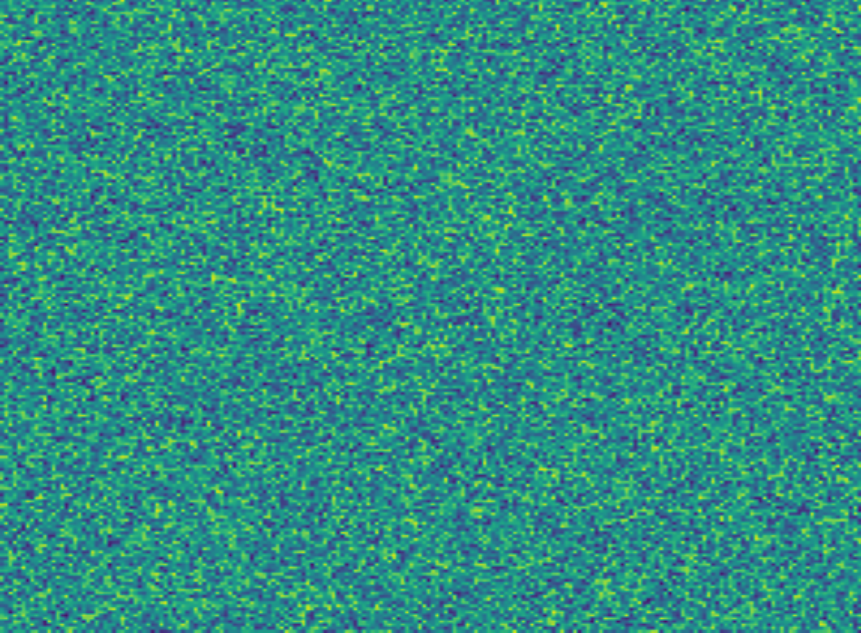};
    
    \nextgroupplot[title=$\sigma\circ f_{10}$]
    \addplot graphics[xmin=0,xmax=1,ymin=0,ymax=1] {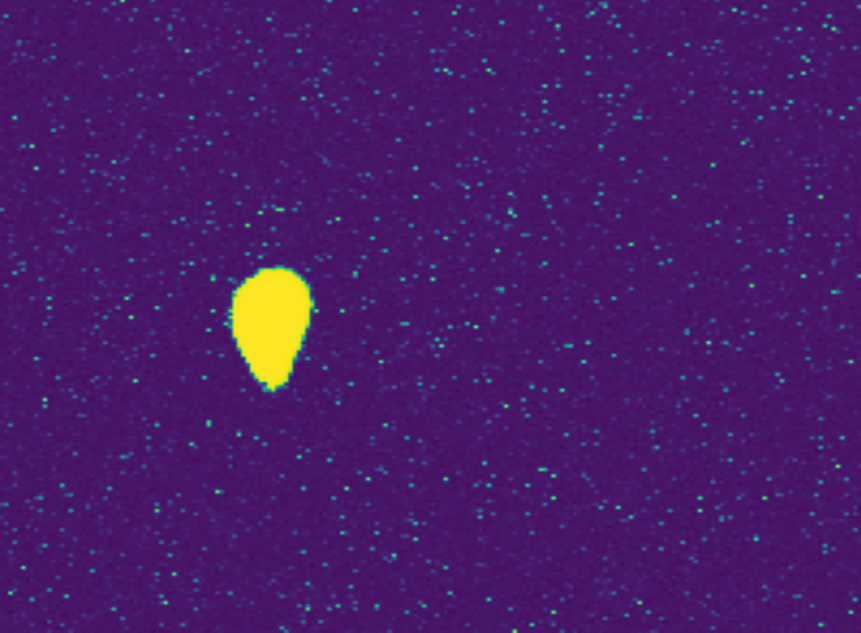};
  
    \nextgroupplot[title=$\sigma\circ f_{20}$]
    \addplot graphics[xmin=0,xmax=1,ymin=0,ymax=1] {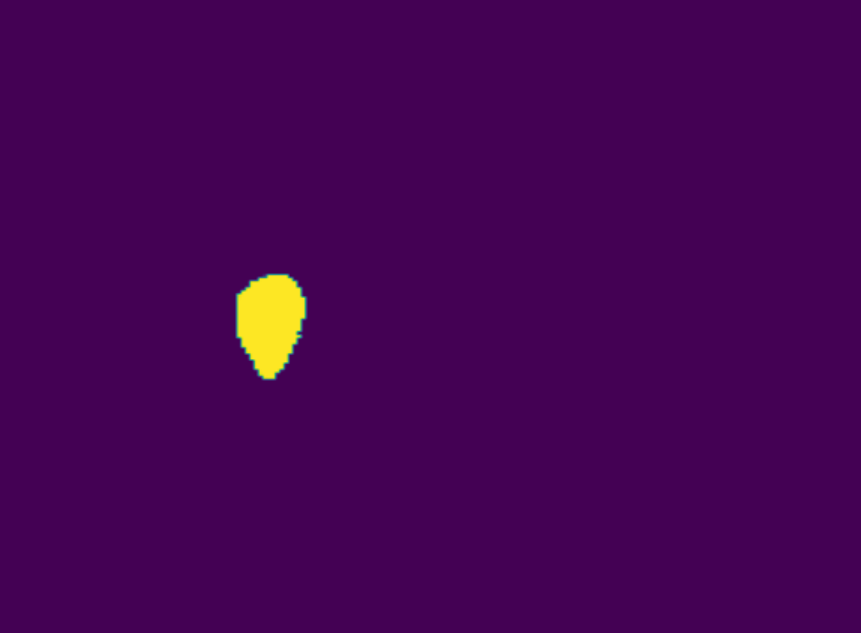};
 
    \nextgroupplot[title=$\sigma\circ f_{100}$]
    \addplot graphics[xmin=0,xmax=1,ymin=0,ymax=1] {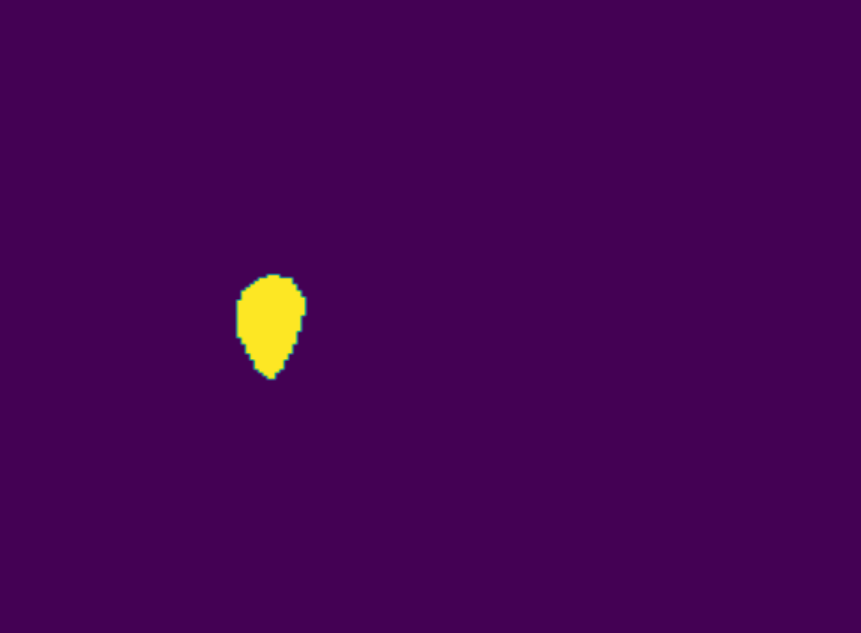};
    
    \nextgroupplot[title=$\sigma\circ f_{200}$]
    \addplot graphics[xmin=0,xmax=1,ymin=0,ymax=1] {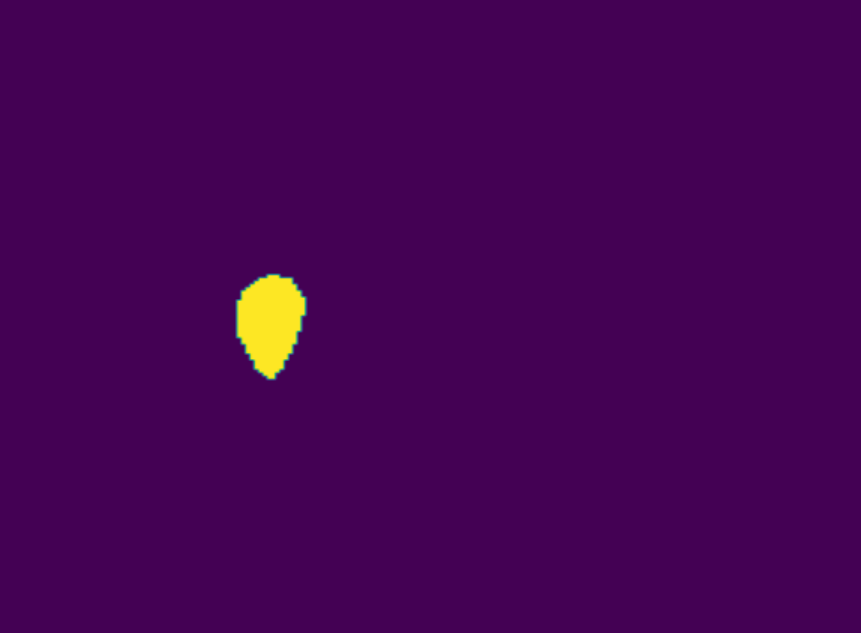};

    \nextgroupplot[title=$s$]
    \addplot graphics[xmin=0,xmax=1,ymin=0,ymax=1] {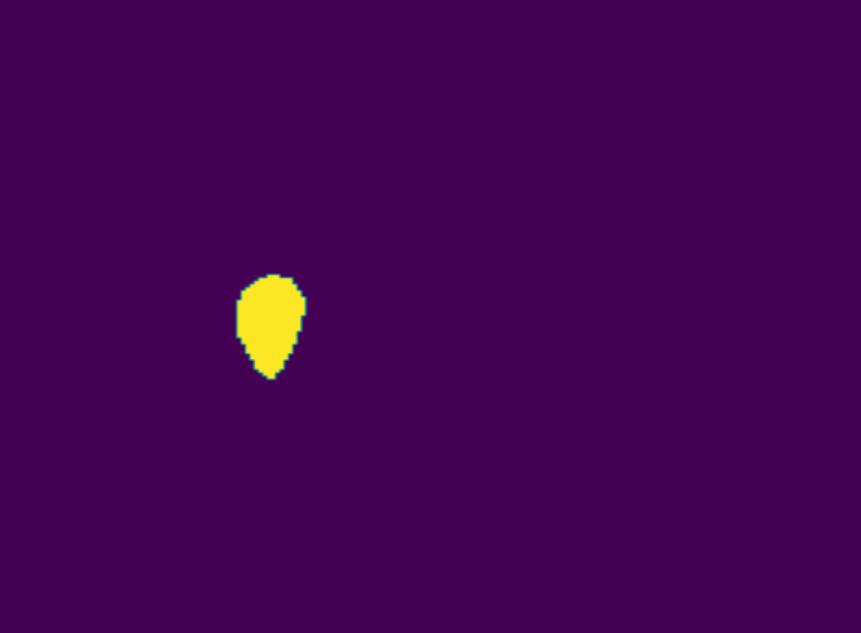};
    %%%%%%%%%%%%%%%%%%
    
    \nextgroupplot[ylabel={$\rho=0.02$} ]
    \addplot graphics[xmin=0,xmax=1,ymin=0,ymax=1] {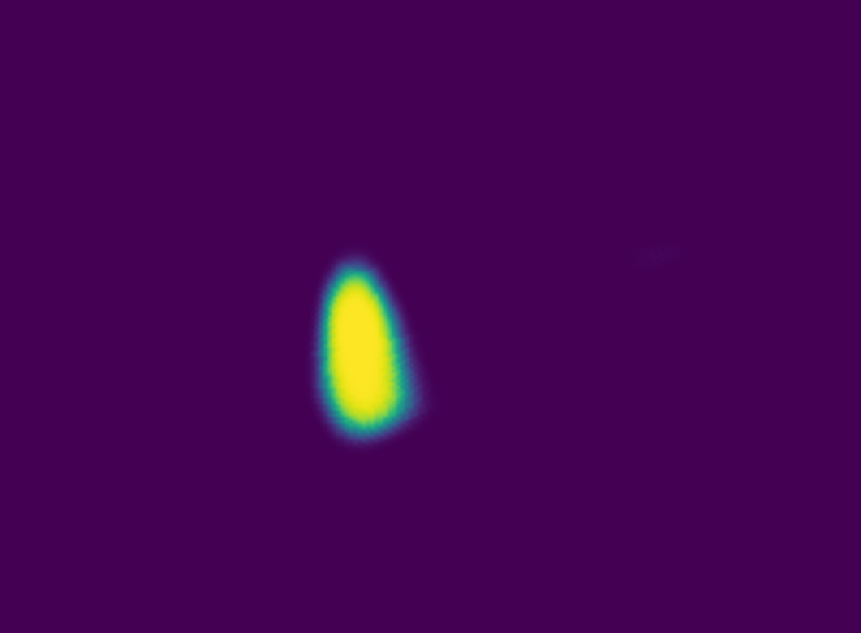};
    
    \nextgroupplot
    \addplot graphics[xmin=0,xmax=1,ymin=0,ymax=1] {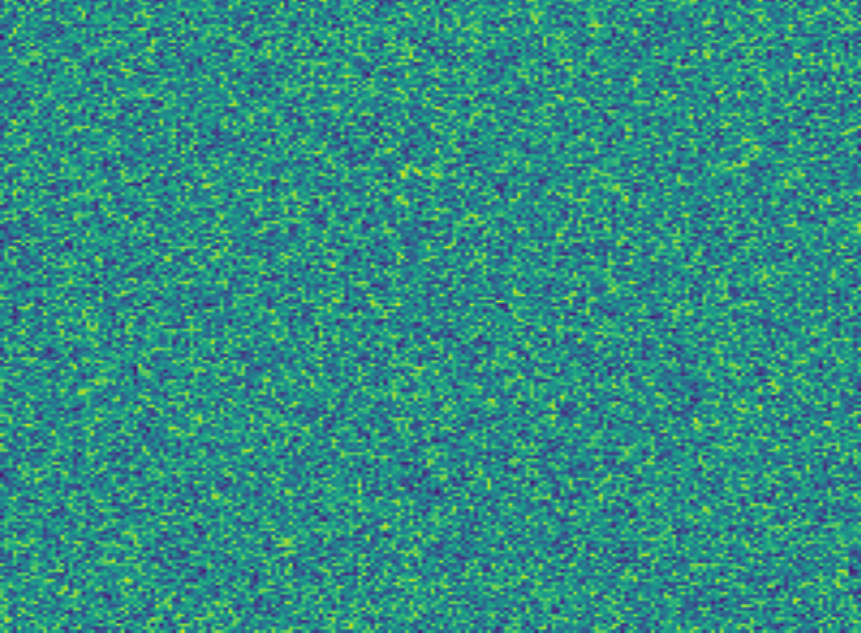};
    
    \nextgroupplot
    \addplot graphics[xmin=0,xmax=1,ymin=0,ymax=1] {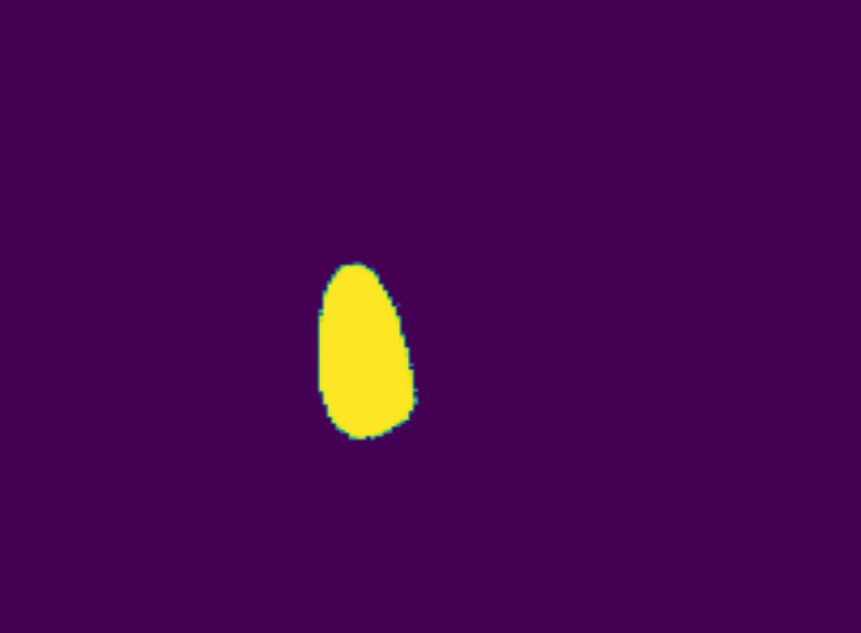};
  
    \nextgroupplot
    \addplot graphics[xmin=0,xmax=1,ymin=0,ymax=1] {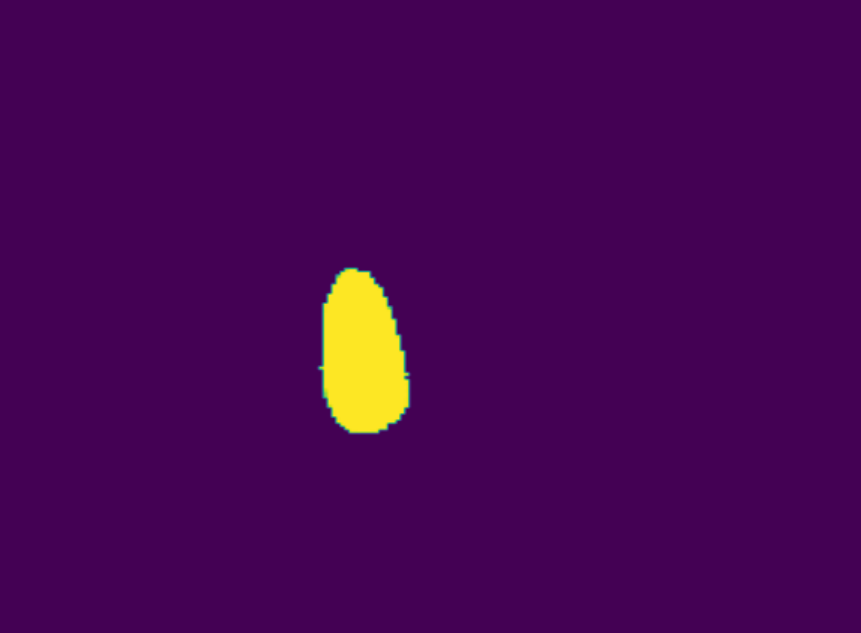};
 
    \nextgroupplot
    \addplot graphics[xmin=0,xmax=1,ymin=0,ymax=1] {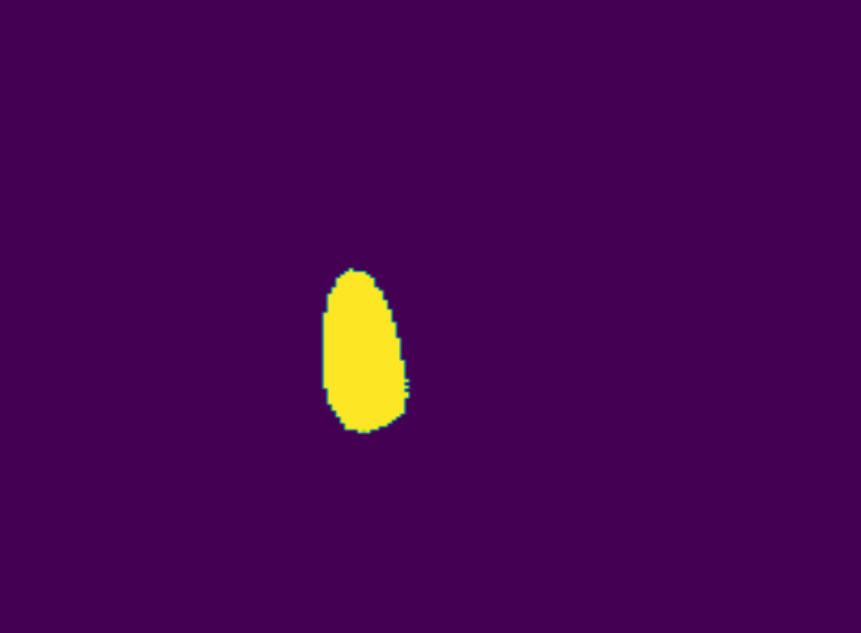};
    
    \nextgroupplot
    \addplot graphics[xmin=0,xmax=1,ymin=0,ymax=1] {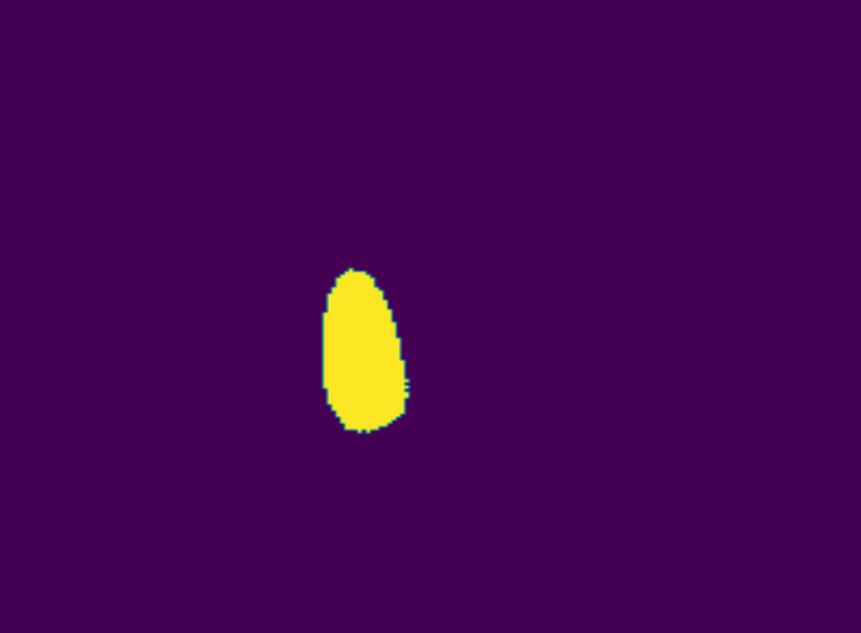};
    
    \nextgroupplot
    \addplot graphics[xmin=0,xmax=1,ymin=0,ymax=1] {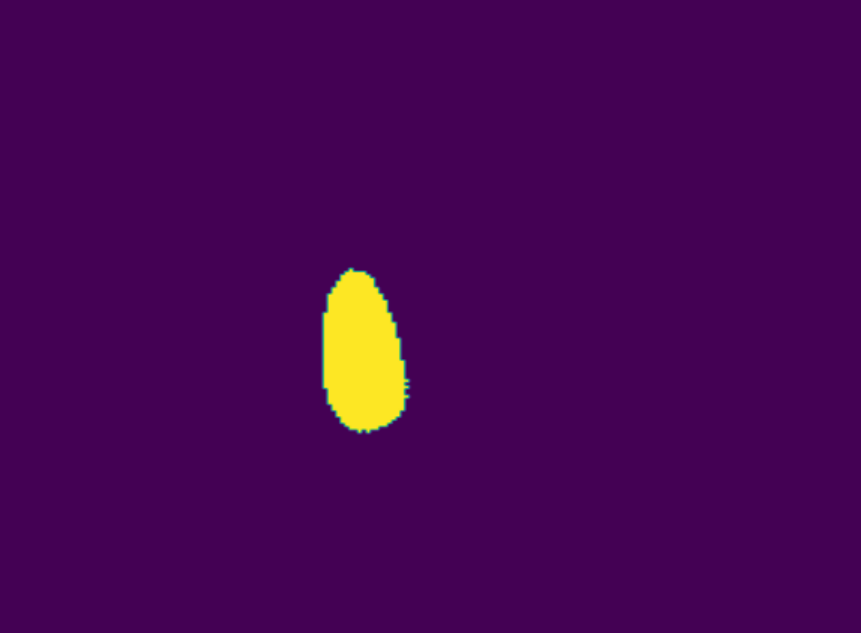};
    %%%%%%%%%%%%%%%%%%%%
    
    \nextgroupplot[ylabel={$\rho=0.03$}]
    \addplot graphics[xmin=0,xmax=1,ymin=0,ymax=1] {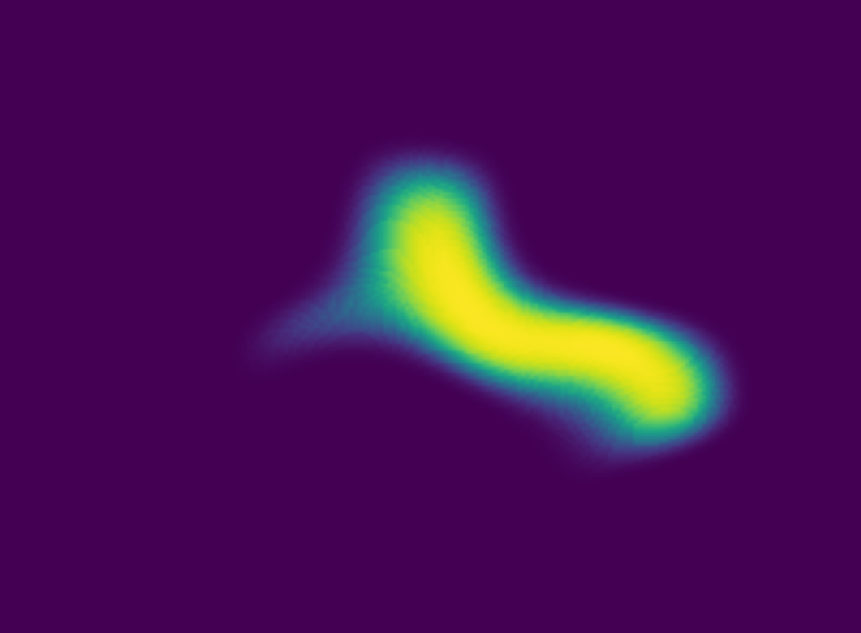};
    
    \nextgroupplot
    \addplot graphics[xmin=0,xmax=1,ymin=0,ymax=1] {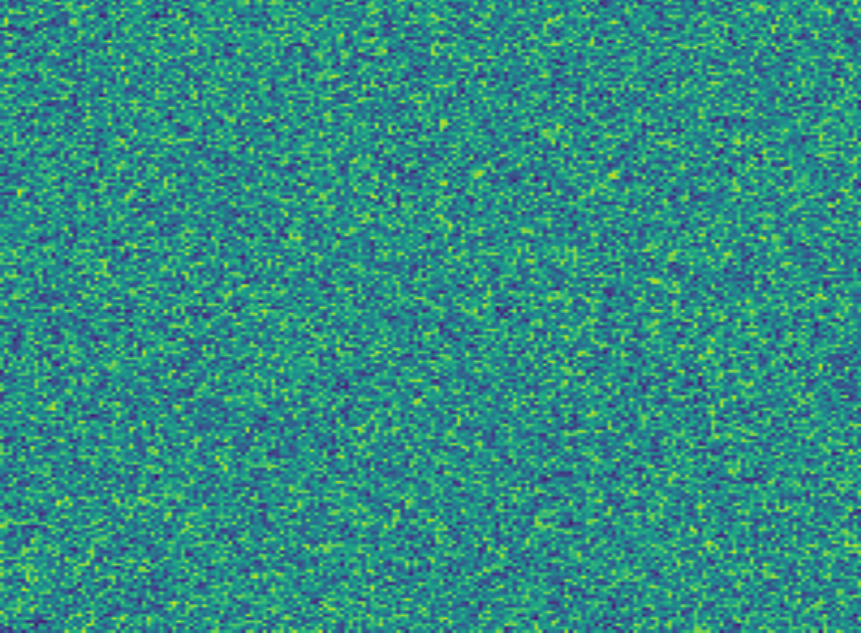};
    
    \nextgroupplot
    \addplot graphics[xmin=0,xmax=1,ymin=0,ymax=1] {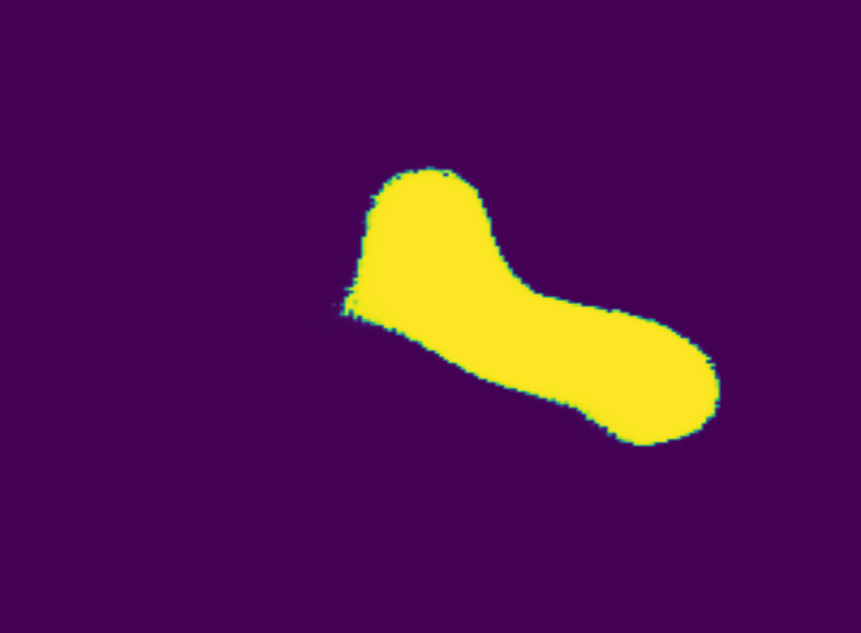};
  
    \nextgroupplot
    \addplot graphics[xmin=0,xmax=1,ymin=0,ymax=1] {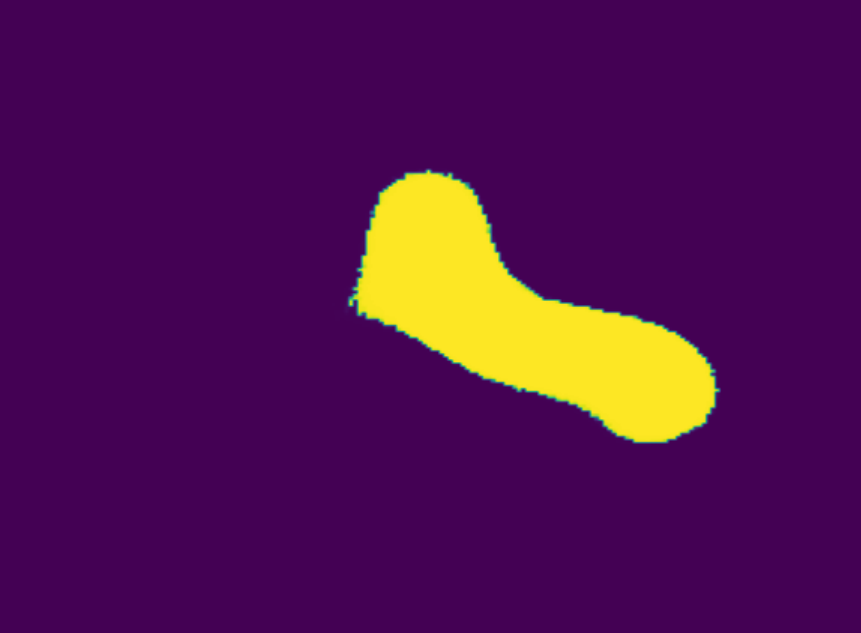};
 
    \nextgroupplot
    \addplot graphics[xmin=0,xmax=1,ymin=0,ymax=1] {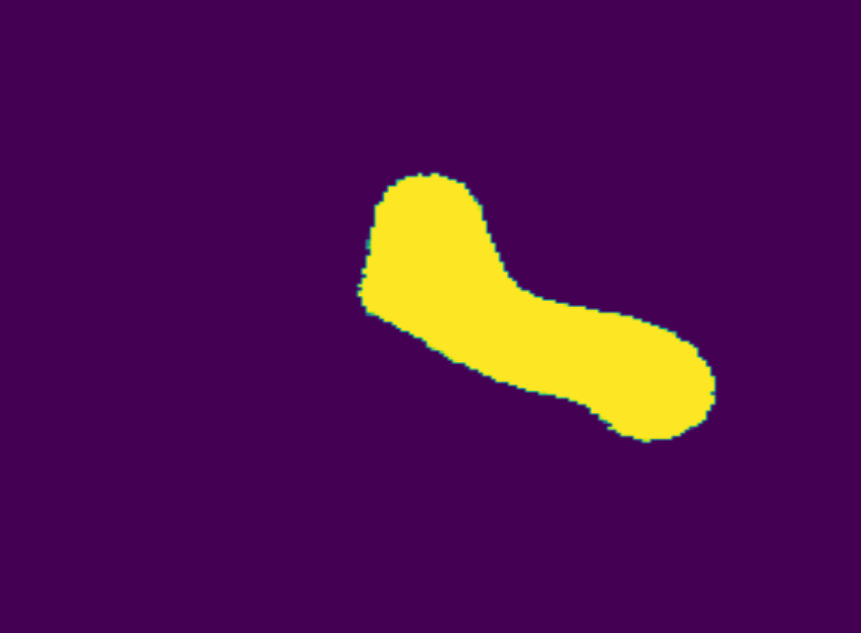};
    
    \nextgroupplot
    \addplot graphics[xmin=0,xmax=1,ymin=0,ymax=1] {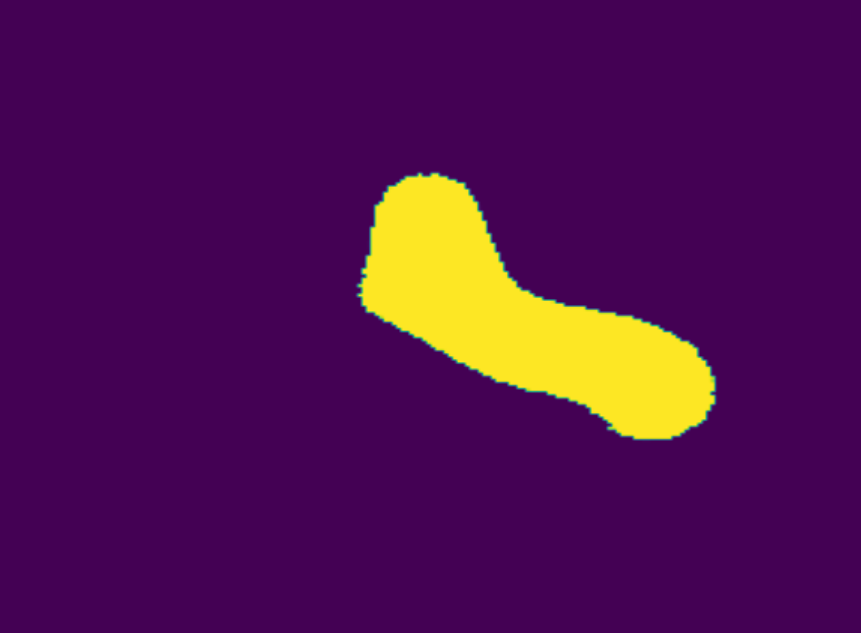};
    
    \nextgroupplot
    \addplot graphics[xmin=0,xmax=1,ymin=0,ymax=1] {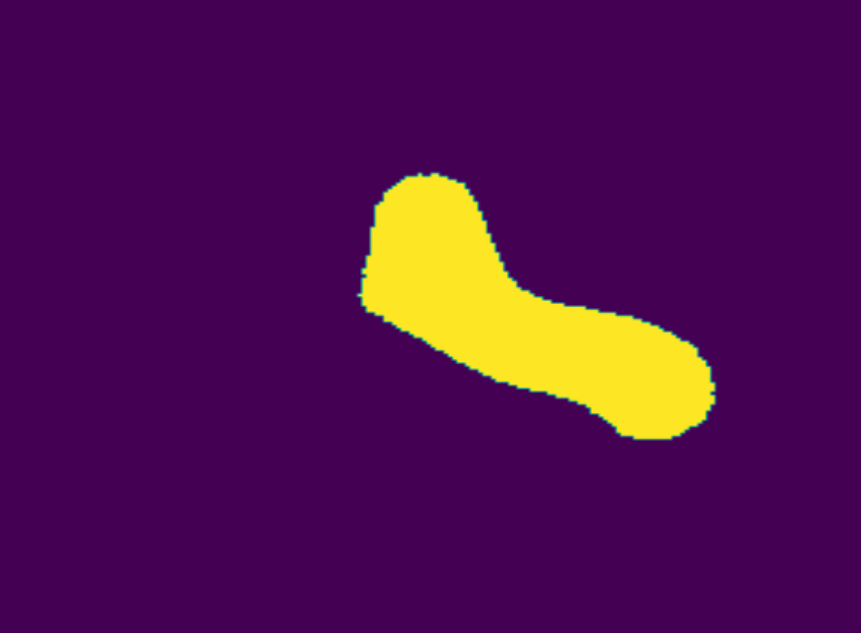};
    
    %%%%%%%%%%%%%%%%%%%%
    
    \nextgroupplot[ylabel={$\rho=0.04$}]
    \addplot graphics[xmin=0,xmax=1,ymin=0,ymax=1] {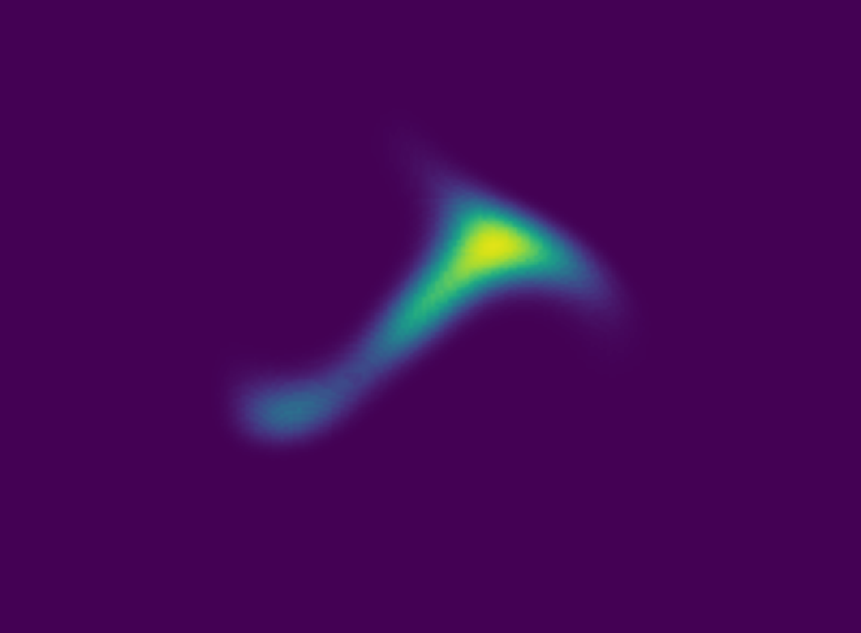};
    
    \nextgroupplot
    \addplot graphics[xmin=0,xmax=1,ymin=0,ymax=1] {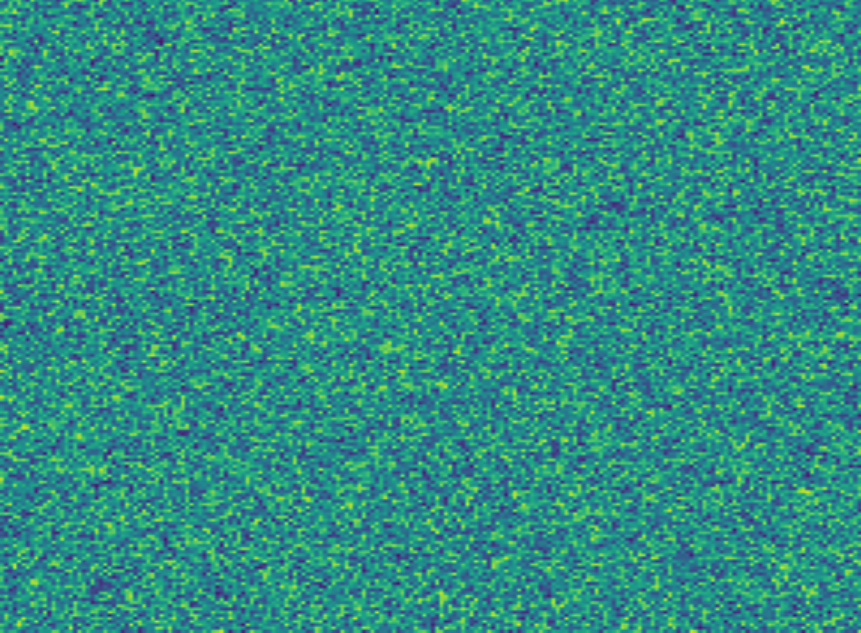};
    
    \nextgroupplot
    \addplot graphics[xmin=0,xmax=1,ymin=0,ymax=1] {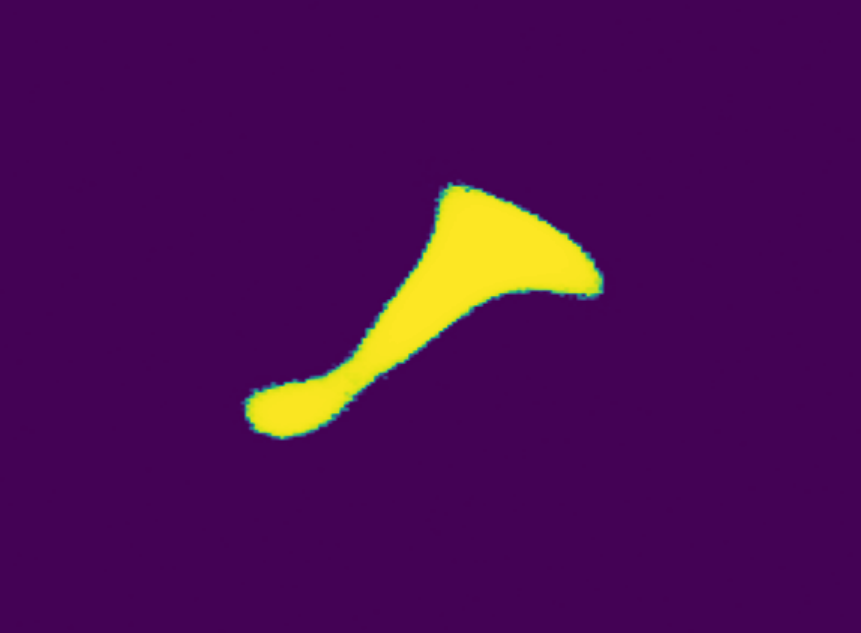};
  
    \nextgroupplot
    \addplot graphics[xmin=0,xmax=1,ymin=0,ymax=1] {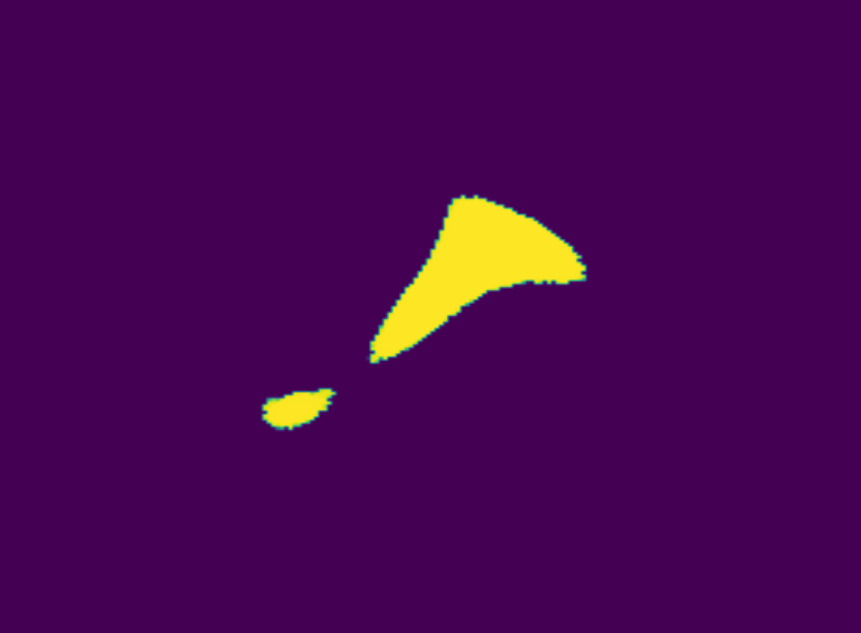};
 
    \nextgroupplot
    \addplot graphics[xmin=0,xmax=1,ymin=0,ymax=1] {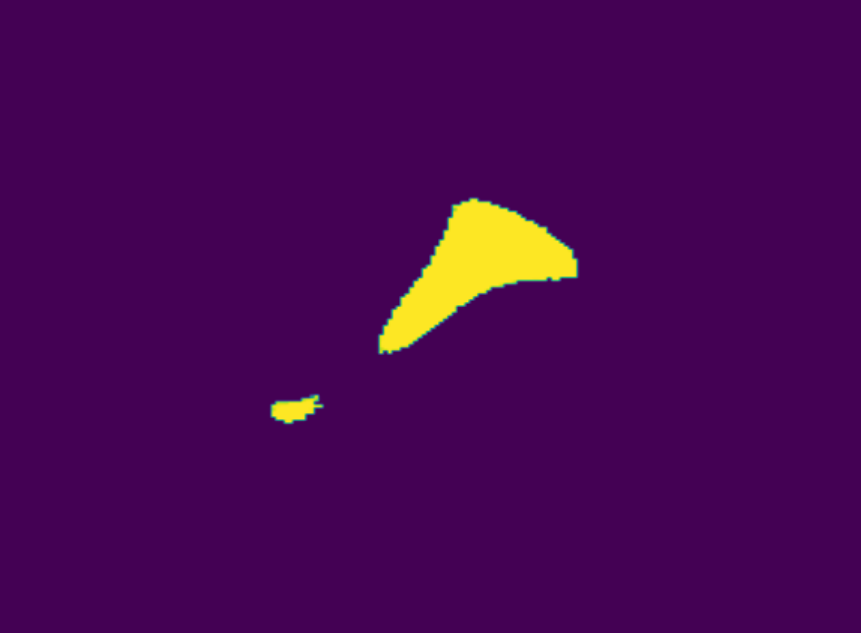};
    
    \nextgroupplot
    \addplot graphics[xmin=0,xmax=1,ymin=0,ymax=1] {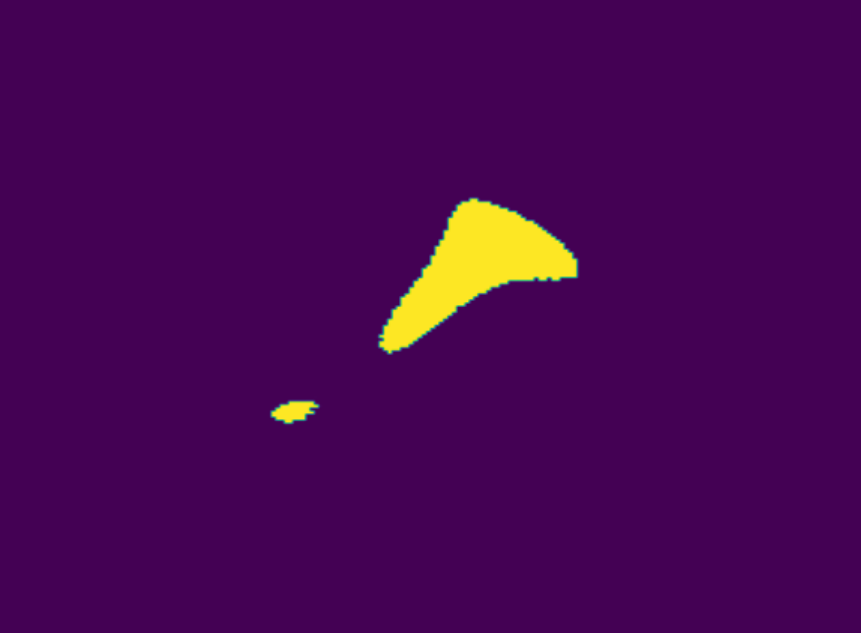};
    
    \nextgroupplot
    \addplot graphics[xmin=0,xmax=1,ymin=0,ymax=1] {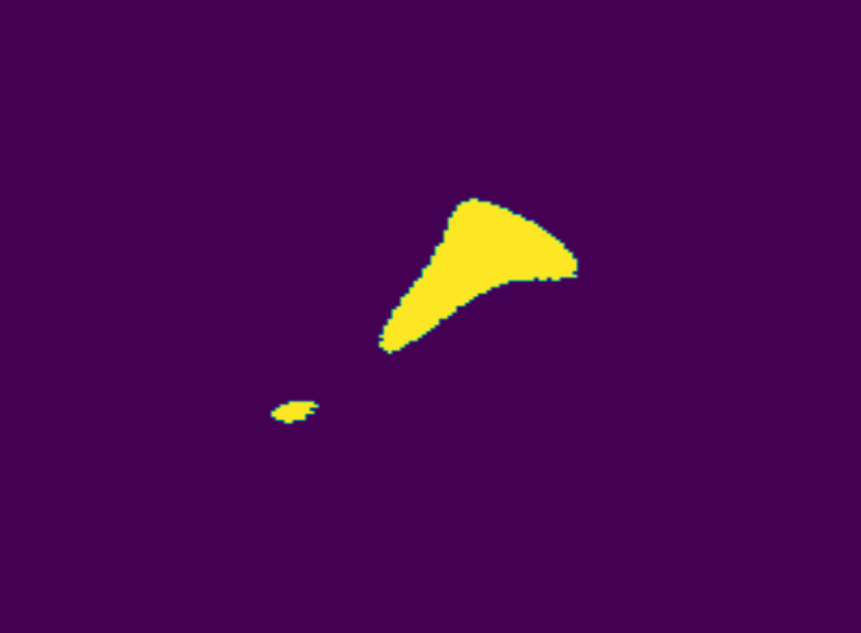};
    
    %%%%%%%%%%%%%%%%%%%%
    
    \nextgroupplot[ylabel={$\rho=0.05$}]
    \addplot graphics[xmin=0,xmax=1,ymin=0,ymax=1] {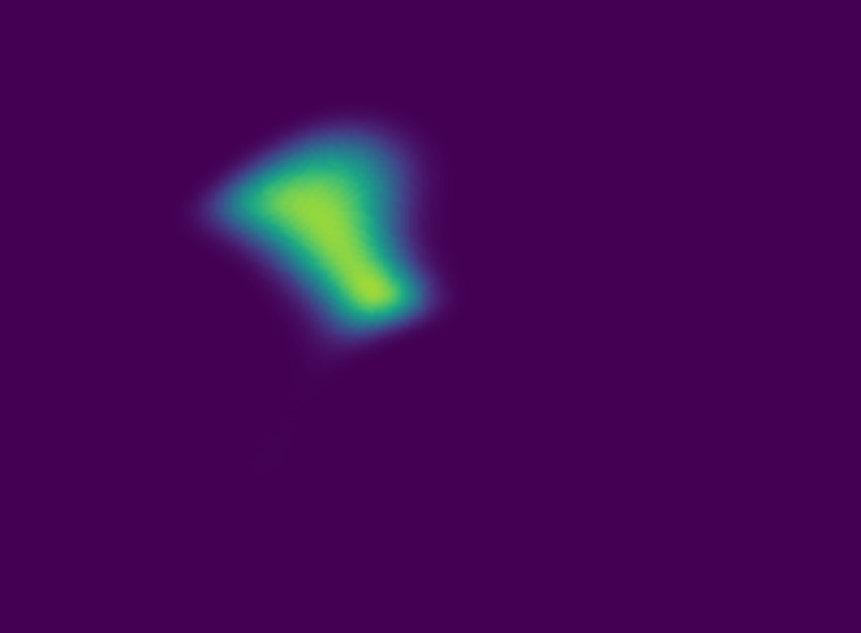};
    
    \nextgroupplot
    \addplot graphics[xmin=0,xmax=1,ymin=0,ymax=1] {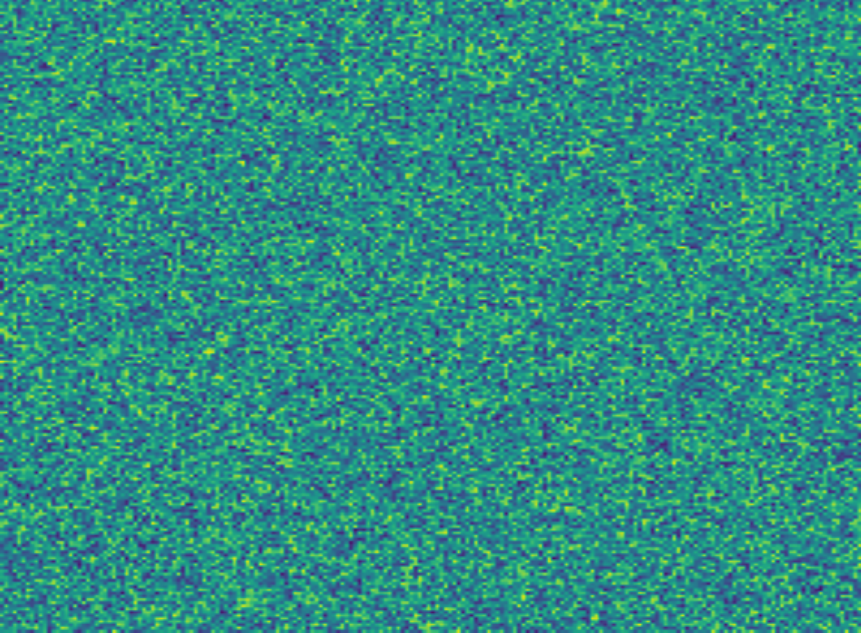};
    
    \nextgroupplot
    \addplot graphics[xmin=0,xmax=1,ymin=0,ymax=1] {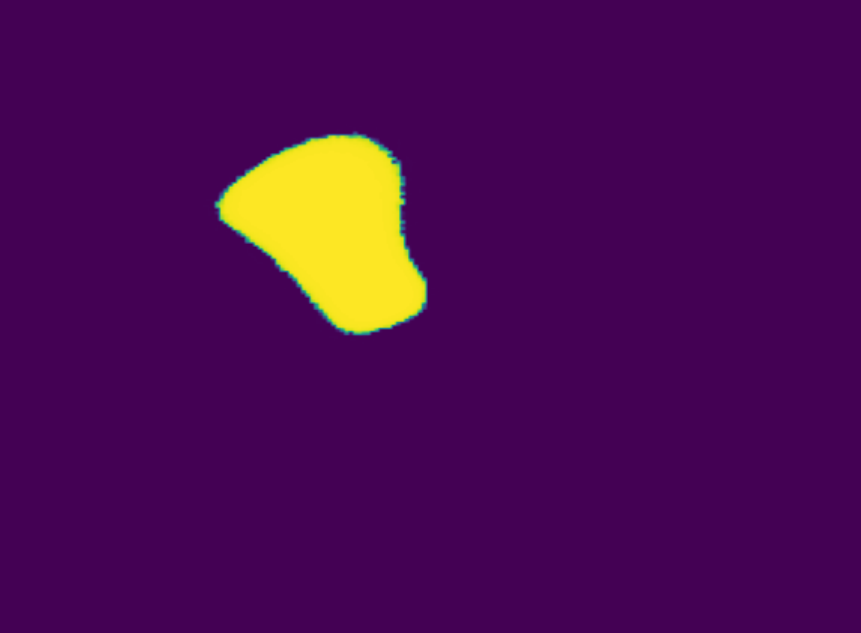};
  
    \nextgroupplot
    \addplot graphics[xmin=0,xmax=1,ymin=0,ymax=1] {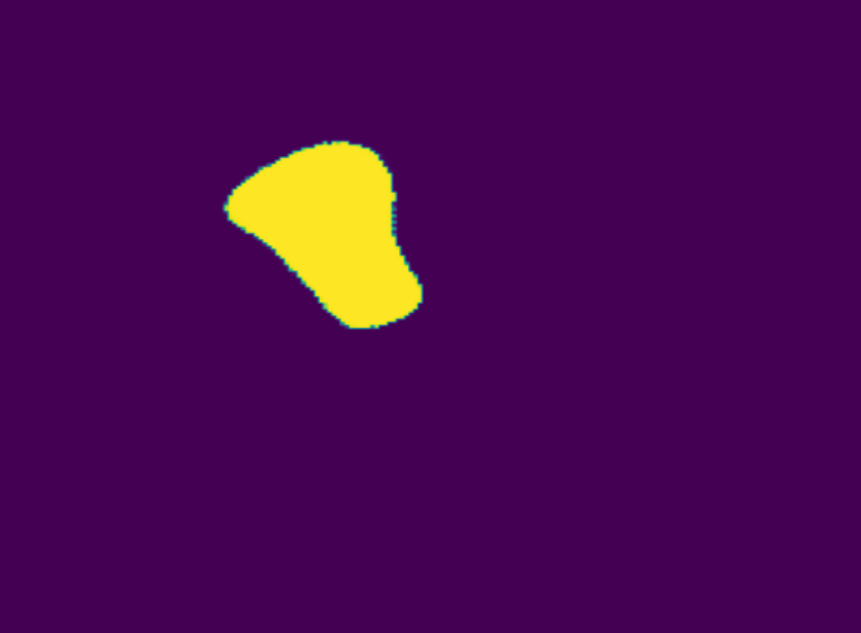};
 
    \nextgroupplot
    \addplot graphics[xmin=0,xmax=1,ymin=0,ymax=1] {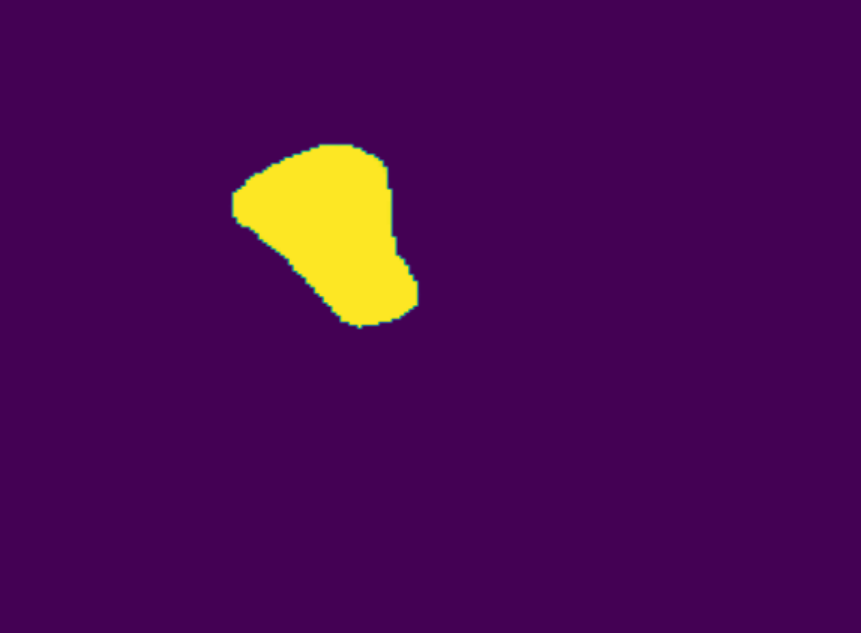};
    
    \nextgroupplot
    \addplot graphics[xmin=0,xmax=1,ymin=0,ymax=1] {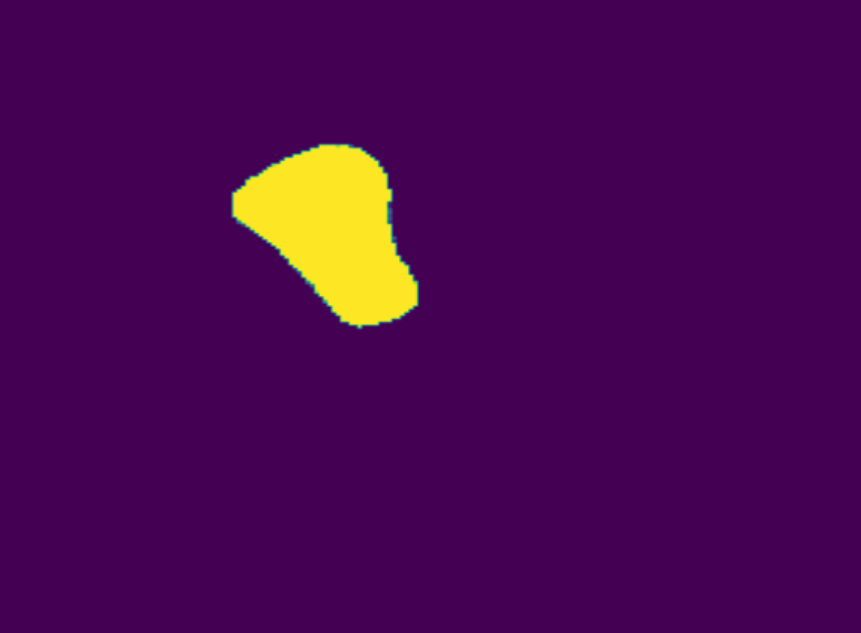};
    
    \nextgroupplot
    \addplot graphics[xmin=0,xmax=1,ymin=0,ymax=1] {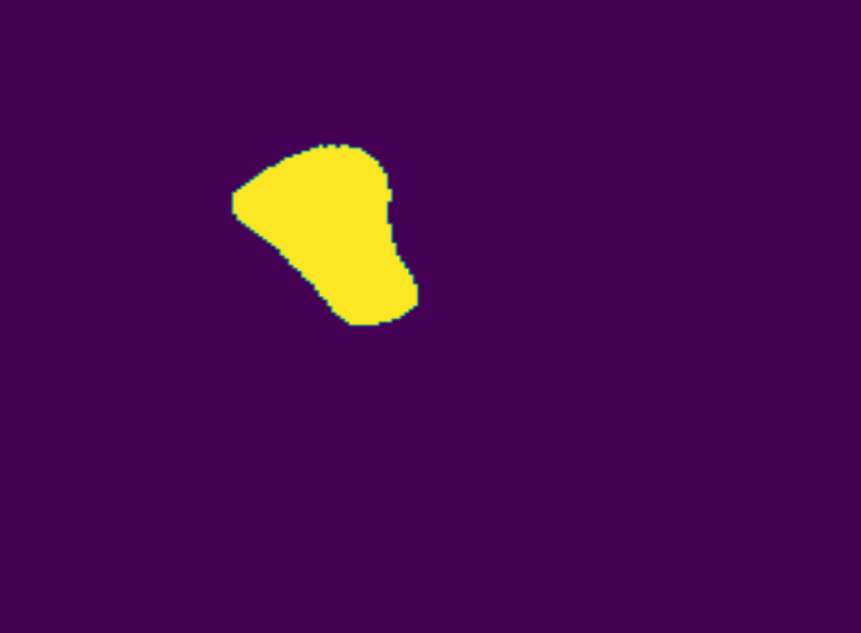};
    
    %%%%%%%%%%%%%%%%%%%%
    
    \nextgroupplot[ylabel={$\rho=0.06$} ]
    \addplot graphics[xmin=0,xmax=1,ymin=0,ymax=1] {raw/seqS/p_5_m.pdf};
    \addplot graphics[xmin=0,xmax=1,ymin=0,ymax=1] {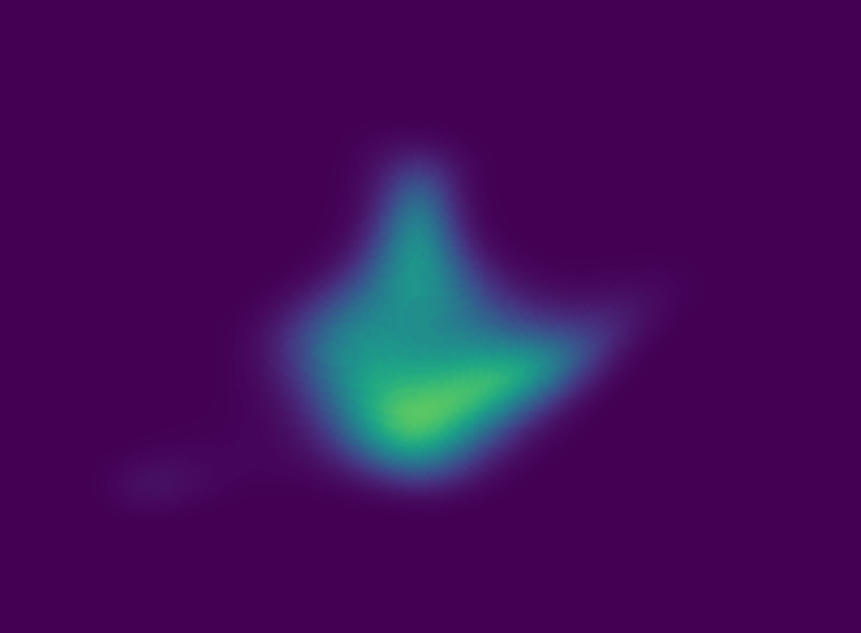};
    
    \nextgroupplot
    \addplot graphics[xmin=0,xmax=1,ymin=0,ymax=1] {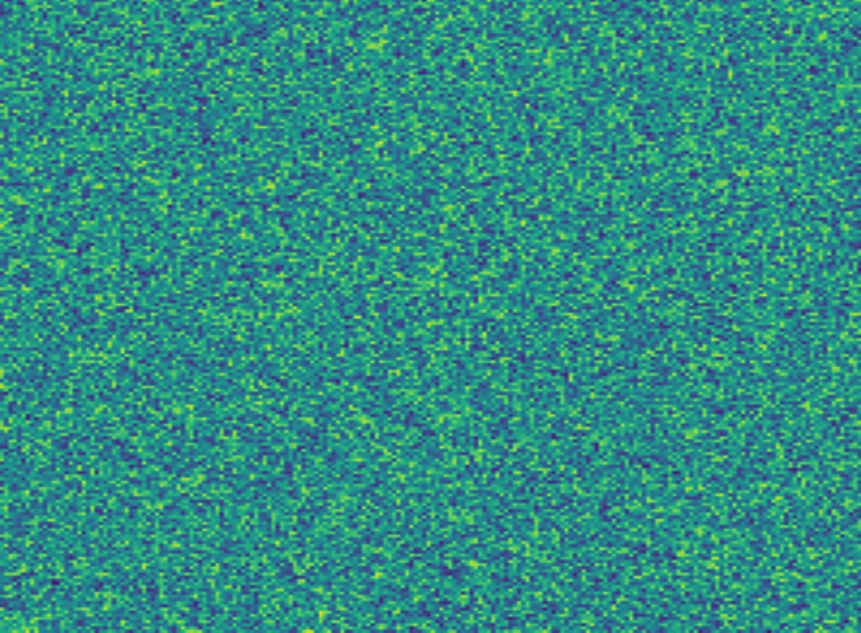};
    
    \nextgroupplot
    \addplot graphics[xmin=0,xmax=1,ymin=0,ymax=1] {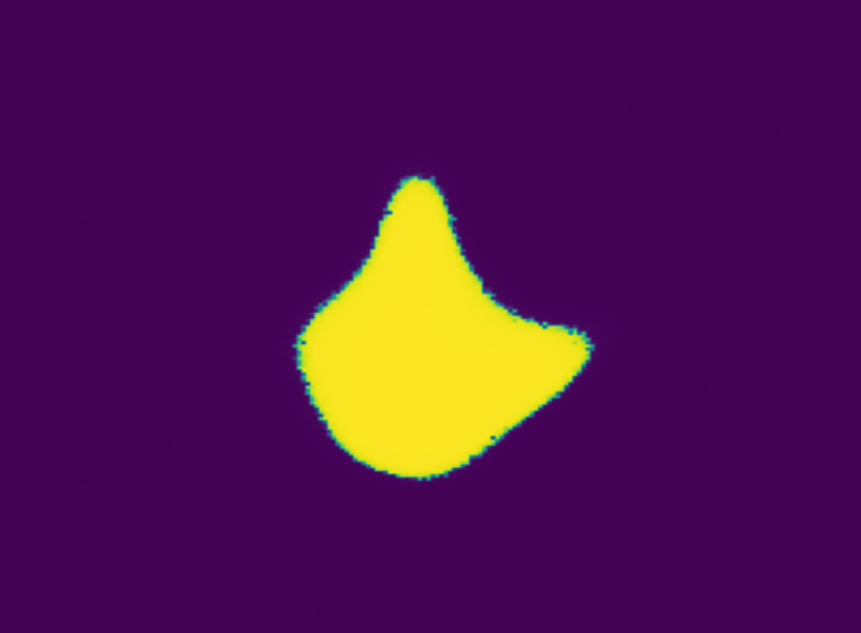};
  
    \nextgroupplot
    \addplot graphics[xmin=0,xmax=1,ymin=0,ymax=1] {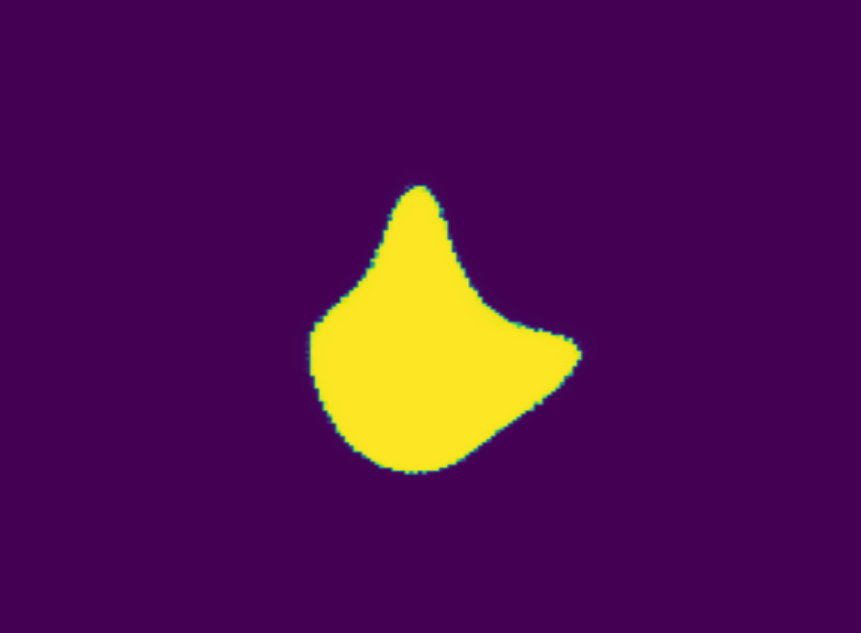};
 
    \nextgroupplot
    \addplot graphics[xmin=0,xmax=1,ymin=0,ymax=1] {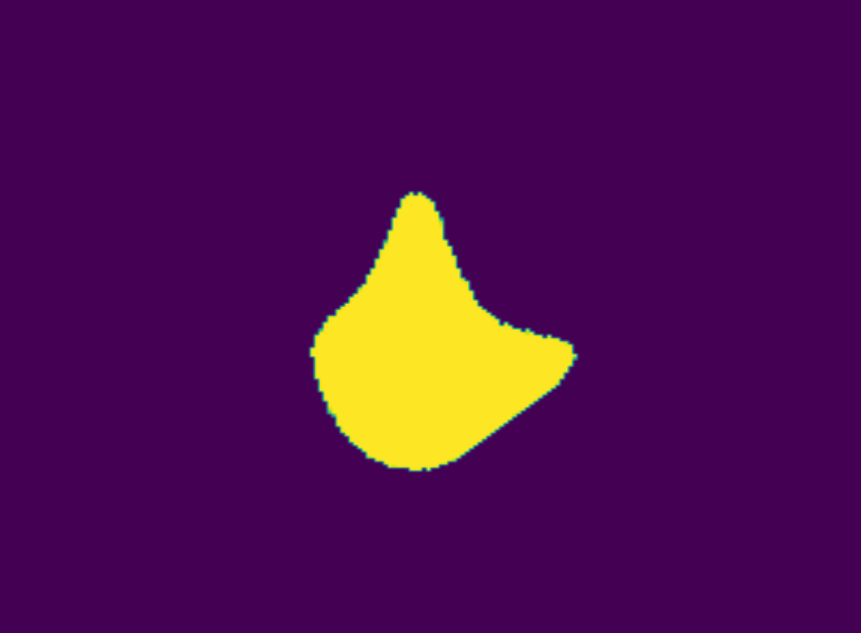};
    
    \nextgroupplot
    \addplot graphics[xmin=0,xmax=1,ymin=0,ymax=1] {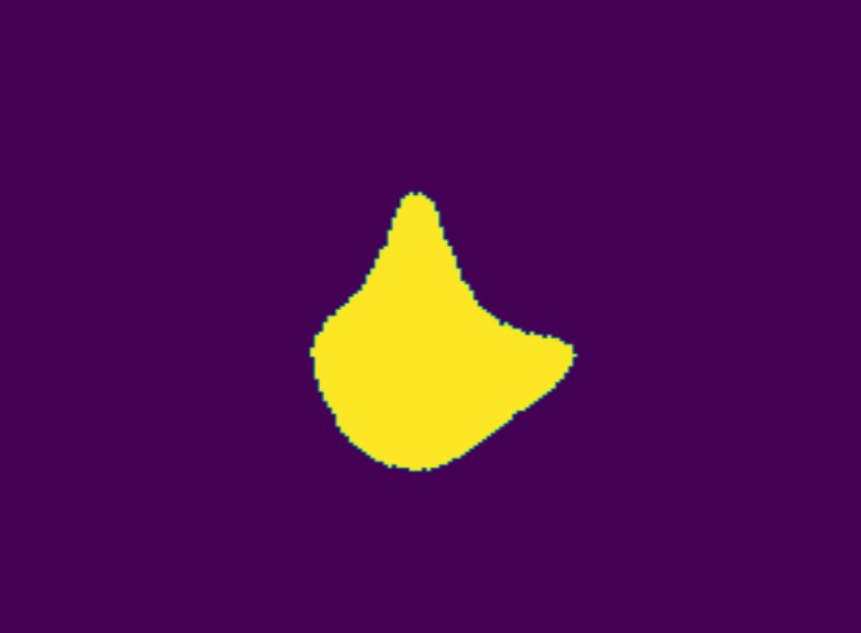};
    
    \nextgroupplot
    \addplot graphics[xmin=0,xmax=1,ymin=0,ymax=1] {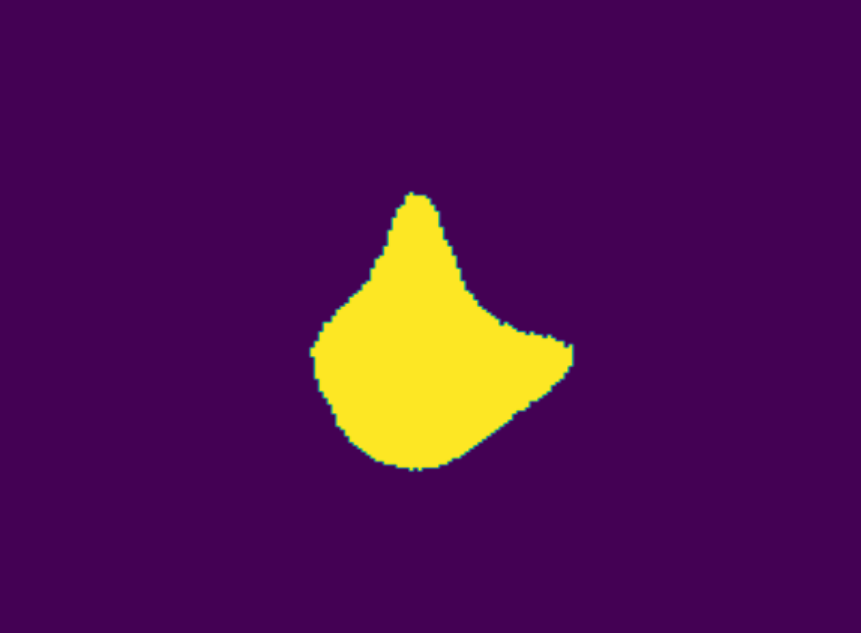};
    
    %%%%%%%%%%%%%%%%%%%%
    
    \nextgroupplot[ylabel={$\rho=0.07$} ]
    \addplot graphics[xmin=0,xmax=1,ymin=0,ymax=1] {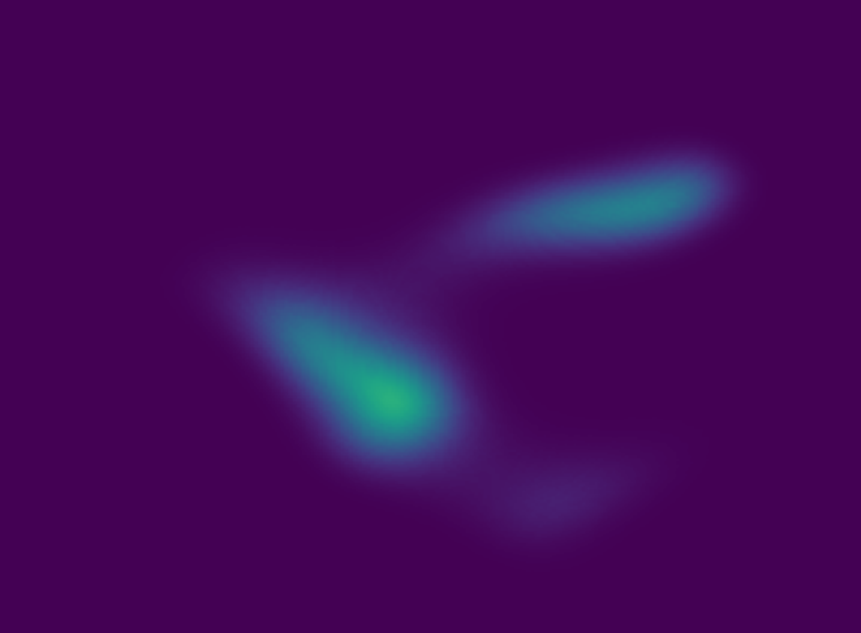};
    
    \nextgroupplot
    \addplot graphics[xmin=0,xmax=1,ymin=0,ymax=1] {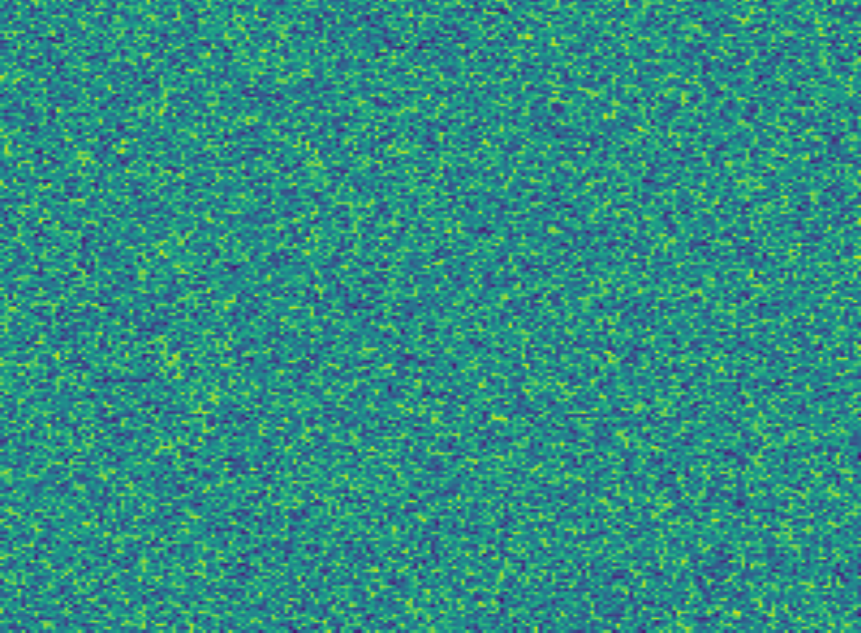};
    
    \nextgroupplot
    \addplot graphics[xmin=0,xmax=1,ymin=0,ymax=1] {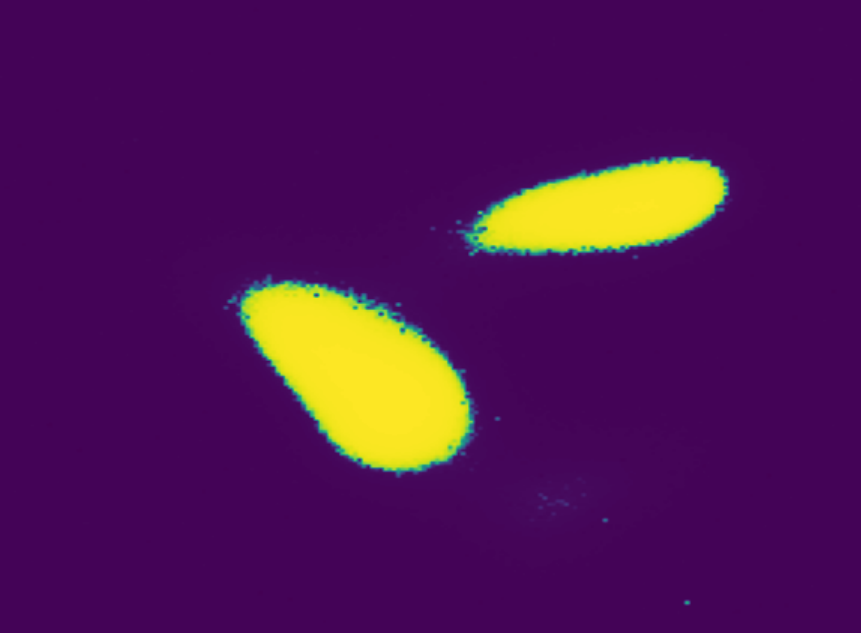};
  
    \nextgroupplot
    \addplot graphics[xmin=0,xmax=1,ymin=0,ymax=1] {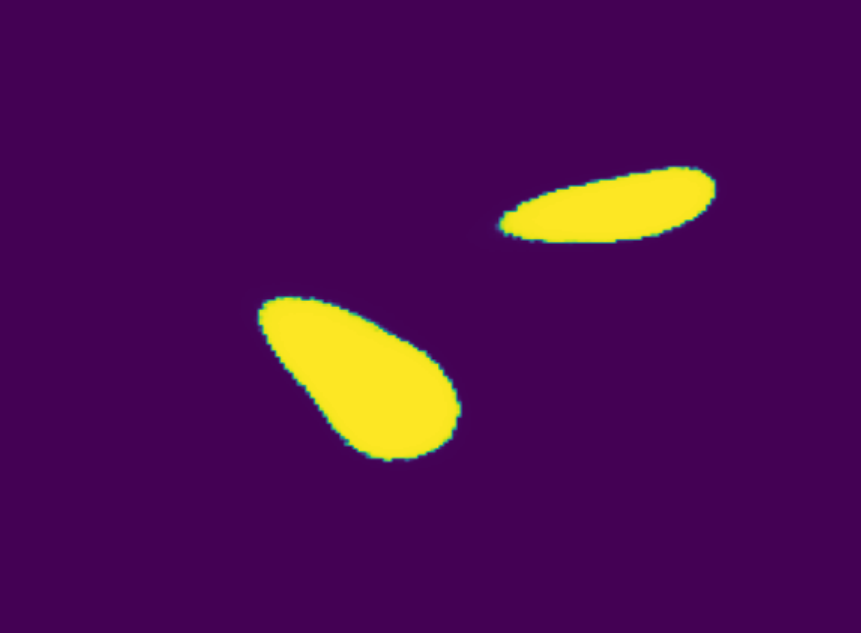};
 
    \nextgroupplot
    \addplot graphics[xmin=0,xmax=1,ymin=0,ymax=1] {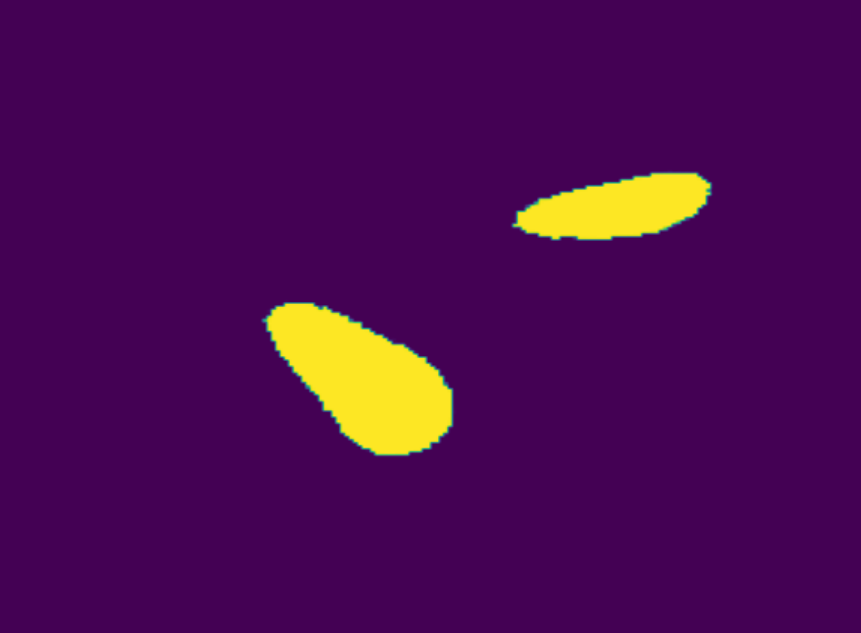};
    
    \nextgroupplot
    \addplot graphics[xmin=0,xmax=1,ymin=0,ymax=1] {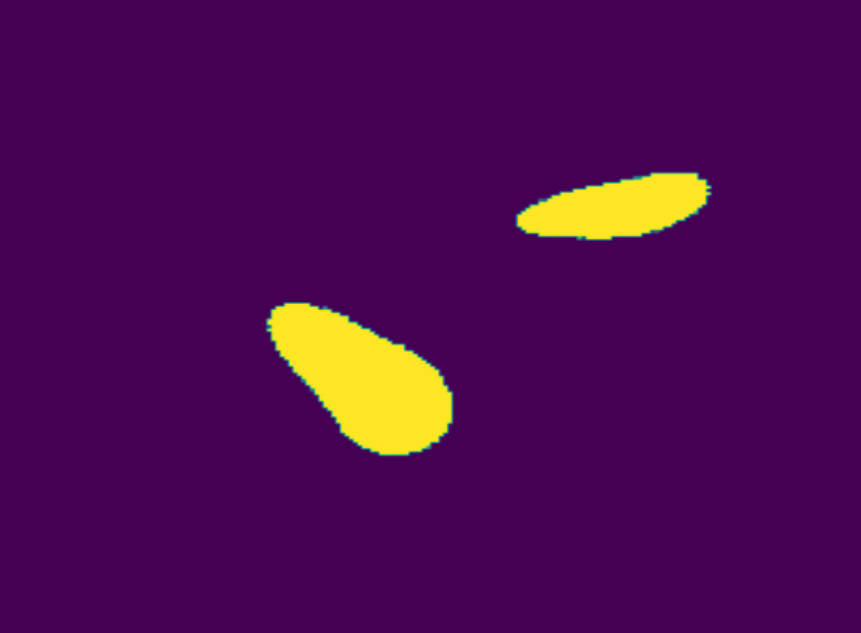};
    
    \nextgroupplot
    \addplot graphics[xmin=0,xmax=1,ymin=0,ymax=1] {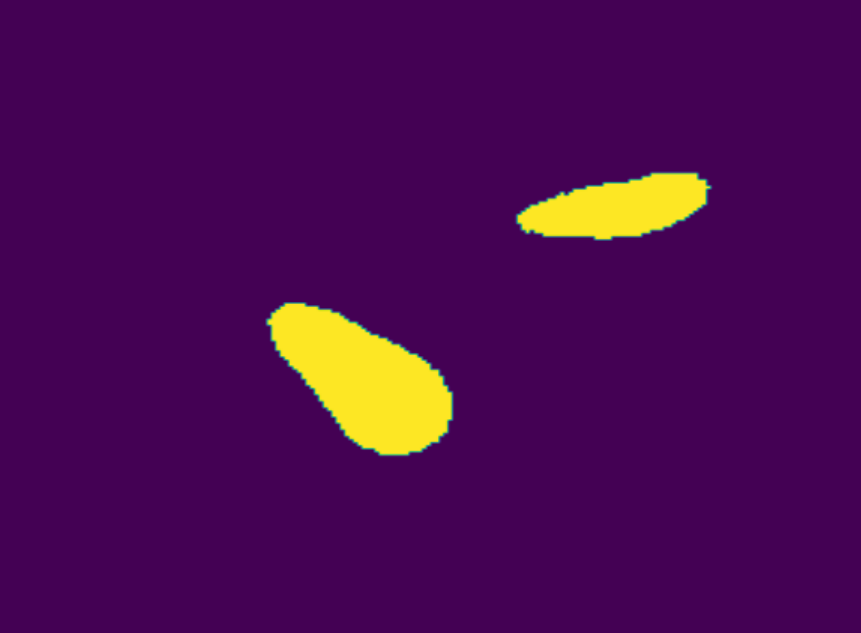};
    
    %%%%%%%%%%%%%%%%%%%%
    
    \nextgroupplot[ylabel={$\rho=0.08$} ]
    \addplot graphics[xmin=0,xmax=1,ymin=0,ymax=1] {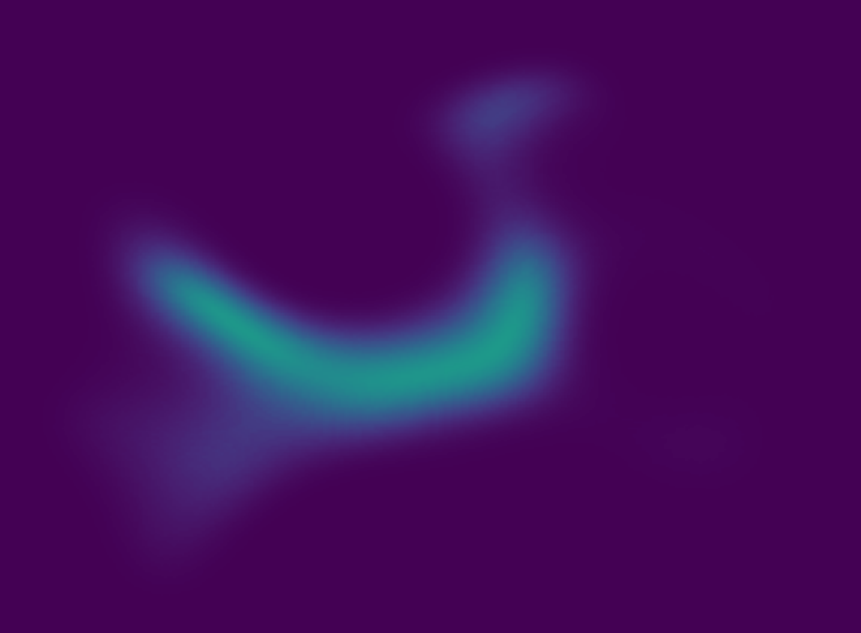};
    
    \nextgroupplot
    \addplot graphics[xmin=0,xmax=1,ymin=0,ymax=1] {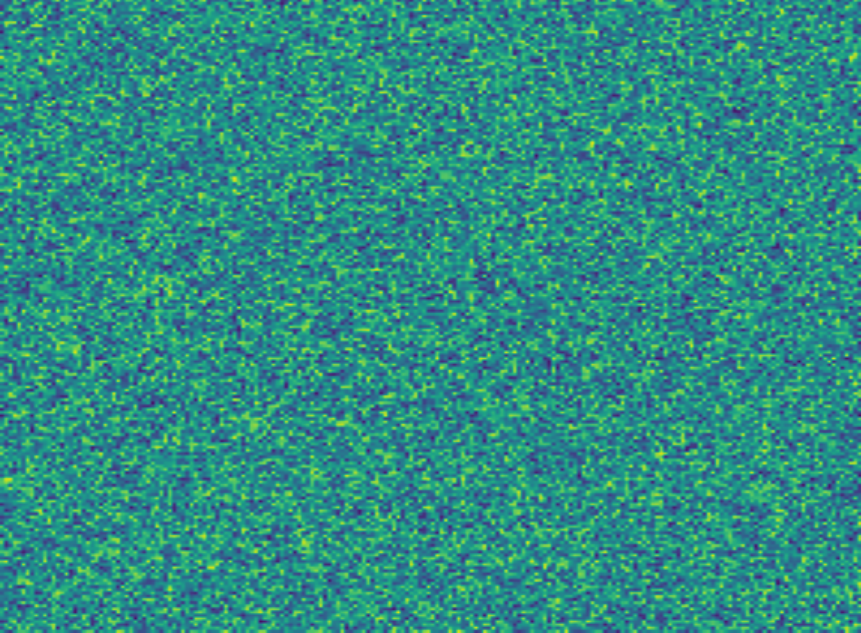};
    
    \nextgroupplot
    \addplot graphics[xmin=0,xmax=1,ymin=0,ymax=1] {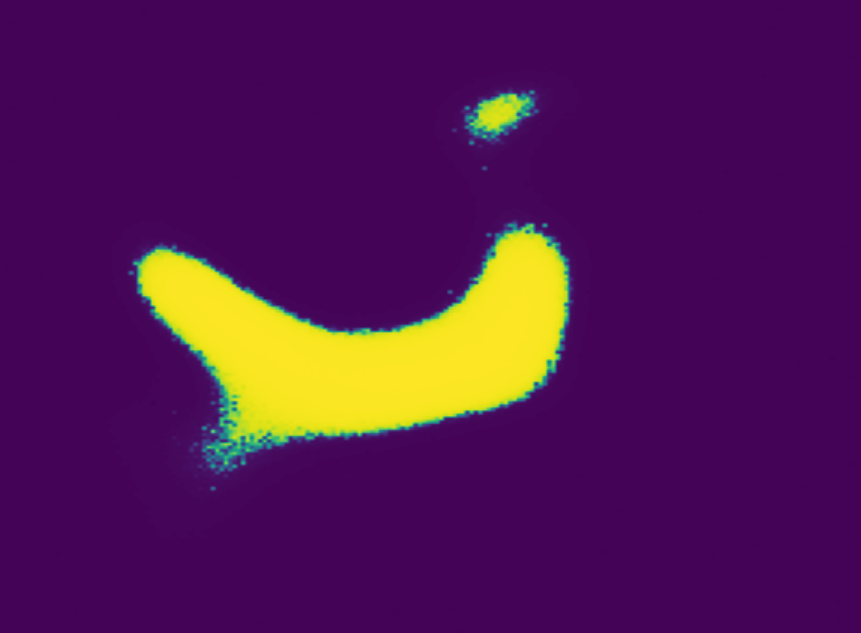};
  
    \nextgroupplot
    \addplot graphics[xmin=0,xmax=1,ymin=0,ymax=1] {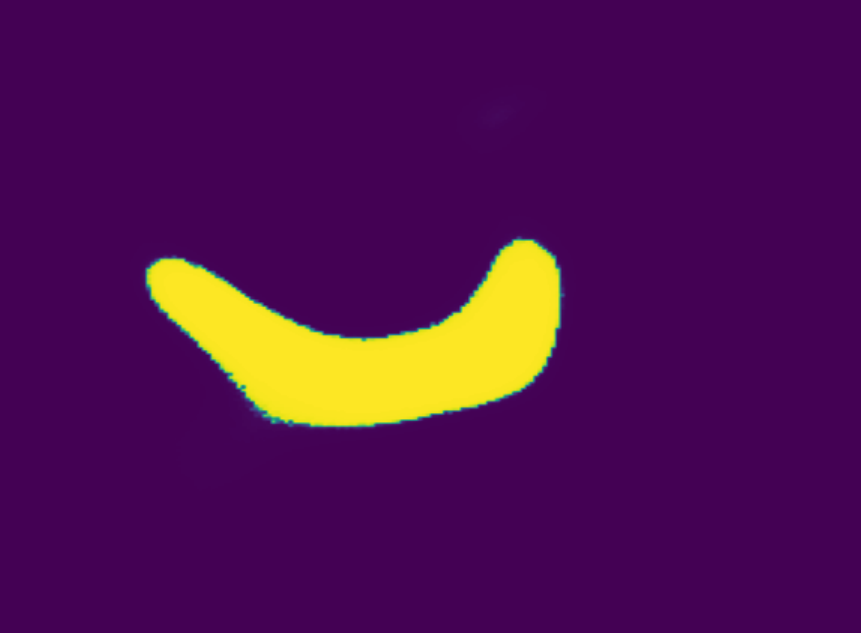};
 
    \nextgroupplot
    \addplot graphics[xmin=0,xmax=1,ymin=0,ymax=1] {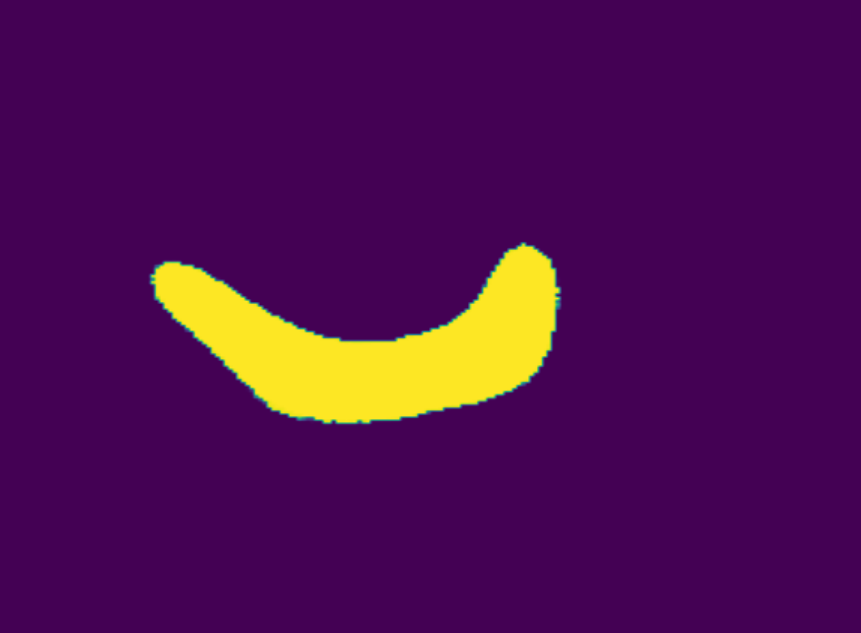};
    
    \nextgroupplot
    \addplot graphics[xmin=0,xmax=1,ymin=0,ymax=1] {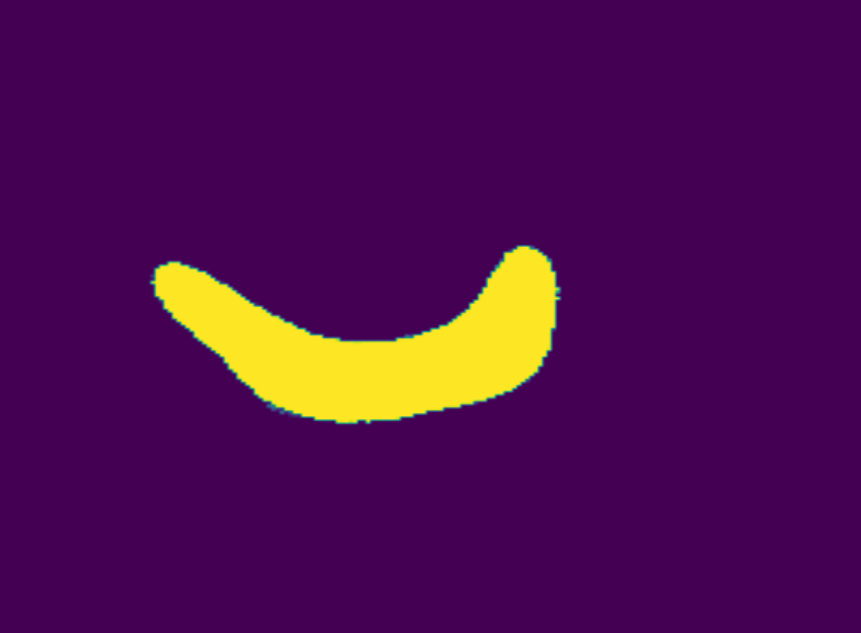};
    
    \nextgroupplot
    \addplot graphics[xmin=0,xmax=1,ymin=0,ymax=1] {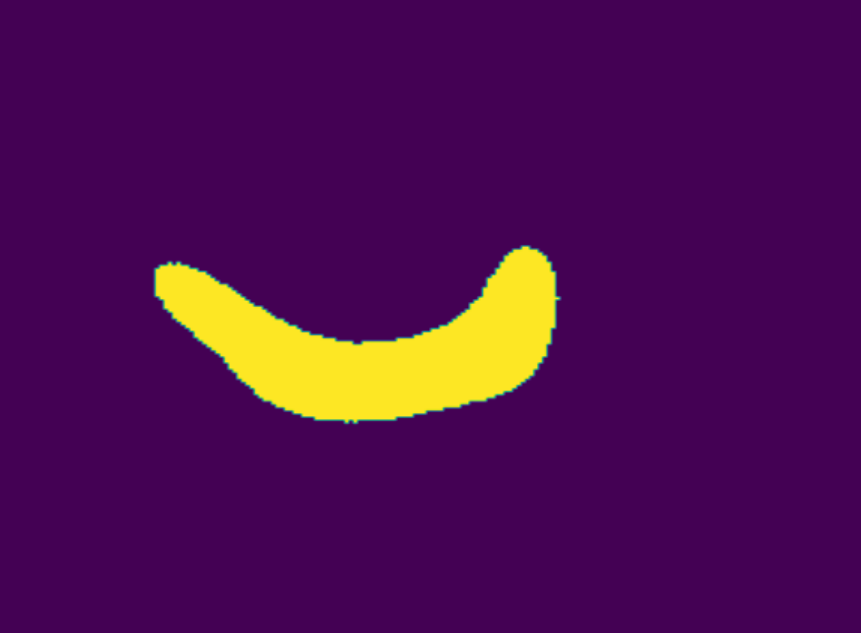};

    %%%%%%%%%%%%%%%%%%%%
    
    \nextgroupplot[ylabel={$\rho=0.09$} ]
    \addplot graphics[xmin=0,xmax=1,ymin=0,ymax=1] {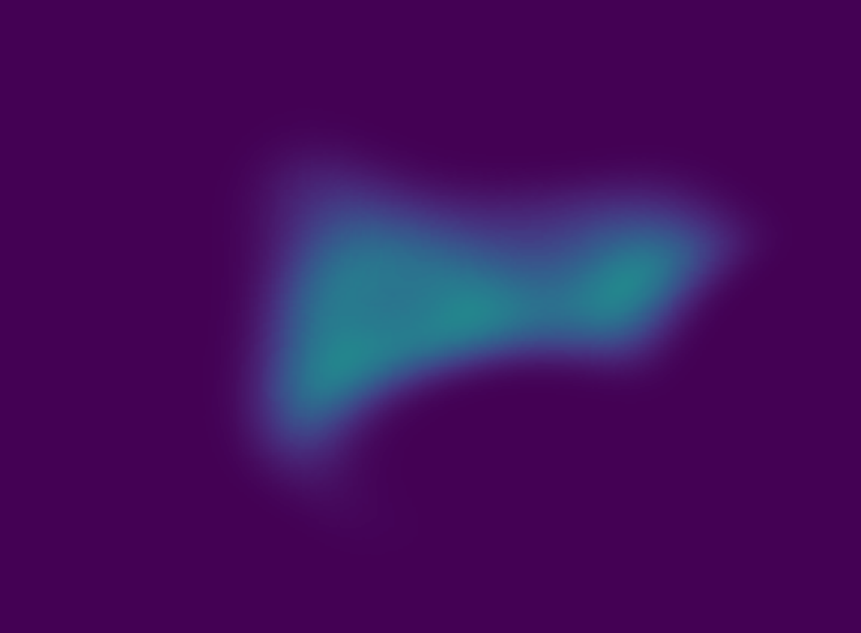};
    
    \nextgroupplot
    \addplot graphics[xmin=0,xmax=1,ymin=0,ymax=1] {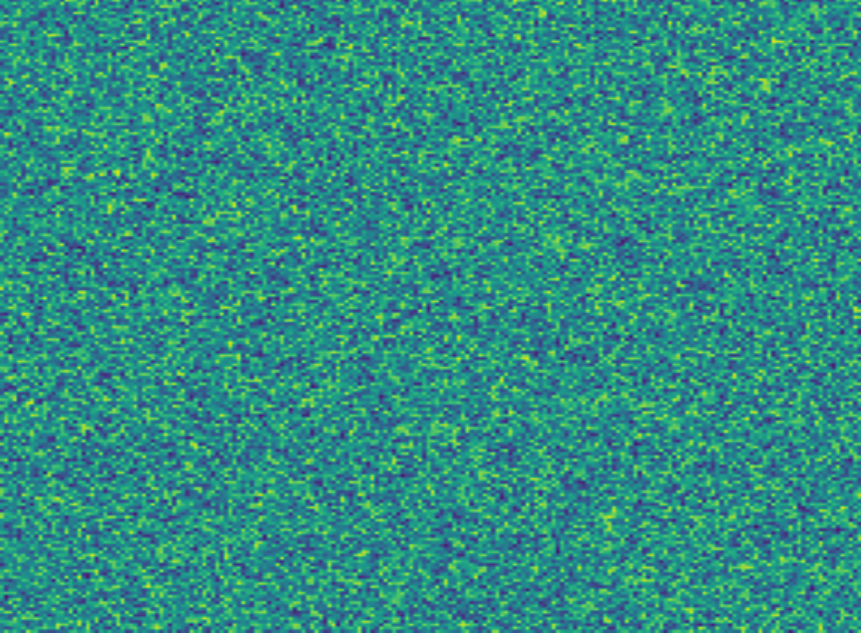};
    
    \nextgroupplot
    \addplot graphics[xmin=0,xmax=1,ymin=0,ymax=1] {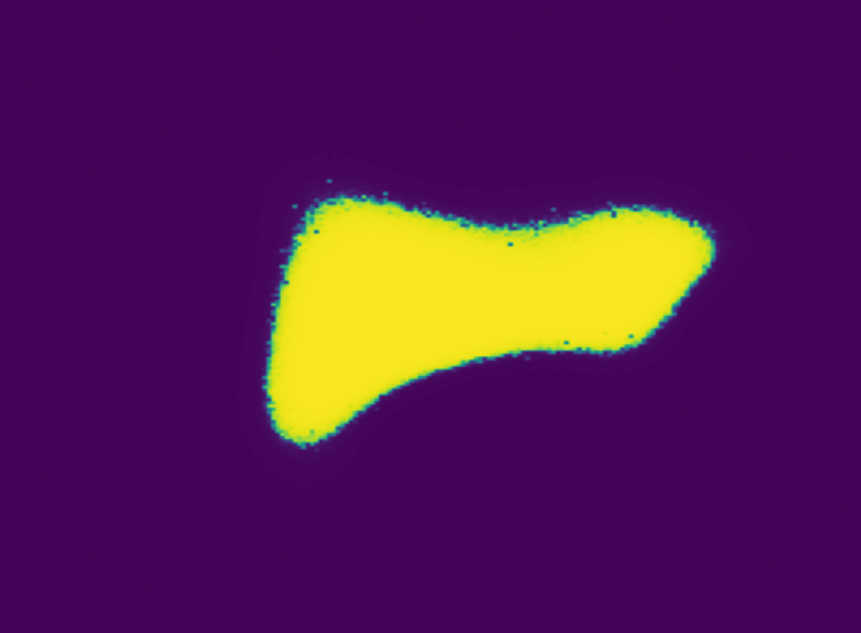};
  
    \nextgroupplot
    \addplot graphics[xmin=0,xmax=1,ymin=0,ymax=1] {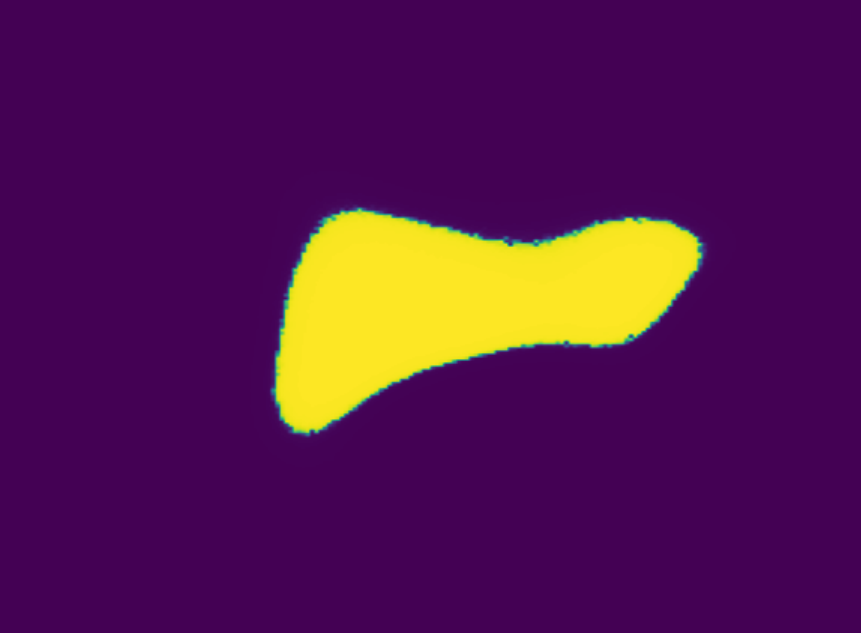};
 
    \nextgroupplot
    \addplot graphics[xmin=0,xmax=1,ymin=0,ymax=1] {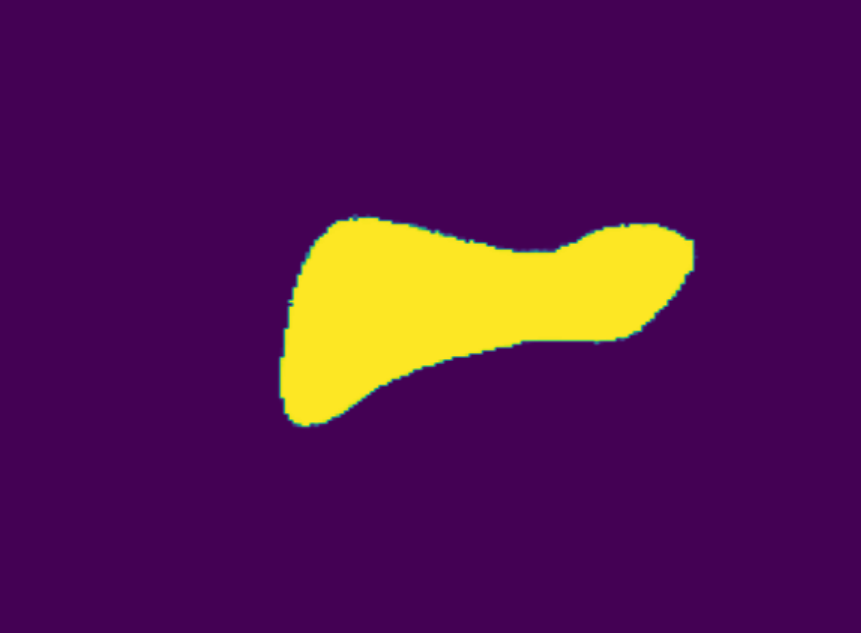};
    
    \nextgroupplot
    \addplot graphics[xmin=0,xmax=1,ymin=0,ymax=1] {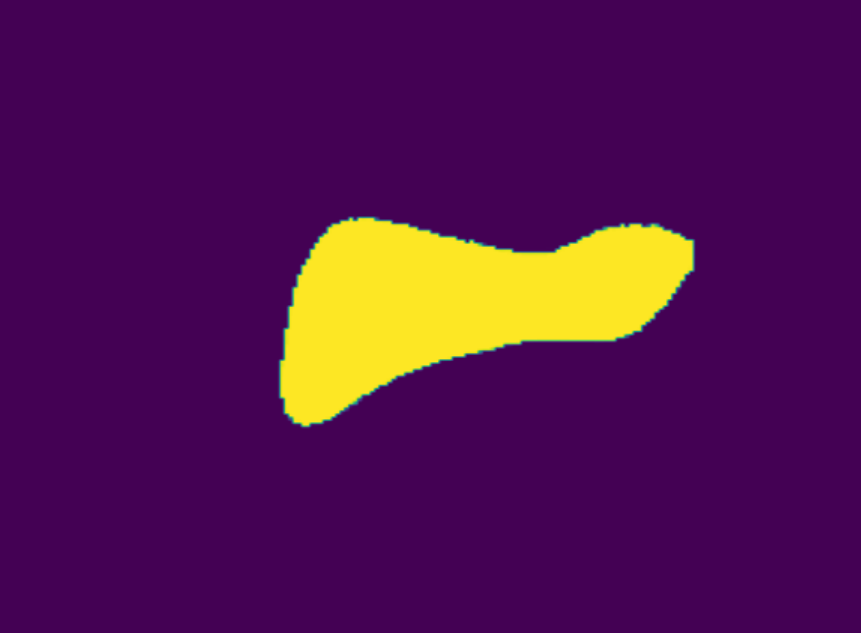};
    
    \nextgroupplot
    \addplot graphics[xmin=0,xmax=1,ymin=0,ymax=1] {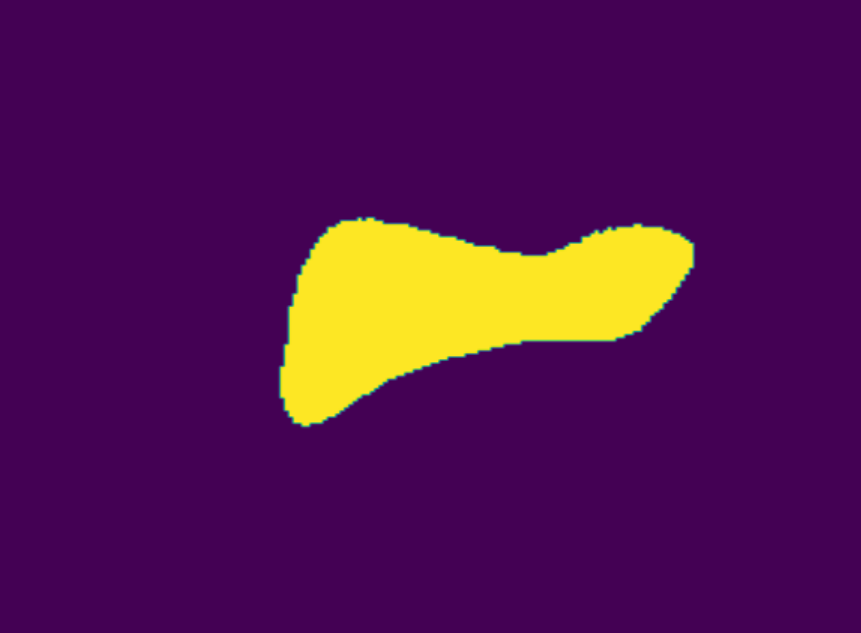};
    
  \end{groupplot}
\end{tikzpicture}
  \caption{
Each row corresponds to an example from one of the experiments in $(S)$.
The first column depicts a marginal function $m$.
In column 2-6 various iterations from the gradient descent minimization of soft-Dice $\mathrm{SD}_m(\sigma\circ f)$ is depicted. 
The last column depicts the theoretical optimal solutions $s=I_{[\sup_{s'\in\mathcal{S}}\mathrm{D}_m(s')/2,1]}\circ m$ described in Theorem~\ref{theorem1}.
% The rows are generated using the synthetic model parameters $\sigma=0.01,0.02,0.03,0.04,0.05,0.06,0.07,0.08$.
}
  \label{seqSfig}
\end{figure*}

\FloatBarrier
\end{appendices}

{\small
\bibliographystyle{ieee_fullname}
\bibliography{bibliography}

\begin{thebibliography}{10}\itemsep=-1pt

\bibitem{armato2011lung}
Samuel~G Armato~III, Geoffrey McLennan, Luc Bidaut, Michael~F McNitt-Gray,
  Charles~R Meyer, Anthony~P Reeves, Binsheng Zhao, Denise~R Aberle, Claudia~I
  Henschke, Eric~A Hoffman, et~al.
\newblock {The Lung Image Database Consortium (LIDC) and Image Database
  Resource Initiative (IDRI): A Completed Reference Database of Lung Nodules on
  CT Scans}.
\newblock {\em Medical Physics}, 38(2):915--931, 2011.

\bibitem{bao2020calibrated}
Han Bao and Masashi Sugiyama.
\newblock Calibrated surrogate maximization of linear-fractional utility in
  binary classification.
\newblock In {\em International Conference on Artificial Intelligence and
  Statistics}, pages 2337--2347. PMLR, 2020.

\bibitem{bartlett2006convexity}
Peter~L Bartlett, Michael~I Jordan, and Jon~D McAuliffe.
\newblock Convexity, classification, and risk bounds.
\newblock {\em Journal of the American Statistical Association},
  101(473):138--156, 2006.

\bibitem{bertels2019optimizing}
Jeroen Bertels, Tom Eelbode, Maxim Berman, Dirk Vandermeulen, Frederik Maes,
  Raf Bisschops, and Matthew~B Blaschko.
\newblock {Optimizing the Dice Score and Jaccard Index for Medical Image
  Segmentation: Theory and Practice}.
\newblock In {\em International Conference on Medical Image Computing and
  Computer-Assisted Intervention}, pages 92--100. Springer, 2019.

\bibitem{bertels2021theoretical}
Jeroen Bertels, David Robben, Dirk Vandermeulen, and Paul Suetens.
\newblock {Theoretical Analysis and Experimental Validation of Volume Bias of
  Soft Dice Optimized Segmentation Maps in the Context of Inherent
  Uncertainty}.
\newblock {\em Medical Image Analysis}, 67:101833, 2021.

\bibitem{bridge2016intraobserver}
Pete Bridge, Andrew Fielding, Pamela Rowntree, and Andrew Pullar.
\newblock {Intraobserver Variability: Should We Worry?}
\newblock {\em Journal of Medical Imaging and Radiation Sciences},
  47(3):217--220, 2016.

\bibitem{drozdzal2016importance}
Michal Drozdzal, Eugene Vorontsov, Gabriel Chartrand, Samuel Kadoury, and Chris
  Pal.
\newblock {The Importance of Skip Connections in Biomedical Image
  Segmentation}.
\newblock In {\em Deep Learning and Data Labeling for Medical Applications},
  pages 179--187. Springer, 2016.

\bibitem{du2020medical}
Getao Du, Xu Cao, Jimin Liang, Xueli Chen, and Yonghua Zhan.
\newblock Medical image segmentation based on u-net: A review.
\newblock {\em Journal of Imaging Science and Technology}, 64:1--12, 2020.

\bibitem{eelbode2020optimization}
Tom Eelbode, Jeroen Bertels, Maxim Berman, Dirk Vandermeulen, Frederik Maes,
  Raf Bisschops, and Matthew~B Blaschko.
\newblock Optimization for medical image segmentation: theory and practice when
  evaluating with dice score or jaccard index.
\newblock {\em IEEE Transactions on Medical Imaging}, 39(11):3679--3690, 2020.

\bibitem{gros2021softseg}
Charley Gros, Andreanne Lemay, and Julien Cohen-Adad.
\newblock {SoftSeg: Advantages of Soft Versus Binary Training for Image
  Segmentation}.
\newblock {\em Medical Image Analysis}, 71:102038, 2021.

\bibitem{heller2018imperfect}
Nicholas Heller, Joshua Dean, and Nikolaos Papanikolopoulos.
\newblock Imperfect segmentation labels: How much do they matter?
\newblock In {\em Intravascular Imaging and Computer Assisted Stenting and
  Large-Scale Annotation of Biomedical Data and Expert Label Synthesis}, pages
  112--120. Springer, 2018.

\bibitem{isensee2021nnu}
Fabian Isensee, Paul~F Jaeger, Simon~AA Kohl, Jens Petersen, and Klaus~H
  Maier-Hein.
\newblock nnu-net: a self-configuring method for deep learning-based biomedical
  image segmentation.
\newblock {\em Nature methods}, 18(2):203--211, 2021.

\bibitem{jiang2019two}
Zeyu Jiang, Changxing Ding, Minfeng Liu, and Dacheng Tao.
\newblock {Two-Stage Cascaded U-Net: 1st Place Solution to BraTS Challenge 2019
  Segmentation Task}.
\newblock In {\em International MICCAI Brainlesion Workshop}, pages 231--241.
  Springer, 2019.

\bibitem{karimi2020deep}
Davood Karimi, Haoran Dou, Simon~K Warfield, and Ali Gholipour.
\newblock {Deep Learning With Noisy Labels: Exploring Techniques and Remedies
  in Medical Image Analysis}.
\newblock {\em Medical Image Analysis}, 65:101759, 2020.

\bibitem{kats2019soft}
Eytan Kats, Jacob Goldberger, and Hayit Greenspan.
\newblock {Soft Labeling by Distilling Anatomical Knowledge for Improved MS
  Lesion Segmentation}.
\newblock In {\em 2019 IEEE 16th International Symposium on Biomedical Imaging
  (ISBI 2019)}, pages 1563--1566. IEEE, 2019.

\bibitem{lemay2022label}
Andreanne Lemay, Charley Gros, and Julien Cohen-Adad.
\newblock {Label Fusion and Training Methods for Reliable Representation of
  Inter-Rater Uncertainty}.
\newblock {\em arXiv preprint arXiv:2202.07550}, 2022.

\bibitem{li2020superpixel}
Hang Li, Dong Wei, Shilei Cao, Kai Ma, Liansheng Wang, and Yefeng Zheng.
\newblock {Superpixel-Guided Label Softening for Medical Image Segmentation}.
\newblock In {\em International Conference on Medical Image Computing and
  Computer-Assisted Intervention}, pages 227--237. Springer, 2020.

\bibitem{lipton2014optimal}
Zachary~C Lipton, Charles Elkan, and Balakrishnan Naryanaswamy.
\newblock {Optimal Thresholding of Classifiers to Maximize F1 Measure}.
\newblock In {\em Joint European Conference on Machine Learning and Knowledge
  Discovery in Databases}, pages 225--239. Springer, 2014.

\bibitem{liu2022deep}
Jun Liu, Xiangyue Wang, and Xue-cheng Tai.
\newblock {Deep Convolutional Neural Networks with Spatial Regularization,
  Volume and Star-Shape Priors for Image Segmentation}.
\newblock {\em Journal of Mathematical Imaging and Vision}, pages 1--21, 2022.

\bibitem{ma2021loss}
Jun Ma, Jianan Chen, Matthew Ng, Rui Huang, Yu Li, Chen Li, Xiaoping Yang, and
  Anne~L Martel.
\newblock Loss odyssey in medical image segmentation.
\newblock {\em Medical Image Analysis}, 71:102035, 2021.

\bibitem{nir2018automatic}
Guy Nir, Soheil Hor, Davood Karimi, Ladan Fazli, Brian~F Skinnider, Peyman
  Tavassoli, Dmitry Turbin, Carlos~F Villamil, Gang Wang, R~Storey Wilson,
  et~al.
\newblock {Automatic Grading of Prostate Cancer in Digitized Histopathology
  Images: Learning From Multiple Experts}.
\newblock {\em Medical Image Analysis}, 50:167--180, 2018.

\bibitem{nordstrom2020calibrated}
Marcus Nordstr{\"o}m, Han Bao, Fredrik L{\"o}fman, Henrik Hult, Atsuto Maki,
  and Masashi Sugiyama.
\newblock {Calibrated Surrogate Maximization of Dice}.
\newblock In {\em International Conference on Medical Image Computing and
  Computer-Assisted Intervention}, pages 269--278. Springer, 2020.

\bibitem{nordstrom2022image}
Marcus Nordstrom, Henrik Hult, Fredrik L{\"o}fman, and Jonas S{\"o}derberg.
\newblock On image segmentation with noisy labels: Characterization and volume
  properties of the optimal solutions to accuracy and dice.
\newblock {\em Advances in Neural Information Processing Systems},
  35:34321--34333, 2022.

\bibitem{nyholm2018mr}
Tufve Nyholm, Stina Svensson, Sebastian Andersson, Joakim Jonsson, Maja Sohlin,
  Christian Gustafsson, Elisabeth Kjell{\'e}n, Karin S{\"o}derstr{\"o}m, Per
  Albertsson, Lennart Blomqvist, et~al.
\newblock {MR and CT Data With Multiobserver Delineations of Organs in the
  Pelvic Area—Part of the Gold Atlas Project}.
\newblock {\em Medical Physics}, 45(3):1295--1300, 2018.

\bibitem{popordanoska2021relationship}
Teodora Popordanoska, Jeroen Bertels, Dirk Vandermeulen, Frederik Maes, and
  Matthew~B Blaschko.
\newblock {On the Relationship Between Calibrated Predictors and Unbiased
  Volume Estimation}.
\newblock In {\em International Conference on Medical Image Computing and
  Computer-Assisted Intervention}, pages 678--688. Springer, 2021.

\bibitem{ronneberger2015u}
Olaf Ronneberger, Philipp Fischer, and Thomas Brox.
\newblock {U-net: Convolutional Networks for Biomedical Image Segmentation}.
\newblock In {\em International Conference on Medical Image Computing and
  Computer-Assisted Intervention}, pages 234--241. Springer, 2015.

\bibitem{rousseau2021post}
Axel-Jan Rousseau, Thijs Becker, Jeroen Bertels, Matthew~B Blaschko, and Dirk
  Valkenborg.
\newblock {Post Training Uncertainty Calibration of Deep Networks for Medical
  Image Segmentation}.
\newblock In {\em 2021 IEEE 18th International Symposium on Biomedical Imaging
  (ISBI)}, pages 1052--1056. IEEE, 2021.

\bibitem{sharp2010plastimatch}
Gregory~C Sharp, Rui Li, John Wolfgang, G Chen, Marta Peroni, Maria~Francesca
  Spadea, Shinichro Mori, Junan Zhang, James Shackleford, and Nagarajan
  Kandasamy.
\newblock Plastimatch: an open source software suite for radiotherapy image
  processing.
\newblock In {\em Proceedings of the XVI’th International Conference on the
  use of Computers in Radiotherapy (ICCR), Amsterdam, Netherlands}, 2010.

\bibitem{siddique2021u}
Nahian Siddique, Sidike Paheding, Colin~P Elkin, and Vijay Devabhaktuni.
\newblock U-net and its variants for medical image segmentation: A review of
  theory and applications.
\newblock {\em Ieee Access}, 9:82031--82057, 2021.

\bibitem{silva2021using}
Joao~Louren{\c{c}}o Silva and Arlindo~L Oliveira.
\newblock {Using Soft Labels to Model Uncertainty in Medical Image
  Segmentation}.
\newblock {\em arXiv preprint arXiv:2109.12622}, 2021.

\bibitem{sudre2017generalised}
Carole~H Sudre, Wenqi Li, Tom Vercauteren, Sebastien Ourselin, and M
  Jorge~Cardoso.
\newblock {Generalised Dice Overlap as a Deep Learning Loss Function for Highly
  Unbalanced Segmentations}.
\newblock In {\em Deep Learning in Medical Image Analysis and Multimodal
  Learning for Clinical Decision Support}, pages 240--248. Springer, 2017.

\bibitem{tajbakhsh2020embracing}
Nima Tajbakhsh, Laura Jeyaseelan, Qian Li, Jeffrey~N Chiang, Zhihao Wu, and
  Xiaowei Ding.
\newblock {Embracing Imperfect Datasets: A Review of Deep Learning Solutions
  for Medical Image Segmentation}.
\newblock {\em Medical Image Analysis}, 63:101693, 2020.

\bibitem{vorontsov2021label}
Eugene Vorontsov and Samuel Kadoury.
\newblock Label noise in segmentation networks: mitigation must deal with bias.
\newblock In {\em Deep Generative Models, and Data Augmentation, Labelling, and
  Imperfections}, pages 251--258. Springer, 2021.

\bibitem{zhao2013beyond}
Ming-Jie Zhao, Narayanan Edakunni, Adam Pocock, and Gavin Brown.
\newblock {Beyond Fano's Inequality: Bounds on the Optimal F-score, BER, and
  Cost-Sensitive Risk and Their Implications}.
\newblock {\em The Journal of Machine Learning Research}, 14(1):1033--1090,
  2013.

\end{thebibliography}


\begin{thebibliography}{10}\itemsep=-1pt

\bibitem{nordstrom2022image}
Marcus Nordstrom, Henrik Hult, Fredrik L{\"o}fman, and Jonas S{\"o}derberg.
\newblock On image segmentation with noisy labels: Characterization and volume
  properties of the optimal solutions to accuracy and dice.
\newblock {\em Advances in Neural Information Processing Systems},
  35:34321--34333, 2022.

\bibitem{nyholm2018mr}
Tufve Nyholm, Stina Svensson, Sebastian Andersson, Joakim Jonsson, Maja Sohlin,
  Christian Gustafsson, Elisabeth Kjell{\'e}n, Karin S{\"o}derstr{\"o}m, Per
  Albertsson, Lennart Blomqvist, et~al.
\newblock {MR and CT Data With Multiobserver Delineations of Organs in the
  Pelvic Area—Part of the Gold Atlas Project}.
\newblock {\em Medical Physics}, 45(3):1295--1300, 2018.


\bibitem{sharp2010plastimatch}
Gregory~C Sharp, Rui Li, John Wolfgang, G Chen, Marta Peroni, Maria~Francesca
  Spadea, Shinichro Mori, Junan Zhang, James Shackleford, and Nagarajan
  Kandasamy.
\newblock Plastimatch: an open source software suite for radiotherapy image
  processing.
\newblock In {\em Proceedings of the XVI’th International Conference on the
  use of Computers in Radiotherapy (ICCR), Amsterdam, Netherlands}, 2010.

\end{thebibliography}
}

\end{document}

\onecolumn
\maketitle

% \appendix
\tableofcontents

\clearpage

\section{Proofs} \noindent

\subsection{Proof of Theorem 1} \noindent
First introduce the following generalization of Dice to soft labels
\begin{align}
    \mathrm{D}_m'(c) = \frac{2\int_\Omega c(\omega) m(\omega) \lambda(d\omega)}{\lVert c\rVert_1 + \lVert m\rVert_1}, \quad c\in \mathcal{M},
\end{align}
and note that for any $c^*\in\mathcal{M}$ that
\begin{align}
    \mathrm{SD}_m(c^*) = \inf_{c\in\mathcal{C}} \mathrm{SD}_m(c) \iff
    \mathrm{D}'_m(c^*) = \sup_{c\in\mathcal{C}} \mathrm{D}'_m(c).
\end{align}
% This form will be used in the proof because it is closer to how Dice is defined over hard labels.
Now, using this, the theorem is shown in four steps.
Firstly, the case when $\lVert m \rVert_1 = 0$ is considered.
For the rest, it is assumed that $\lVert m\rVert_1>0$.
Secondly, an upper bound is shown.
Thirdly, a characterization of the optimal solutions is shown.
Fourthly, the characterization is connected to the set $\mathcal{M}_m^*$.
Throughout, $\lambda_1$ will denote the standard Lebesgue measure over $[0,1]$ (similarly to $\lambda$ which is similarly defined over $\Omega=[0,1]^n$ for some $n\ge 1$).

\paragraph{Part 1, corner case.}
For the case $\lVert m\rVert_1=0$, $\mathrm{D}_m'(c) = 0$ for any $c\in \mathcal{M}$, and so the supremum is attained for any $c\in\mathcal{M}$.
Since, $m(\omega) = 0$ for $\omega\in\Omega,\lambda$-a.e. and $\sup_{c\in\mathcal{M}}\mathrm{D}_m'(c)=0$ it follows that $m(\omega) = \sup_{s\in\mathcal{S}}\mathrm{D}_m(s)/2$ for $\omega\in\Omega,\lambda$-a.e.
Consequently, $\mathcal{M}_m^* = \mathcal{M}$ whenever $\lVert m\rVert_1=0$.

\paragraph{Part 2, existence.}
First note that the following holds for any soft-segmentation $c\in\mathcal{M}$,
\begin{align}
    c = \int_{(0,1)} I_{[a,1]} \circ c\lambda_1(da), \quad \lambda-\text{a.e}.
\end{align}
Furthermore, since $c(\omega) \ge 0,\omega\in\Omega$, 
\begin{align}
    \lVert c \rVert_1 = \int_{(0,1)}  \lVert I_{[a,1]} \circ c \rVert_1 \lambda_1(da).
\end{align}
This together with the definition of Dice yields
\begin{align}
    \mathrm{D}_m'(c) &=
     \int_{(0,1)}\mathrm{D}_m(I_{[a,1]}\circ c)\frac{\lVert I_{[a,1]}\circ c \rVert_1 + \lVert m\rVert_1}{\lVert c \rVert_1 + \lVert m\rVert_1}  \lambda_1(da) \\
    &\le
    \sup_{s\in\mathcal{S}} \mathrm{D}_m(s) \int_{(0,1)} \frac{\lVert I_{[a,1]}\circ c \rVert_1 + \lVert m\rVert_1}{\lVert c \rVert_1 + \lVert m\rVert_1} \lambda_1(da) \\
    &=
    \sup_{s\in\mathcal{S}} \mathrm{D}_m(s)  \frac{ \lVert \int_{(0,1)} I_{[a,1]}\circ c \lambda_1(da)\rVert_1 + \lVert m\rVert_1}{\lVert c \rVert_1 + \lVert m\rVert_1}  \\
    &=
    \sup_{s\in\mathcal{S}} \mathrm{D}_m(s)  \frac{ \lVert c \rVert_1 + \lVert m\rVert_1}{\lVert c \rVert_1 + \lVert m\rVert_1}  \\
    &=
    \sup_{s\in\mathcal{S}} \mathrm{D}_m(s).
\end{align}
Hence,
\begin{align}
    \sup_{c\in\mathcal{M}} \mathrm{D}_m'(c) \le 
    \sup_{s\in\mathcal{S}} \mathrm{D}_m(s).
\end{align}
however, since $\mathcal{S} \subset \mathcal{M}$,
\begin{align}
    \sup_{c\in\mathcal{M}} \mathrm{D}_m'(c) = 
    \sup_{s\in\mathcal{S}} \mathrm{D}_m(s).
\end{align}

\paragraph{Part 3, characterization.}
Now consider the set
\begin{align}
    A_\epsilon =\{a\in(0,1) : \mathrm{D}_m(I_{[a,1]}\circ c) \ge \sup_{s\in\mathcal{S}} \mathrm{D}_m(s) -\epsilon \}.
\end{align}
and note that there are two situations
\begin{align}
    & (1) \exists \epsilon > 0 \implies \lambda_1(A_\epsilon^C) > 0, \\
    & (2) \forall \epsilon > 0 \implies \lambda_1(A_\epsilon^C) = 0.
\end{align}
For the first case (1), then for some $\epsilon > 0$
\begin{align}
    \mathrm{D}_m'(c) &=
     \int_{(0,1)} \mathrm{D}_m(I_{[a,1]} \circ c)\frac{\lVert I_{[a,1]}\circ c \rVert_1 + \lVert m\rVert_1}{\lVert c\rVert_1 + \lVert m\rVert_1}  \lambda_1(da) \\
     &\le
     \sup_{s\in \mathcal{S}}\mathrm{D}_m(s) \int_{ A_\epsilon } 
     \frac{\lVert I_{[a,1]}\circ c \rVert_1 + \lVert m\rVert_1}{\lVert c\rVert_1 + \lVert m\rVert_1} \lambda_1(da)
     +
     (\sup_{s\in \mathcal{S}}\mathrm{D}_m(s) -\epsilon)\int_{ A_\epsilon^C } 
     \frac{\lVert I_{[a,1]}\circ c \rVert_1 + \lVert m\rVert_1}{\lVert c\rVert_1 + \lVert m\rVert_1} \lambda_1(da)\\
     &=
     \sup_{s\in \mathcal{S}}\mathrm{D}_m(s)
     -\epsilon\int_{ A_\epsilon^C } 
     \frac{\lVert I_{[a,1]}\circ c \rVert_1 + \lVert m\rVert_1}{\lVert c\rVert_1 + \lVert m\rVert_1} \lambda_1(da).
\end{align}
Since, for each $a \in (0,1)$ and $c \in \mathcal{M}$, 
\begin{align}
       \frac{\lVert I_{[a,1]}\circ c \rVert_1 + \lVert m\rVert_1}{\lVert c\rVert_1 + \lVert m\rVert_1} \geq \frac{\lVert m\rVert_1}{1 + \lVert m\rVert_1}, 
\end{align}
it follows that the expression in the previous display is bounded above by
\begin{align}
     \sup_{s\in \mathcal{S}}\mathrm{D}_m(s)
     -\epsilon 
     \lambda_1(A_\epsilon^C) \frac{\lVert m \rVert_1}{1+\lVert m\rVert_1}
    <
    \sup_{s\in\mathcal{S}} \mathrm{D}_m(s),
\end{align}
where the last follows by the assumption $\lVert m \rVert_1>0$. We conclude that in case (1) 
\begin{align}
    \sup_{c\in\mathcal{M}} \mathrm{D}_m'(c)<
    \sup_{s\in\mathcal{S}} \mathrm{D}_m(s). 
\end{align}
For the second case (2),
\begin{align}
    \mathrm{D}_m'(c) &=
     \int_{(0,1)} \mathrm{D}_m(I_{[a,1]} \circ c)\frac{\lVert I_{[a,1]}\circ c \rVert_1 + \lVert m\rVert_1}{\lVert c\rVert_1 + \lVert m\rVert_1}  \lambda_1(da) \\
     &=
     \sup_{s\in\mathcal{S}} \mathrm{D}_m(s)\int_{(0,1)} \frac{\lVert I_{[a,1]}\circ c \rVert_1 + \lVert m\rVert_1}{\lVert c\rVert_1 + \lVert m\rVert_1}  \lambda_1(da) \\
     &=
     \sup_{s\in\mathcal{S}} \mathrm{D}_m(s) \frac{\int_{(0,1)}\lVert I_{[a,1]}\circ c \rVert_1 \lambda_1(da) + \lVert m\rVert_1}{\lVert c\rVert_1 + \lVert m\rVert_1}  \\
     &=
     \sup_{s\in\mathcal{S}} \mathrm{D}_m(s) \frac{\lVert c \rVert_1  + \lVert m\rVert_1}{\lVert c\rVert_1 + \lVert m\rVert_1}  \\
    &=
    \sup_{s\in\mathcal{S}} \mathrm{D}_m(s).
\end{align}

\paragraph{Part 4, final form.}
By Theorem 2 and Theorem 5 in \cite{nordstrom2022image}, it follows that the optimizers $\mathcal{S}_m^*\subset \mathcal{S}$ to $\sup_{s\in\mathcal{S}} \mathrm{D}_m(s)$, are given by the elements satisfying
\begin{align}\label{eq:s}
    s(\omega) \in
    \begin{cases}
        \{ 0 \} &\text{ if } m(\omega) < \sup_{s'\in\mathcal{S}} \mathrm{D}_m(s')/2, \\
        \{ 0,1 \} &\text{ if } m(\omega)= \sup_{s'\in\mathcal{S}} \mathrm{D}_m(s')/2, \\
        \{ 1 \} &\text{ if } m(\omega) > \sup_{s'\in\mathcal{S}} \mathrm{D}_m(s')/2,
    \end{cases}
\end{align}
$\omega\in\Omega$, $\lambda$-a.e. 
If the second case (2) hold, then almost every $a\in(0,1)$ satisfies
\begin{align}
    \mathrm{D}_m(I_{[a,1]}\circ c) = \sup_{s\in\mathcal{S}} \mathrm{D}_m(s),
\end{align}
and consequently that $I_{[a,1]}\circ c \in \mathcal{S}_m^*$ for almost every $a\in(0,1)$.
Now, since
\begin{align}
c = \int_{(0,1)} I_{[a,1]} \circ c \lambda_1(da),
\end{align}
it follows that the optimizers $\mathcal{M}_m^*\subset \mathcal{M}$ to $\sup_{c\in\mathcal{M}} \mathrm{D}_m'(c)$ are the elements satisfying
\begin{align}
    c(\omega) \in
    \begin{cases}
        \{ 0 \} &\text{ if } m(\omega) < \sup_{s'\in\mathcal{S}} \mathrm{D}_m(s')/2, \\
        [ 0,1 ] &\text{ if } m(\omega)= \sup_{s'\in\mathcal{S}} \mathrm{D}_m(s')/2, \\
        \{ 1 \} &\text{ if } m(\omega) > \sup_{s'\in\mathcal{S}} \mathrm{D}_m(s')/2,
    \end{cases}
\end{align}
$\omega\in\Omega$, $\lambda$-a.e.
% This means $I_{[a,1]}\circ c \in\mathcal{S}_m^*$  for almost every $a\in(0,1)$, which can only be the case if $c\in\mathcal{M}_m^*$.
This completes the proof. \qed

\subsection{Proof of Theorem 2} \noindent
This theorem is shown in two simple steps.
Firstly, similarities with previous results are observed.
Secondly, a simple observation is made connecting this with the presented result.

\paragraph{Part 1, hard-segmentations.}
The following is a consequence of Theorem 2 and Theorem 5 in \cite{nordstrom2022image}.
For any $m\in\mathcal{M}$, the class $\mathcal{S}^*_m \subset \mathcal{S}$ of maximizers attaining the supremum to $\sup_{s\in\mathcal{S}}\mathrm{D}_m(s)$
satisfy the bounds
\begin{align}
[\inf_{s\in\mathcal{S}_m^*}\lVert  s\rVert_1,\sup_{s\in\mathcal{S}_m^*} \lVert s \rVert_1] \subseteq [\lVert m \rVert_1^2,1].
\end{align}
Moreover, the bounds are sharp in the sense that there for any $v\in(0,1]$ exist $m_0,m_1\in\mathcal{M}$ such that $\lVert m_0 \rVert_1 = \lVert m_1 \rVert_1 = v$ and
\begin{align}
\inf_{s\in\mathcal{S}_{m_0}^*} \lVert s \rVert_1 = \lVert m_0 \rVert_1^2, \quad \sup_{s\in\mathcal{S}_{m_1}^*}\lVert s \rVert_1 =1.
\end{align}
Furthermore, $\mathcal{S}_m^*$ is given by the elements.
\begin{align}
    s(\omega) \in
    \begin{cases}
        \{ 0 \} &\text{ if } m(\omega) < \sup_{s'\in\mathcal{S}} \mathrm{D}_m(s')/2, \\
        \{ 0,1 \} &\text{ if } m(\omega)= \sup_{s'\in\mathcal{S}} \mathrm{D}_m(s')/2, \\
        \{ 1 \} &\text{ if } m(\omega) > \sup_{s'\in\mathcal{S}} \mathrm{D}_m(s')/2,
    \end{cases}
\end{align}
$\omega\in\Omega$, $\lambda$-a.e.

\paragraph{Part 2, soft-segmentations.}
Let $\mathcal{M}_m^*\subset \mathcal{M}$ be the elements attaining the infimum to $\inf_{c\in\mathcal{M}}\mathrm{SD}_m(c)$, which as shown by Theorem~1 is given by
\begin{align}
    c(\omega) \in
    \begin{cases}
        \{ 0 \} &\text{ if } m(\omega) < \sup_{s'\in\mathcal{S}} \mathrm{D}_m(s')/2, \\
        [ 0,1 ] &\text{ if } m(\omega)= \sup_{s'\in\mathcal{S}} \mathrm{D}_m(s')/2, \\
        \{ 1 \} &\text{ if } m(\omega) > \sup_{s'\in\mathcal{S}} \mathrm{D}_m(s')/2,
    \end{cases}
\end{align}
$\omega\in\Omega$, $\lambda$-a.e.
% In this part, we show that the previous result also holds when the optimization is done over $\mathcal{M}$ rather than $\mathcal{S}$.
Now, observe that for any $m\in\mathcal{M}$,
\begin{align}
    \inf_{s\in\mathcal{S}_m^*}\lVert s \rVert_1 =
    \inf_{s\in\mathcal{M}_m^*}\lVert c \rVert_1, 
\end{align}
and infimum are in both cases attained by:
\begin{align}
s(\omega) = I_{(\sup_{s'\in\mathcal{S}} \mathrm{D}_m(s')/2,1]} \circ m, \quad \omega \in \Omega.
\end{align}
Similarly, for any $m\in\mathcal{M}$,
\begin{align}
    \sup_{s\in\mathcal{S}_m^*}\lVert s \rVert_1 =
    \sup_{s\in\mathcal{M}_m^*}\lVert c \rVert_1.
\end{align}
and supremum are in both cases attained by:
\begin{align}
s(\omega) = I_{[\sup_{s'\in\mathcal{S}} \mathrm{D}_m(s')/2,1]} \circ m, \quad \omega \in \Omega.
\end{align}
Consequently, for any $m\in\mathcal{M}$,
\begin{align}
[\inf_{c\in\mathcal{M}_m^*}\lVert  c\rVert_1,\sup_{c\in\mathcal{M}_m^*} \lVert c \rVert_1] =
[\inf_{s\in\mathcal{S}_m^*}\lVert  s\rVert_1,\sup_{s\in\mathcal{S}_m^*} \lVert s \rVert_1] \subseteq [\lVert m \rVert_1^2,1].
\end{align}
Moreover, the bounds are sharp in the sense that there for any $v\in(0,1]$ exist $m_0,m_1\in\mathcal{M}$ such that $\lVert m_0 \rVert_1 = \lVert m_1 \rVert_1 = v$ and
\begin{align}
\inf_{c\in\mathcal{M}_{m_0}^*}\lVert c \rVert_1 = \inf_{s\in\mathcal{S}_{m_0}^*}\lVert s \rVert_1 = \lVert m_0 \rVert_1^2, \quad 
\sup_{c\in\mathcal{M}_{m_1}^*}\lVert c \rVert_1 = \sup_{s\in\mathcal{S}_{m_1}^*}\lVert s \rVert_1 =1.
\end{align}
\label{theorem2}
This completes the proof. \qed

\subsection{Proof of Theorem 3}
\noindent
First introduce the following generalization of Dice to soft labels
\begin{align}
    \mathrm{D}_m'(c) = \frac{2\int_\Omega c(\omega) m(\omega) \lambda(d\omega)}{\lVert c\rVert_1 + \lVert m\rVert_1}, \quad c\in \mathcal{M},
\end{align}
and note that for any $c^*\in\mathcal{M}$ that
\begin{align}
    \lim_{l\rightarrow\infty} \mathrm{SD}_m(c_l) = \inf_{c\in\mathcal{M}} \mathrm{SD}_m(c) \iff
    \lim_{l\rightarrow\infty} \mathrm{D}_m'(c_l) = \sup_{c\in\mathcal{M}} \mathrm{D}_m'(c).
\end{align}
Now, using this, the theorem is shown in three steps.
Firstly, continuity of $\mathrm{D}_m'$ over $\mathcal{M}$ is shown.
Secondly, an implication of soft labels converging to optimal soft-Dice is shown.
Thirdly, this implication is used to show that the same sequence, when thresholded appropriately, converges to optimal Dice.

\paragraph{Part 1, continuity.}
% We start by showing continuity of $\mathrm{D}_m$ over $\mathcal{M}$.
First, note that if $\lVert m\rVert_1 = 0$ , then $\mathrm{D}_m'(c)=0$ for any $c$ which means that $\mathrm{D}_m'$ is continuous.
Now, assume that $\lVert m \rVert_1 >0$ and let $c,c_0\in\mathcal{M}$ be any two elements satisfying
\begin{align}
    \lVert c-c_0\rVert_1 \le \delta,
\end{align}
Then
\begin{align}
    |\mathrm{D}_m'(c_0) - \mathrm{D}_m'(c)| &=
    \left|\mathrm{D}_m'(c_0)-
    \mathrm{D}_m'(c_0) \frac{\lVert c_0 \rVert_1 + \lVert m \rVert_1}{\lVert c \rVert_1 + \lVert m \rVert_1} -
    \frac{2\int_\Omega (c(\omega)-c_0(\omega)) m(\omega)\lambda(d\omega)}{\lVert c\rVert_1 + \lVert m \rVert_1} \right| \\
    &=
    \left|\mathrm{D}_m'(c_0)
    \frac{\lVert c \rVert_1 - \lVert c_0\rVert_1}{\lVert c \rVert_1 + \lVert m \rVert_1} -
    \frac{2\int_\Omega (c(\omega)-c_0(\omega)) m(\omega)\lambda(d\omega)}{\lVert c\rVert_1 + \lVert m\rVert_1} \right| \\
    &\le
    \mathrm{D}_m'(c_0)
    \frac{|\lVert c \rVert_1 - \lVert c_0\rVert_1 |}{\lVert c \rVert_1 + \lVert m \rVert_1} +
    \frac{2|\int_\omega (c(\omega)-c_0(\omega)) m(\omega)\lambda(d\omega)|}{\lVert c\rVert_1 + \lVert m\rVert_1}  \\
    &\le
    \frac{\lVert c - c_0\rVert_1 }{\lVert c \rVert_1 + \lVert m\rVert_1} +
    \frac{2\lVert c - c_0\rVert_1}{\lVert s\rVert_1 + \lVert m \rVert_1}  \\
    &\le
    \frac{3\delta}{\lVert m\rVert_1 }
\end{align}
Consequently, for any $\epsilon > 0$,
\begin{align}
    \lVert c-c_0\rVert_1 \le \delta = \frac{\lVert m \rVert_1}{3}\epsilon
\end{align}
implies that
\begin{align}
    |\mathrm{D}_m'(c_0) - \mathrm{D}_m'(c)| \le \epsilon.
\end{align}
% The exact same reasoning holds if $\mathrm{D}_m$ is restricted to $\mathcal{M}$ or to $\mathcal{S}$.

\paragraph{Part 2, convergence of soft-labels.} 
In this part, we show that for any relatively compact sequence $\{c_l\}_{l\ge 1} \subset \mathcal{M}$, then
\begin{align}
     \lim_{l\rightarrow\infty}\mathrm{D}_m'(c_l) = \sup_{c\in\mathcal{M}} \mathrm{D}_m'(c)
     \quad\implies\quad
     \lim_{l\rightarrow\infty} \inf_{c^*\in\mathcal{M}_m^*} \lVert c_l - c^*\rVert_1 = 0.
\end{align}
For this, introduce the following $\epsilon$-expansion of $\mathcal{M}_m^*$
\begin{align}
    \mathcal{M}_m^{*,\epsilon}  = \{ c \in \mathcal{M} : \inf_{c^*\in\mathcal{M}_m^*}\lVert c-c^*\rVert_1 \le \epsilon  \}.
\end{align}
Since the closure to a subset of a relatively compact set is compact, it follows that $\text{closure}(\{c_l\}_{l\ge 1}\setminus \mathcal{M}_m^{*,\epsilon})$ is compact, and by the Weierstrass extreme value theorem that
\begin{align}
\exists c' \in \text{closure}(\{c_l\}_{l\ge 1}\setminus \mathcal{M}_m^{*,\epsilon}) \quad &\implies\quad
    \mathrm{D}_m'(c') = \sup_{c\in \{c_l\}_{l\ge 1}\setminus \mathcal{M}_m^{*,\epsilon}} \mathrm{D}_m'(c). 
\end{align}
Now, note that if
\begin{align}
    \mathrm{D}_m'(c') = \sup_{c\in\mathcal{M}} \mathrm{D}_m'(c),
\end{align}
then it follows that $c'\in\mathcal{M}_m^*$ which is a contradiction by construction, consequently
\begin{align}
    \sup_{c\in \{ c_l\}_{l\ge 1} \setminus\mathcal{M}_m^{\epsilon,*}} \mathrm{D}_m'(c) <
    \sup_{c\in\mathcal{M}} \mathrm{D}_m'(c),
\end{align}
and thus %, for any relatively compact sequence $\{c_l\}_{l\ge 1}\subset \mathcal{M}$ 
\begin{align}
     \lim_{l\rightarrow\infty}\mathrm{D}_m'(c_l) = \sup_{c\in\mathcal{M}} \mathrm{D}_m'(c)
     \quad\implies\quad
     \lim_{l\rightarrow\infty} \inf_{c^*\in\mathcal{M}_m^{*,\epsilon}} \lVert c_l - c^*\rVert_1 = 0.
\end{align}
But since $\epsilon>0$ is arbitrary, we must have that
\begin{align}
     \lim_{l\rightarrow\infty}\mathrm{D}_m'(c_l) = \sup_{c\in\mathcal{M}} \mathrm{D}_m'(c)
     \quad\implies\quad
     \lim_{l\rightarrow\infty} \inf_{c^*\in\mathcal{M}_m^*} \lVert c_l - c^*\rVert_1 = 0.
\end{align}

\paragraph{Part 3. convergence of thresholded soft-labels.} 
In this part, we show that for any sequence $\{c_l\}_{l\ge 1} \subset \mathcal{M}$ and constant $a\in (0,1)$ it follows that
\begin{align}
     \lim_{l\rightarrow\infty} \inf_{c^*\in\mathcal{M}_m^*} \lVert c_l - c^*\rVert_1 = 0
     \implies 
    \lim_{l\rightarrow\infty} \mathrm{D}_m(I_{[a,1]} \circ c_l) = \sup_{s\in\mathcal{S}} \mathrm{D}_m(s).
 \end{align}
For this, take any $a\in (0,1)$ and $\epsilon > 0$.
Then there exist some $L\ge 1$ such that for $l\ge L$ 
\begin{align}
       \inf_{c^*\in\mathcal{M}_m^*}\lVert c_l - c^*\rVert_1 \le \epsilon.
\end{align}
Furthermore since
\begin{align}
    & \inf_{c^*\in\mathcal{M}_m^*}\lVert c_l - c^*\rVert_1 =  \\ 
    &\quad \int_\Omega I_{[0,\sup_{s\in\mathcal{S}} \mathrm{D}_m(s)/2)}( m(\omega))\rvert c_l(\omega) - 0 \rvert \lambda(d\omega) 
    +
    \int_\Omega I_{(\sup_{s\in\mathcal{S}} \mathrm{D}_m(s)/2,1]}(m(\omega)) \rvert c_l(\omega) - 1 \rvert \lambda(d\omega),
\end{align}
we have that 
\begin{align}
    \int_\Omega I_{[0,\sup_{s\in\mathcal{S}} \mathrm{D}_m(s)/2)} (m(\omega))\rvert c_l(\omega) - 0 \rvert \lambda(d\omega)  \le \epsilon, \\
    \int_\Omega I_{(\sup_{s\in\mathcal{S}} \mathrm{D}_m(s)/2,1]}(m(\omega)) \rvert c_l(\omega) - 1 \rvert \lambda(d\omega) \le \epsilon.
\end{align}
This in turn implies that
\begin{align}
    \int_\Omega I_{[0,\sup_{s\in\mathcal{S}} \mathrm{D}_m(s)/2)} (m(\omega))\rvert I_{[a,1]}(c_l(\omega)) - 0 \rvert \lambda(d\omega)  \le \epsilon/\min\{a,1-a\},\\
    \int_\Omega I_{(\sup_{s\in\mathcal{S}} \mathrm{D}_m(s)/2,1]}(m(\omega)) \rvert I_{[a,1]}(c_l(\omega)) - 1 \rvert \lambda(d\omega) \le \epsilon / \min\{a,1-a\},
\end{align}
 and consequently since
\begin{align}
    &\inf_{s^*\in\mathcal{S}_m^*}\lVert I_{[a,1]}\circ c_l - s^*\rVert_1 = \\
    &\int_\Omega I_{[0,\sup_{s\in\mathcal{S}} \mathrm{D}_m(s)/2)} (m(\omega))\rvert I_{[a,1]}(c_l(\omega)) - 0 \rvert \lambda(d\omega)  +
    \int_\Omega I_{(\sup_{s\in\mathcal{S}} \mathrm{D}_m(s)/2,1]}(m(\omega)) \rvert I_{[a,1]}(c_l(\omega)) - 1 \rvert \lambda(d\omega)
\end{align}
also that
\begin{align}
    \inf_{s^*\in\mathcal{S}_m^*}\lVert I_{[a,1]}\circ c_l - s^*\rVert_1 \le 2\epsilon/\min\{a,1-a\}.
\end{align}
Since $\epsilon>0$ is arbitrary, this means that
\begin{align}
    \lim_{l\rightarrow\infty}\inf_{s^*\in\mathcal{S}_m^*}\lVert I_{[a,1]} \circ c_l - s^*\rVert_1 = 0.
\end{align}
By continuity of $\mathrm{D}_m$, we finally have that
\begin{align}
    \lim_{l\rightarrow\infty} \mathrm{D}_m(I_{[a,1]} \circ c_l) = \sup_{s\in\mathcal{S}} \mathrm{D}_m(s).
\end{align}
This completes the proof. \qed

\section{Experiments} \noindent
All of the experiments simply compute the steps in a gradient descent scheme.
Of interest is the marginal function (soft labels) under consideration $m\in\mathcal{M}$, the sequence $\{f_l\}_{l\ge 1}\subset \mathcal{F}$ generated with a gradient descent scheme, and the theoretical optimal segmentation $s=I_{[\sup_{s'\in\mathcal{S}}\mathrm{D}_m(s')/2, 1] }\circ m\in\mathcal{S}_m^*$.
For the computation, the continuous domain is discretized to a set of voxels.
Let $N$ be the number of voxels, and the voxels be given by the paritioning $\{\Omega_i\}_{i=1}^N$ satisfying $\Omega = \Omega_1\cup \dots \cup\Omega_n$ with $\Omega_i\cap \Omega_j = \emptyset$ when $i\not=j$ and $\lambda(\Omega_i) =1/N$ for all $1\le i \le N$.
The computation is then carried out on this voxelization by restricting the functions to be constant on each voxel
\begin{align}
    m(\omega) &\doteq M(i), \quad \omega \in \Omega_i, \\
    f_l(\omega) &\doteq F_l(i), \quad \omega \in \Omega_i, \\
    s(\omega) &\doteq S(i), \quad \omega \in \Omega_i,
\end{align}
for $i=1,\dots,N$ and $l=1, 2,\dots.$.
At start, $M$ is provided in the discretized form, whereas $\{F_l\}_{l\ge 1}$ and $S$ are computed.
More precisely, $\{F_l\}_{l\ge 1}$ is generated by first assigning the elements in $F_1$ with independent draws of a standard normal distribution
\begin{align}
    F_1(i) \sim \text{normal}(0,1),
\end{align}
and then sequentially computing the rest
\begin{align}
    F_{l+1}(i) &=
    F_l(i) - \gamma\frac{d}{dF_l(i)} \left[
    \frac{2\frac{1}{N}\sum_{j=1}^N \sigma(F_l(j))M(j)}{\frac{1}{N}\sum_{j=1}^N M(j) + \frac{1}{N}\sum_{j=1}^N \sigma(F_l(i))}
    \right] \\
    &=
    F_l(i) - \gamma  \frac{2 M(i) - 2\frac{1}{N}\sum_{j=1}^N \sigma(F_l(j))M(j) }{(\sum_{j=1}^N M(j) + \sum_{j=1}^N \sigma(F_l(i)))^2} \sigma(F_l(i)) \sigma(-F_l(i)).
\end{align}
The theoretically optimal solution is computed by taking
\begin{align}
    S(i) &= I\left\{ M(i) \ge \max_{t\in [0,1]} 
    \frac{2\frac{1}{N}\sum_{j=1}^N I\{M(j) \ge t\} M(j)}{\frac{1}{N}\sum_{j=1}^N M(j) + \frac{1}{N}\sum_{j=1}^N I\{M(j) \ge t\}} \frac{1}{2}
    \right\} \\
    &=
    I\left\{ M(i) \ge \max_{t\in \{M(1),\dots,M(N)\}} 
    \frac{\sum_{j=1}^N I\{M(j) \ge t\} M(j)}{\sum_{j=1}^N M(j) + \sum_{j=1}^N I\{M(j) \ge t\}}
    \right\},
\end{align}
and the final results that are reported in the paper are
\begin{align}
    e_{0,l} &= \frac{1}{N}\sum_{j=1}^N \lvert \sigma(F_{l}(j)) - S(j)\rvert, \\
    e_{1,l} &= \frac{1}{N}\sum_{j=1}^N \lvert I\{\sigma(F_{l}(j)) \ge 1/2\} - S(j)\rvert.
\end{align}
The experiments are done with three different data sets which we will now describe.
\subsection{Experiment (G)} \noindent
The experiment (G) is in 3D and based on pelvic data from the Gold Atlas project~\cite{nyholm2018mr}.
More specifically, it includes nine different ROIs (region of interests) urinary bladder, rectum, anal canal, penile bulb, neurovascular bundles, femoral head right, femoral head left, prostate and seminical vesicles.
Each ROI has been delineated by $5$ different medical practitioners.
The soft-labels used in the experiments are formed, for each case and ROI, by taking the voxel-wise average of the segmentation's done by the different practitioners.
% Let $l_i^{r,c}\in\mathcal{S}$ be the segmentation of ROI with index $r=1,\dots,9$, medical pracitioner with index $i=1,\dots,4$, where the case has index $c=1,\dots,19$.
\\ \\
\noindent
To get the raw data, visit the following link and request access.
\begin{center}
\url{https://doi.org/10.5281/zenodo.583096}
\end{center}
When granted, download all of the files and put them in the folder
 \path{Experiments_G/dicom}.
 Make sure also to rename \path{3_03_P (1).zip}  to \path{3_03_P.zip}.
When done, the following files should be located in the folder:
\path{1_01_P.zip},
\path{1_02_P.zip},
\path{1_03_P.zip},
\path{1_04_P.zip},
\path{1_05_P.zip},
\path{1_06_P.zip},
\path{1_07_P.zip},
\path{1_08_P.zip},
\path{2_03_P.zip},
\path{2_04_P.zip},
\path{2_05_P.zip},
\path{2_06_P.zip},
\path{2_09_P.zip},
\path{2_10_P.zip},
\path{2_11_P.zip},
\path{3_01_P.zip},
\path{3_02_P.zip},
\path{3_03_P.zip},
\path{3_04_P.zip}.
\\
\\
\noindent
To extract the data, the \emph{Plastimatch}~\cite{sharp2010plastimatch} software with version 1.9.3 is used.
The installer and information about the license can be found by following the following link.
\begin{center}
    \url{http://plastimatch.org/}
\end{center}
\noindent
Code is attached together with a Python requirements file.
The privacy concerns of the patients have been handled by letting the participants sign informed consent forms.
There is furthermore to the best of our knowledge no explicit license attached to the data, however, it is stated in the original reference that it is freely available for use in non-commercial applications.
The experiment is CPU based and conducted on a Windows 10 machine with Python 3.10.4.

\subsection{Experiment (S)} \noindent
The experiment (S) is based on synthetic data.
The ambition with this synthetic data is to generate marginal functions that are similar to \emph{real} marginal functions we expect occur in practice.
Note that the experiments (G) are with respect to real world data, however, finite sample approximations are used in the sense that voxel-wise averages over the samples are computed.
This approximation is limited by the number of samples and if we had access to an infinite amount we expect a smooth looking marginal function, rather than the step-wise type of marginals that we get with finite samples.
The synthetic data we generate is smooth in this sense.
It is computed with the following three steps.
\begin{figure*}[htb!]
  \centering
\begin{tikzpicture}
  \begin{groupplot}[group style={group size=3 by 1, horizontal sep=0.2cm},height=6.0cm,width=6.0cm,xmajorgrids,ymajorgrids,xtick={0.4,0.7,1.0,1.3,1.6},xmin=0.4,xmax=1.6,ytick={0.4,0.7,1.0,1.3,1.6},ymin=0.4,ymax=1.6]
  
    \nextgroupplot[title={$m_0$},xtick=\empty,ytick=\empty,yticklabel pos = right, xmin=0, xmax=1,ymin=0,ymax=1]
    \addplot graphics[xmin=0,xmax=1,ymin=0,ymax=1] {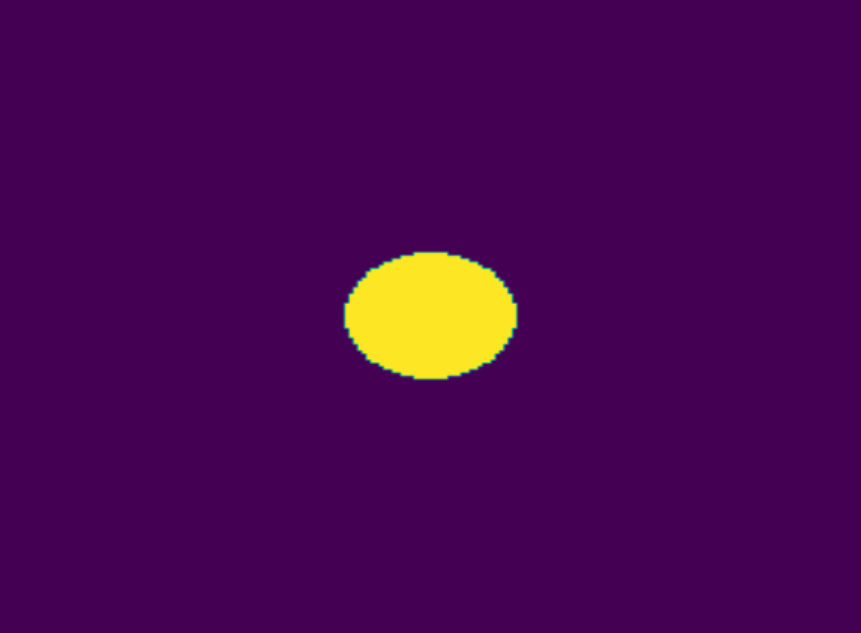};
    
    \nextgroupplot[title=$m_0^\rho$,xtick=\empty,ytick=\empty,yticklabel pos = right, xmin=0, xmax=1,ymin=0,ymax=1]
    \addplot graphics[xmin=0,xmax=1,ymin=0,ymax=1] {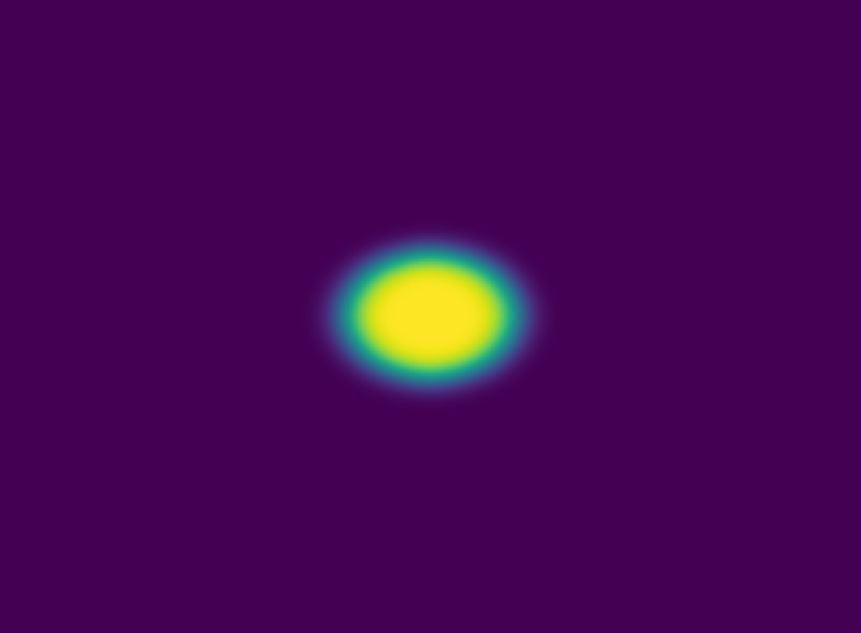};
    
    \nextgroupplot[title=$m^\rho$,xtick=\empty,ytick=\empty,yticklabel pos = right, xmin=0, xmax=1,ymin=0,ymax=1]
    \addplot graphics[xmin=0,xmax=1,ymin=0,ymax=1] {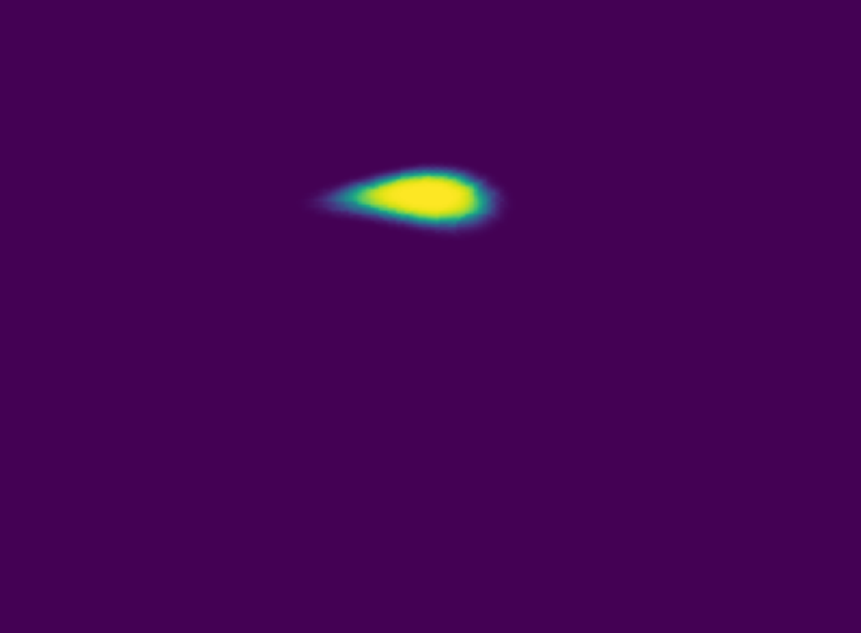};
    
  \end{groupplot}
\end{tikzpicture}
  \caption{
  Illustration of the three steps used to generate the data in experiment (S).
}
  \label{c:supfig}
\end{figure*}
\\
\\
\noindent
\textbf{Step 1.} A ball $m_0$ is generated in 2D. \\ \\
\noindent
\textbf{Step 2.} The function $m_0^\rho$ is generated by convolving $m_0$ with an isotropic Gaussian (zero covariance) with variance $\rho^2$. \\ \\
\noindent 
\textbf{Step 3.} Finally, $m^\rho$ is generated by randomly deforming $m_0^\rho$ with a continuous Gaussian field. \\
\\
\noindent 
Code is attached together with a Python requirements file.
Since the data is synthetic, there is no privacy or license concerns.
The experiment is CPU based and conducted on a Windows 10 machine with Python 3.10.4.

% \textbf{Step 1.} First, we start by generating a ball in 2D
% \begin{align}
%     m_0(\omega) = I\{ (\omega_1-0.5)^2 + (\omega_2-0.5)^2 \le 0.2^2\}, \quad \omega \in \Omega=[0,1]^2.
% \end{align}
% \textbf{Step 2.}
% We then convolve this with a centered Gaussian with variance $\rho$ in each dimension
% \begin{align}
%     m_0^\rho(\omega) = \int_\Omega K_\rho(\omega-\omega')K_\rho(\omega-\omega') \lambda(d\omega'), \quad \omega \in \Omega,
% \end{align}
% where
% \begin{align}
%     K_\rho(x) =
%     \frac{1}{\rho\sqrt{2\pi}} \exp\left\{ -\frac{1}{2}\frac{x^2}{\rho^2} \right\}, \quad  x\in(-\infty,\infty).
% \end{align}
% \textbf{Step 3.}
% A random deformation $T$ is generated by convolving whitenoise with the same Gaussian kernel
% \begin{align}
%     T(\omega_1,\omega_2) = \int_\Omega K_{0.1}(\omega_1-\omega_2')K_{0.1}(\omega_2-\omega_2') dB(\omega').
% \end{align}
% and used to randomly deform the marginal function $m_0^\rho$ by taking
% \begin{align}
%     m^\rho(\omega) = m_0(\omega - 0.1\cdot T(\omega)), \quad \omega\in\Omega.
% \end{align}
% The choice of constants are hand tuned by the authors to make interesting looking marginal functions.

{\small
\bibliographystyle{ieee_fullname}
\bibliography{bibliography}
}